\documentclass[10pt,twocolumn,letterpaper]{article}

\usepackage[final]{cvpr}      

\usepackage{graphicx}
\usepackage{amsmath}
\usepackage{amssymb}
\usepackage{booktabs}
\usepackage{xspace}
\usepackage{wrapfig}
\usepackage{xspace}
\usepackage{multirow}
\usepackage[font=small,labelfont=bf]{caption}
\usepackage[numbers,sort]{natbib}
\usepackage{amssymb}
\usepackage{commath}
\usepackage{adjustbox}
\usepackage{mathtools}    
\usepackage{verbatim} 
\usepackage{soul} 
\usepackage{shortbold}
\usepackage{xcolor,colortbl}

\definecolor{ibm1}{HTML}{648FFF}
\definecolor{ibm2}{HTML}{DC267F}
\definecolor{ibm3}{HTML}{FE6100}
\definecolor{ibm4}{HTML}{FFB000}
\definecolor{ibm5}{HTML}{785EF0}
\definecolor{ibm6}{HTML}{88CCEE}
\usepackage[pagebackref,breaklinks,colorlinks]{hyperref}

\newcommand{\E}{\textrm{E}_S}
\newcommand{\F}{\Phi}

\DeclareMathOperator{\sign}{sgn}
\newcommand{\parnobf}[1]{\vspace{1mm} \par \noindent {\bf {#1}.}}
\newcommand{\parbf}[1]{\vspace{0.0mm} \par \noindent {\bf {#1}}}

\newcommand{\parnoit}[1]{\noindent {\it {#1}.\xspace}}

\newcommand{\matterport}[0]{Matterport3D \cite{chang2017matterport3d}\xspace}
\newcommand{\scannet}[0]{ScanNet \cite{dai2017scannet}\xspace}
\newcommand{\tdf}[0]{3DFront \cite{fu20203dfront}\xspace}
\newcommand{\shapenet}[0]{ShapeNet \cite{chang2015shapenet}\xspace}

\newcommand{\drdf}[0]{DRDF\xspace}
\newcommand{\tsdf}[0]{DRDF\xspace}
\newcommand{\urdf}[0]{URDF\xspace}  
\newcommand{\udf}[0]{URDF\xspace} 
\newcommand{\scened}[0]{UDF\xspace}
\newcommand{\mpd}[0]{LDI\xspace}

\newcommand{\sal}[0]{SAL\xspace}
\newcommand{\occ}[0]{ORF\xspace}

\newcommand{\dur}{d_\textrm{UR}}
\newcommand{\durp}{d_{\textrm{UR}'}}
\newcommand{\dsr}{d_\textrm{SR}}
\newcommand{\dsrp}{d_{\textrm{SR}'}}
\newcommand{\ddr}{d_\textrm{DRDF}}
\newcommand{\dor}{d_\textrm{ORF}}

\newcommand{\rayr}{\overrightarrow{\rB}}

\definecolor{dred}{RGB}{242, 220, 219}
\definecolor{dblue}{RGB}{220, 230, 242}
\definecolor{dpurple}{RGB}{235, 212, 225}

\newcolumntype{m}{>{\columncolor{dred}}c}
\newcolumntype{t}{>{\columncolor{dblue}}c}
\newcolumntype{s}{>{\columncolor{dpurple}}c}

\newcommand{\redtext}[1]{\textcolor{red}{#1}}

\newcommand\blfootnote[1]{%
  \begingroup
  \renewcommand\thefootnote{}\footnote{#1}%
  \addtocounter{footnote}{-1}%
  \endgroup
}

\newcommand{\hlc}[2][yellow]{{%
    \colorlet{foo}{#1}%
    \sethlcolor{foo}\hl{#2}}%
}

\definecolor{tgreen}{HTML}{19CC19}
\definecolor{tblue}{HTML}{1919CC}
\definecolor{tgrey}{HTML}{AAAAAA}
\definecolor{tred}{HTML}{CC0000}
\definecolor{ptgold}{HTML}{DCC709}

\usepackage[capitalize]{cleveref}
\crefname{section}{Sec.}{Secs.}
\Crefname{section}{Section}{Sections}
\Crefname{table}{Table}{Tables}
\crefname{table}{Tab.}{Tabs.}

\def\cvprPaperID{2212} 
\def\confName{CVPR}
\def\confYear{2022}

\begin{document}

\title{What's Behind the Couch? Directed Ray Distance Functions (DRDF) for 3D Scene Reconstruction} 

\author{Nilesh Kulkarni

\and
Justin Johnson\\
University of Michigan\\
{\tt\small \{nileshk, justincj, fouhey\}@umich.edu}
\and
David F. Fouhey
}
\maketitle

\begin{abstract}
We present an approach for full 3D scene reconstruction from a single unseen image. We train on dataset of realistic non-watertight scans of scenes. Our approach
predicts a distance function, since these have shown
promise in handling complex topologies and large spaces. We identify and
analyze two key challenges for predicting such image conditioned distance functions that have prevented their success on real 3D scene data. First, we show that predicting a conventional scene distance from an image
requires reasoning over a large receptive field. Second, we analytically show that the
optimal output of the network trained to predict these distance functions does not obey all the distance function properties. We propose an alternate distance function, the {\it Directed Ray
Distance Function} (DRDF), that tackles both challenges. We show that 
a deep network trained to predict DRDFs outperforms all other methods quantitatively
and qualitatively on 3D reconstruction from single image on Matterport3D, 3DFront, and ScanNet.\blfootnote{Project Page \url{https://nileshkulkarni.github.io/scene_drdf}}
\vspace{-2ex}
\end{abstract}

\section{Introduction}

\definecolor{uppink}{HTML}{FA8F8D}
\definecolor{uppurple}{HTML}{7F7FFE}

\begin{figure}[t]
    \centering
    \noindent
    \begin{tabular}{cc}
        (a) Image & (b) Ray through the scene \\
    \noindent\includegraphics[width=0.4\linewidth]{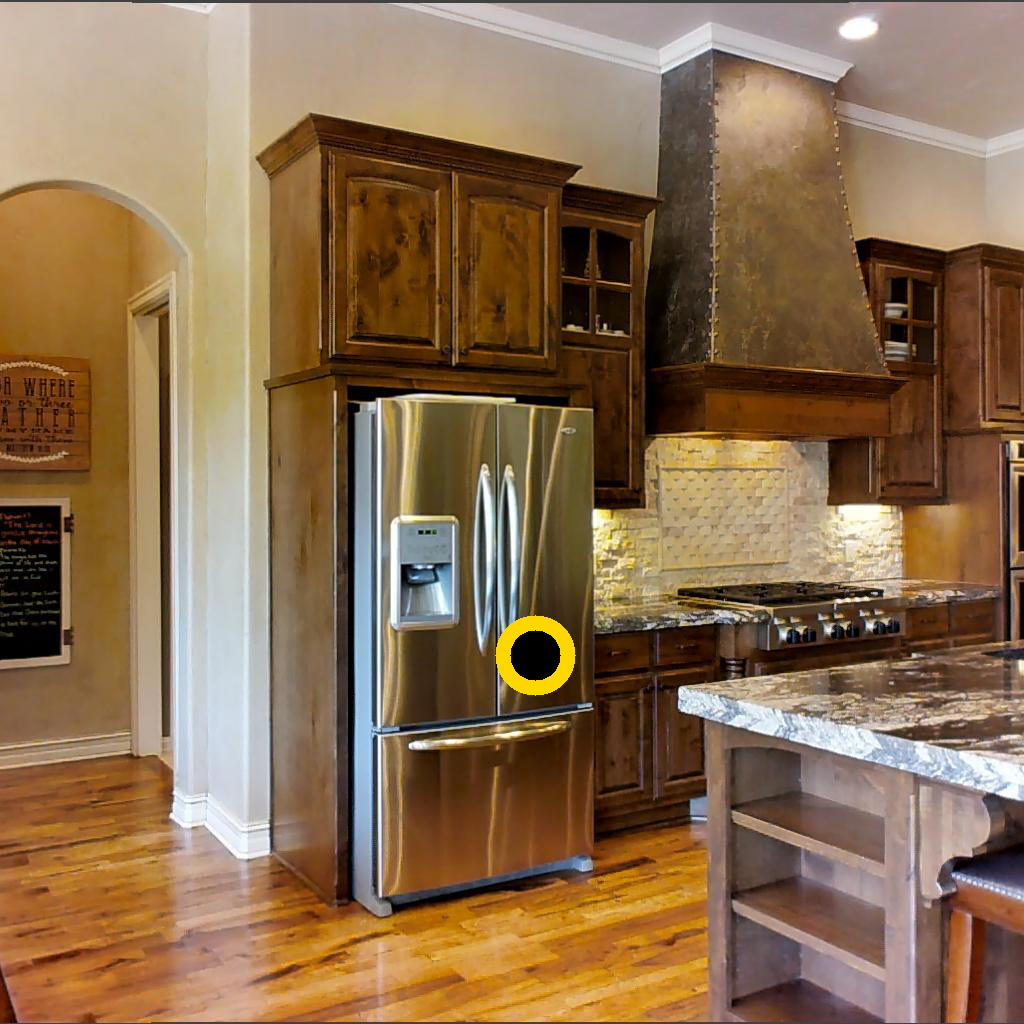} &
    \noindent\includegraphics[width=0.50\linewidth]{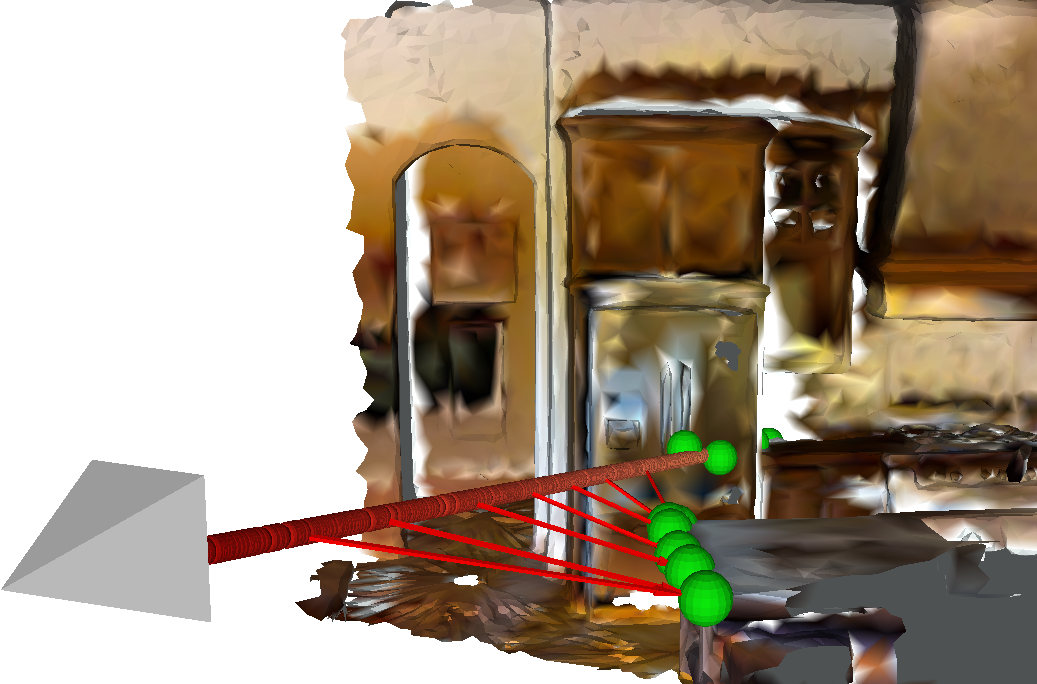} \\
    \noindent\includegraphics[width=0.5\linewidth]{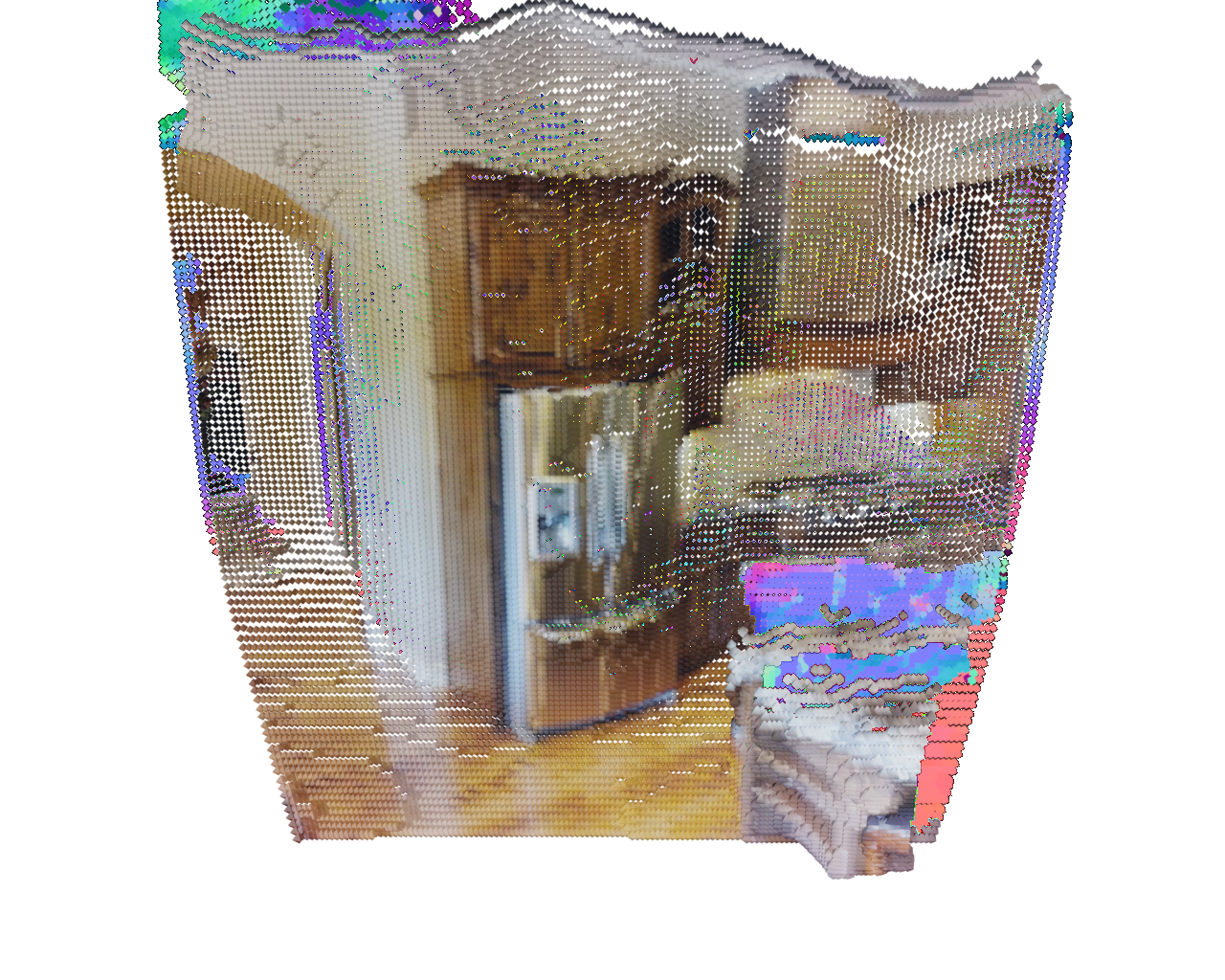} &
    \noindent\includegraphics[width=0.5\linewidth]{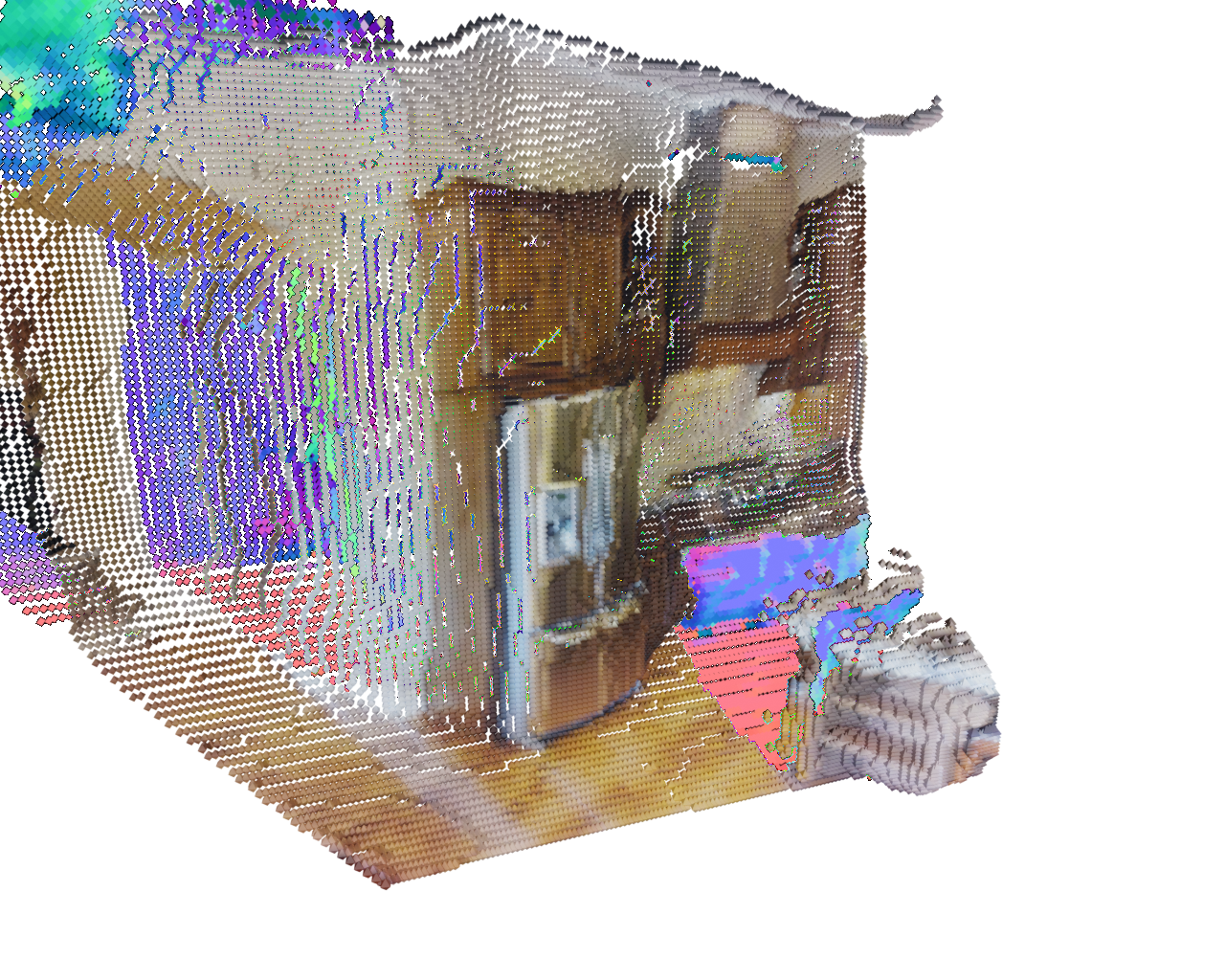} \\
    \multicolumn{2}{c}{(c) 3D outputs rendered from our model} 
    \end{tabular}
    \caption{Given a single input image (a) our model generates its full 3D. In (b) we show a third person view of the scene with a \textcolor{tred}{red-ray} from \textcolor{tgrey}{camera}. The ray projects at the \textcolor{ptgold}{yellow-dot} in the image (a). Nearest points to the ray shown as \textcolor{tgreen}{green} spheres.   In (c) we show the full 3D as two rendered novel views of our method's 3D output revealing the predicted \hlc[uppurple]{occluded cabinet} and \hlc[uppink]{floor}. Visible surfaces are colored with image pixels; occluded ones show surface normals (\hlc[uppink]{pink: upwards}; \hlc[uppurple]{lavender: towards camera}). }

    \label{fig:teaser}
    \vspace{-5mm}
\end{figure}

Consider the image in Figure~\ref{fig:teaser}. What happens if you look behind the kitchen counter? To a layman, this single image shows a rich 3D world in which the floor continues behind the counter, and there are cabinets below the kitchen top. Our work aims to learn a mapping from a single image to the complete 3D, including visible {\it and} occluded surfaces. We learn such mapping from real, unstructured scans like  \matterport or \scannet. Unstructured scans are currently one of the richest sources of real-world 3D ground truth, and as more sensors like LIDAR scanners become ubiquitous, their importance will only grow.

Learning from these real-world scans poses significant challenges to the existing methods in computer vision.
Voxel-based methods~\cite{girdhar2016learning,choy20163d} scale poorly with
size due to their memory requirements, and mesh-based ones~\cite{wang2018pixel2mesh} struggle with varying
topology. Implicit functions~\cite{mescheder2019occupancy,saito2019pifu} 
show promise for overcoming these size and topology challenges,
but mostly focus on watertight
data~\cite{mescheder2019occupancy,chen2019learning,park2019deepsdf,saito2019pifu,sitzmann2020implicit} with a 
well-defined inside and outside regions for objects. This watertightness property enables {\it
signed distance functions} (SDF) or occupancy functions, but
limits them to data like \shapenet,
humans~\cite{saito2019pifu}, or memorizing single watertight
scenes~\cite{sitzmann2020implicit}. Real 3D scans like
\matterport are off-limits for these methods.  
Exceptions include~\cite{chibane2020ndf}, which fits a single model
with {\it unsigned distance function} (UDF) to a scene, and
SAL~\cite{atzmon2020sal,atzmon2020sal++} which learns SDFs on objects with
well-defined insides and outsides that have scan holes.
We believe that the lack of success in predicting implicit functions conditioned on an unseen image on datasets like \matterport stems from two key challenges. 

First, conventional distance functions depend on the distance to the nearest
point in the full 3D scene. We show that this requires complex reasoning across
an image. To see this, consider Fig.~\ref{fig:teaser}. The \textcolor{ptgold}{yellow point}
in (a) is the projection of the \textcolor{tred}{red ray} in (c). We show the
nearest point in the scene to each point on the ray in \textcolor{tgreen}{green}. 
Near the \textcolor{tgrey}{camera}, these are all over the kitchen counter to the right. Closer to
the refrigerator, they finally are on the refrigerator. This illustrates that
the projection of the nearest points to a point is often far from the
\textcolor{ptgold}{projection of that point}. Models estimating scene distances must
integrate information across vast receptive fields to find the nearest points, which
makes learning hard. We examine this in more detail in \S\ref{sec:scene_vs_ray}.

Second,  we will show that the uncertainty in predicting 3D from an unseen single image incentivizes predicted distance functions to not actually have {\it critical} properties of distance functions. The root cause of such behavior is that networks trained with a MSE loss are optimal when they produce the expectation, ($\mathbb{E}$). This issue is common to other domains like colorization~\cite{zhang2016colorful,pix2pix2017} and 3D pose estimation~\cite{Chen_2021_CVPR,Wang15}. In \S\ref{sec:methoduncertainty}, we analytically derive the expected value of multiple distance functions under a model where the surface location has Gaussian noise. We show that even before accounting for other factors, this simple setting ensures that the optimal output lacks basic distance function properties (e.g., reaching 0, a derivative of $\pm1$) that is required to extract surfaces from them.

We propose to overcome these issues with a new distance-like function named
the {\it Directed Ray Distance Function} (DRDF). Unlike the {\it Unsigned Distance Function}
(defined by the nearest points in the scene), the DRDF is defined by
points along the ray through a pixel; these project to the
same pixel, facilitating learning. Unlike standard distance functions, DRDF's expected value under uncertainty behaves like a true distance function close to the surface. We learn to predict the DRDF with a PixelNerf~\cite{yu2020pixelnerf}-style architecture and compare it 
with other distance functions. We also compare it to other conventional methods such as Layered Depth Images
(LDI)~\cite{shade1998layered}. Our experiments (\S\ref{sec:experiments}) 
on \matterport, \tdf, and \scannet show that
the DRDF is substantially better at 3D scene recovery (visible and occluded) across
all (three) metrics.


\section{Related Work}
\label{sec:related}

Our approach aims to infer the full 3D structure of a scene from a single image using implicit functions, which relates with many tasks in 3D computer vision. 

\parnobf{Scenes from a Single Image}
Reconstructing the 3D scene from image cues is a long-term goal of computer vision. Most early work in 3D learning focuses on 2.5D properties~\cite{barrow1978recovering} that are visible in the image, like qualitative geometry~\cite{hoiem2005geometric,fidler20123d},  depth~\cite{saxena2008make3d} and normals~\cite{fouhey2013data}. Our work instead aims to infer the full 3D of the scene, including invisible parts. Most work on invisible surfaces focuses on single objects with voxels~\cite{girdhar2016learning,choy20163d,hane2017hierarchical}, point-clouds~\cite{lin2017learning,fan2017point}, CAD models ~\cite{izadinia2017im2cad} and meshes~\cite{gkioxari2019mesh,groueix2018papier}. These approaches are often trained with synthetic data, e.g., \shapenet or images that have been aligned with synthetic ground-truth 3D~\cite{sun2018pix3d}. Existing scene-level work, e.g.,~\cite{tulsiani2017learning,LiSilhouette19,kulkarni20193d,nie2020total3dunderstanding} trains on synthetic datasets with pre-segmented, watertight objects like SunCG~\cite{song2017semantic}. Our work instead can be learned on real 3D like \matterport. In summary, our work aims to understand the interplay between 3D, uncertainty, and learning~\cite{kendall2017uncertainties,poggi2020uncertainty,bae2021estimating} that has largely been explored in the depth-map space.

\parnobf{Implicit Functions for 3D Reconstruction} 
We approach the problem with learning implicit functions~\cite{mescheder2019occupancy,park2019deepsdf,chen2019learning}, which have shown promise in addressing scale and varying topology. These implicit functions have also been used in novel view synthesis~\cite{mildenhall2020nerf,martin2020nerf,zhang2020nerfplusplus,yu2020pixelnerf}, which differs from our work in goals. In reconstruction, implicit functions have shown impressive results on two styles of task: fitting to a single model to a fixed 3D scene (e.g., SIREN~\cite{sitzmann2020implicit,chibane2020ndf}) and predicting new single objects (e.g., PIFu~\cite{saito2019pifu,xu2019disn}). Our work falls in the latter category as it predicts new scenes. While implicit functions have shown results on humans~\cite{saito2019pifu,saito2020pifuhd} and ShapeNet objects~\cite{xu2019disn}, most work relies on watertight meshes. Our non-watertight setting is more challenging. Two solutions have been proposed: assuming the SDF's existence and supervising it indirectly (SAL: ~\cite{atzmon2020sal,atzmon2020sal++}), and predicting an unsigned distance function (UDF)~\cite{chibane2020ndf} -- we stress that \cite{chibane2020ndf} does not predict from RGB images. Our work can be trained with non-watertight 3D meshes and outperforms these approaches.

\parnobf{Recovering Occluded Surfaces} 
Our system produces the full 3D of a scene, including occluded parts, from a single image. This topic has been of interest to the community beyond previously mentioned volumetric 3D work (e.g.,~\cite{girdhar2016learning,choy20163d}). Early work often used vanishing-point-aligned box~\cite{hedau2009recovering,Pero_2013_CVPR} trained on annotated data. While our approach predicts floors, this is learned, not baked in, unlike modern inheritors that have explicit object and layout components~\cite{tulsiani2018factoring,jiang2020peek} or the ability to query for two layers~\cite{issaranon2019counterfactual,jiang2020peek}. An alternate approach is layered depth images (LDI)~\cite{shade1998layered,dhamo2019object} or multi-plane depthmaps. LDIs can be learned without supervision~\cite{lsiTulsiani18}, but when trained directly, they fare worse than our method.
\section{Learning Pixel Aligned Distance Functions}

\label{sec:methodbase}
\begin{figure*}[t]
    \centering
    \includegraphics[width=\textwidth]{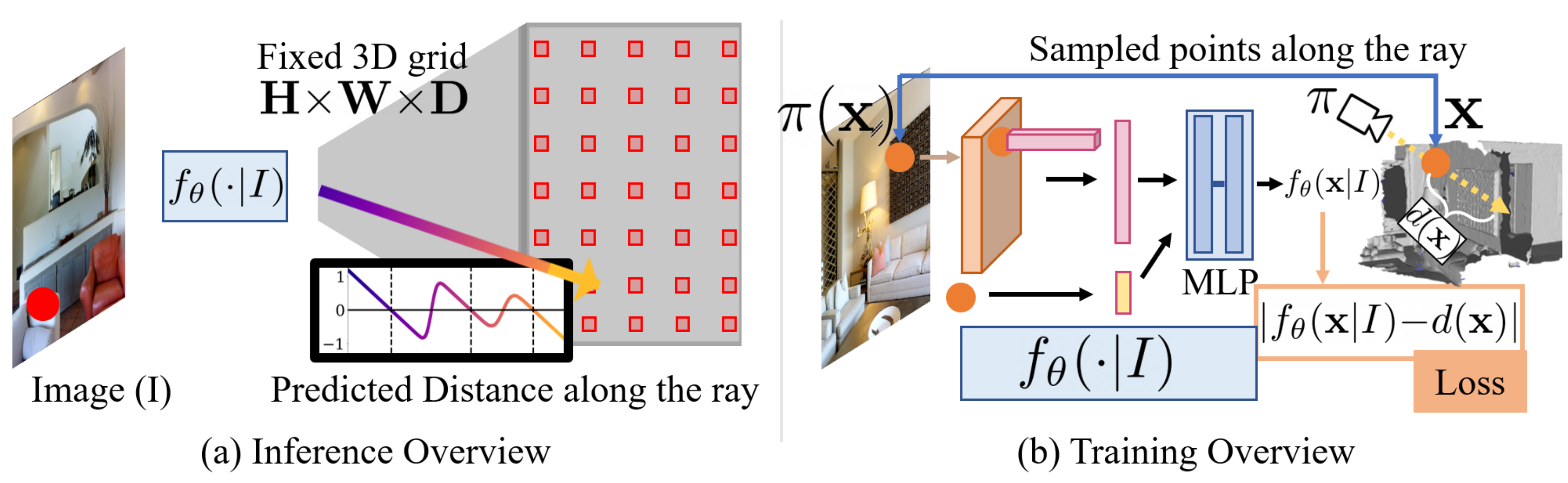}
    \caption{{\bf Approach Overview.} (a) At inference our model, $f_{\theta}(\cdot|I)$, conditioned on an input image ($I$) predicts a pixel conditioned distance for each point in a 3D grid. This frustum volume is then converted to surface locations using a  decoding strategy. (b) At training time, our model takes an image and set of 3D points. It is supervised with the ground truth distance function for scene at these points. More details in \S\ref{sec:methodbase}.}
    \label{fig:overview}
    \vspace{-5mm}
\end{figure*}
\label{sec:method}

We aim to reconstruct the full 3D of an unseen scene, including occluded parts, from a single RGB image while training on real scene captures~\cite{chang2017matterport3d}. Towards this goal we train an image-conditioned neural network to predict distance functions for 3D points in the camera frustum. Our training set consists of images and 3D meshes for supervision. We supervise our network with any ground-truth distance function \eg, the {\it Unsigned Distance Function} (UDF)

At test time, we consider only a single input image and a fixed set of 3D points in the camera view frustum. Our network predicts the distance function for each point using pixel aligned image features. The inference produces a grid of distances instead of a surface; we extract a surface with a {\it decoding strategy} (e.g., a thresholding strategy that defines values close to zero as surface locations).

Our setup is generic and can be paired with any distance function and a decoding strategy. We will discuss particular distance functions and decoding strategies later while discussing experiments. Experimentally, we will show that commonly used distance functions~\cite{chibane2020ndf,atzmon2020sal} do not work well when they are predicted in pixel conditioned way from a single image when trained on raw 3D data.

\parnobf{Inference} Given a input image like in Fig.~\ref{fig:overview} (left), we evaluate our model $f_{\theta}(\xB; I)$ on pre-defined grid of points, $H{\times}W{\times}D$, in the 3D camera frustum to predict the distance function. It is then {\it decoded} to recover surface locations.

\parnobf{Training} At train time we are given $n$ samples $\{(\xB_i, I_i, d(\xB_i)\}_{i=1}^n$ representing the 3D points ($\xB_{i}$), input image ($I_{i}$) and the ground truth distance, $d(\xB_{i})$, computed using the 3D mesh. We find parameters $\theta$ that minimize the empirical risk $\frac{1}{n}\sum_{i=1}^n \mathcal{L}(f_\theta(\xB_i,I_i),d(\xB_{i}))$ with a loss function $\mathcal{L}$ \eg the  L1-Loss. 

\parnobf{Model Architecture} We use a PixelNerf~\cite{yu2020pixelnerf}-like architecture containing an encoder and multi layer perceptron (MLP). The encoder maps the image $I$ to a feature map $\BB$. Given a point $\xB$ and the camera ($\pi$) we compute its projection on the image $\pi(\xB)$. We extract a feature at $\pi(\xB)$ from $\BB$ with bilinear interpolation; the MLP uses the extracted image feature and a positional encoding~\cite{mildenhall2020nerf} of $\xB$ to make a final prediction $f_\theta(\xB;I)$.

\section{Behavior of Pixel Conditioned Distance Functions}

\label{sec:methoddistances}

Recent works have demonstrated overfitting of neural networks to single scenes \cite{chibane2020ndf,sitzmann2020implicit,atzmon2020sal,atzmon2020sal++} but none attempt to predict {\it scene-level 3D} conditioned on an image. We believe this problem has not been tackled due to two challenges. First, predicting a standard scene distance from a single image requires reasoning about large portions of the image. As we will show in \S\ref{sec:scene_vs_ray}, this happens because predicting scene distance for a point $\xB$ requires finding the nearest point to $\xB$ in the scene. This nearest point often projects to a part of the image that is far from $\xB$'s projection in the image. Secondly, we will show in \S\ref{sec:methoduncertainty} that the uncertainty present in predicting pixel conditioned distance function incentivizes networks to produce outputs that lack basic distance function properties. These distorted distance functions do not properly decode into surfaces. To overcome the above challenges, we introduce a new distance function, the {\it Directed Ray Distance Function (DRDF)}. We will show analytically that {\drdf} retains distance-function like properties near the 3D surface under uncertainty.

All distance functions are denoted with $d(\xB)$ where $\xB$ is the query point in 3D space. We use $M$ to denote the mesh of the 3D scene and $\rayr$ to denote the ray originating from the camera passing through $\xB$.

\subsection{Scene \vs Ray Distances}
\label{sec:scene_vs_ray}

\begin{figure*}[t]
    \centering
    \noindent
    \begin{adjustbox}{max width=\linewidth}
    \begin{tabular}{c@{\hskip4pt}c@{\hskip1pt}c@{\hskip1pt}c}
    \noindent\includegraphics[width=0.30\linewidth]{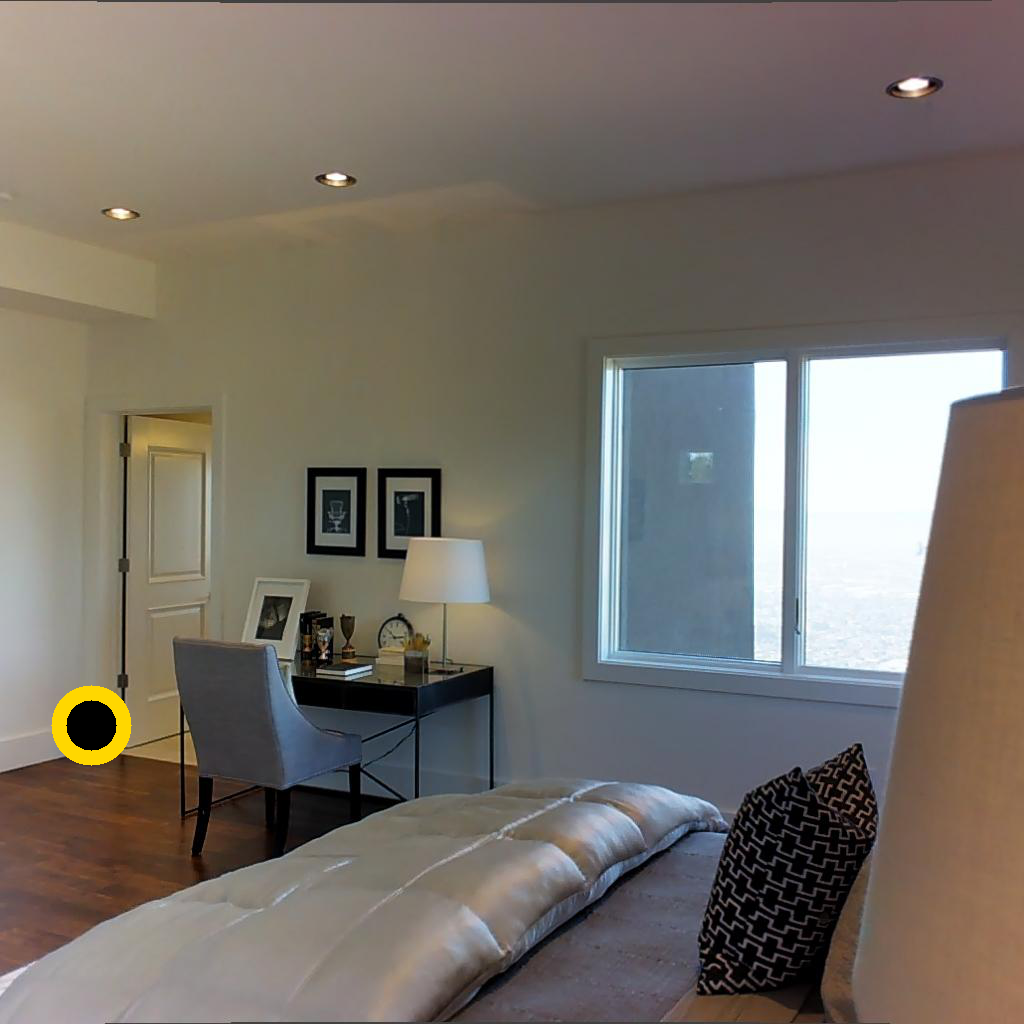} &
    \noindent\includegraphics[width=0.50\linewidth]{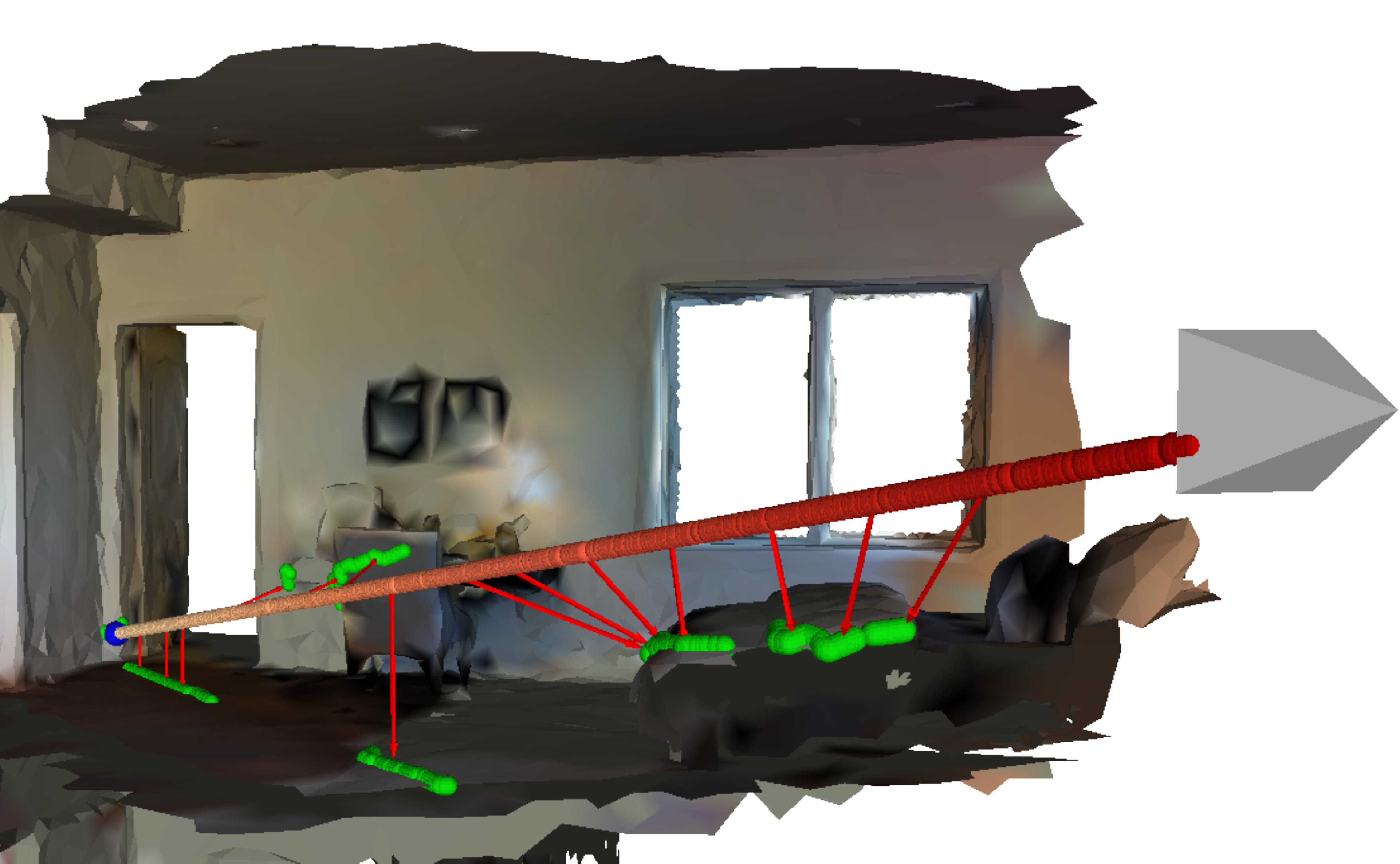} &
    \noindent\includegraphics[width=0.25\linewidth]{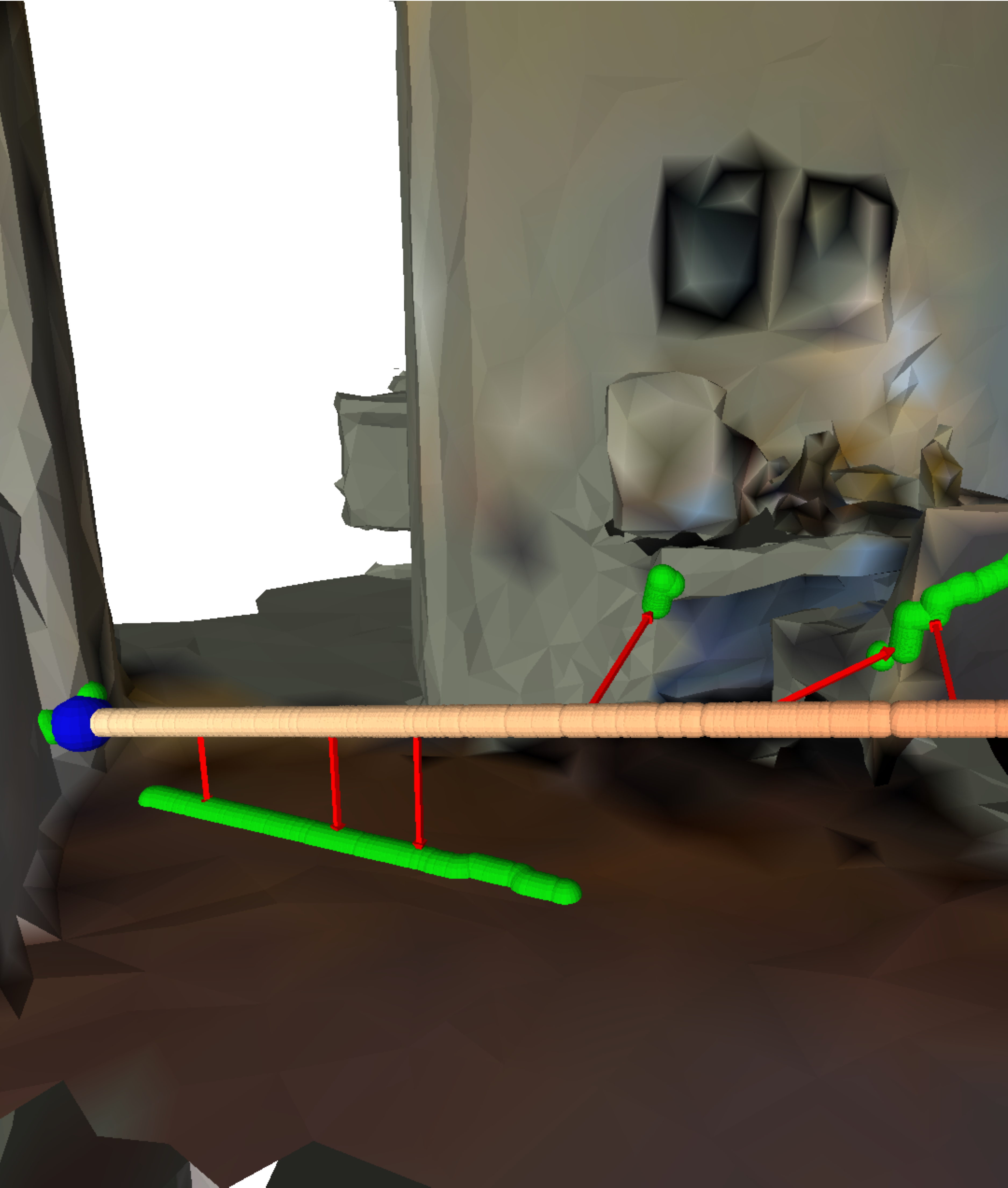} &
    \noindent\includegraphics[width=0.30\linewidth]{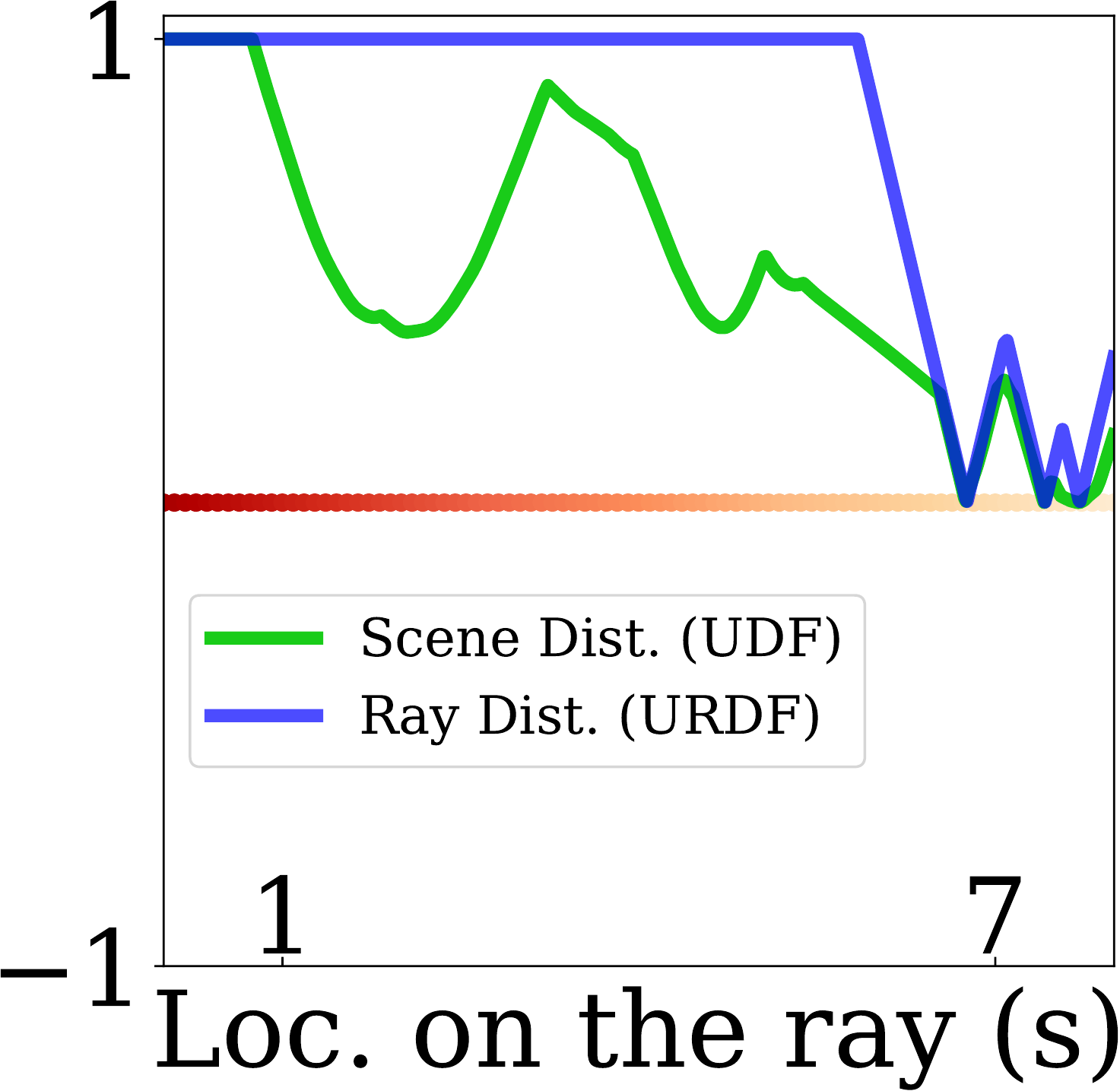} \\
    (a) Image with ray center & \multicolumn{2}{c}{(b) Third person 3D views with the \textcolor{tred}{red ray} and \textcolor{tgreen}{nearest points}} & (c) Plot of Dist. Func.
    \end{tabular}
    \end{adjustbox}
    \caption{
    {\bf Scene vs ray distances }. (a) The \textcolor{red}{red ray} intersects the scene at the black and \textcolor{ptgold}{yellow point} in the image. 
\textcolor{tgreen}{Scene} vs \textcolor{tblue}{Ray} distances along the points on \textcolor{tred}{red-shaded ray} through the \textcolor{tgrey}{camera}. 
(b) Two different 3D views showing \textcolor{tblue}{intersections between the ray and the scene (which define the ray distance) in blue} and \textcolor{tgreen}{the nearest points in the scene to the ray in green}. These nearest scene points define scene distance. A network predicting scene distance must look all over the image (e.g., looking at the bed and chair to determine it for the ray). (c) Ground truth \textcolor{tgreen}{scene} \vs  \textcolor{tblue}{ray} distance functions for points on the ray. There are occluded intersections not visible in the image.}
    \label{fig:ray_vs_scene}
    \vspace{-5mm}
\end{figure*} 
A standard scene distance for a point $\xB$ in a 3D scene $M$ is the minimum distance from $\xB$ to the points in $M$. If there are no modifications, this distance is called the {\it Unsigned Distance Function} (\scened) and can be operationalized by finding the nearest point $\xB'$ in $M$ to $\xB$ and returning $||\xB-\xB'||$. We now define a {\it ray distance} for a point $\xB$ as the minimum distance of $\xB$ to any of the intersections between $\rayr$ and $M$, which is operationalized similarly. The main distinction between scene \vs ray distances boils down to which points define the distance. 
When calculating scene distances, all points in $M$ are candidates for the nearest point. When calculating ray distances, only the intersections of $\rayr$ and $M$ are candidates for the nearest point. These intersections are a much smaller set.

We will now illustrate the above observation qualitatively with Fig.~\ref{fig:ray_vs_scene}. We show in Fig~\ref{fig:ray_vs_scene}(a) the projection of $\rayr$ (and all points on it) onto the image as the \textcolor{ptgold}{yellow-center}. We show in (b) a third person view of the scene with $\rayr$ as the \textcolor{tred}{red-shaded-ray}. We show the intersection point of $\rayr$ with the scene $M$ as \textcolor{tblue}{blue points}.
For each point on the red ray, we show the nearest point on the mesh in \textcolor{tgreen}{green} with an arrow going to that green point. The scene distance for points on the ray is defined by these \textcolor{tgreen}{nearest points in green}. These \textcolor{green}{nearest points} are distributed all over $M$ including the  floor, bed, and chair, \etc. A pixel conditioned neural network predicting scene distances needs to integrate information of all the green regions to estimate scene distance for points projecting to the yellow ray projection.  To show that this is not an isolated case, we quantify the typical projection of nearest points for scene distance to get an estimate of the minimum receptive field need to predict a distance using a neural network. We measure the distance between projections of the \textcolor{tgreen}{nearest points} from the \textcolor{ptgold}{ray center}.
The average maximum distance to the ray center is $0.375{\times}\textit{image width}$ -- averaged over 50K rays on \matterport. Thus, a neural network predicting the scene distance needs to look at least this far to predict it.

This problem of integrating evidence over large regions vanishes for a ray distance function. By definition, the only points involved in defining a ray distance function for a point $\xB$ lie on the ray $\rayr$ since they are at the intersection of the mesh and the ray; these points project to the same location as $\xB$. This simplifies a network's job considerably. We define the {\it Unsigned Ray Distance Function} (URDF) as the Euclidean distance to the nearest ray intersection.

We finally plot  the UDF (scene) and URDF (ray) for the points along \textcolor{red}{the red $\rayr$}, both truncated at $1$m, in Fig~\ref{fig:ray_vs_scene}(c). The UDF is fairly complex because different parts of the scene are nearest to the points along the ray at different distances. In contrast, the URDF is piecewise linear due to the few points defining it. We hypothesize this simplified form of a ray distance aids learning.

\newcommand{\xr}{\mathrm{x}}

\subsection{Ray Distance Functions}
It is convenient when dealing with a ray $\rayr$ to parameterize each point $\xB$ on the ray by a scalar multiplier $z$ such that $\xB = z\rayr$. Now the distance functions are purely defined via the scalar multiplier along the ray. Suppose we define the set of scalars along the ray $\rayr$ that correspond to intersections as $D_{\rayr}=\{s_{i}\}_{0}^{k}$ (i.e., each point $s\rayr$ for $s \in D_{\rayr}$ is an intersection location). We can then define a variety of ray distances using these intersections. For instance, given any point along the ray, $z\rayr$, we can define $\dur(z)=\min_{s \in D_{\rayr}}\norm{s-z}$ as the minimum distance to the intersections. As described earlier, we call this {\it Unsigned Ray Distance Function} (\urdf) -- {\it R} here indicates it is a ray distance function. For watertight meshes, one can have a predicate $\textrm{inside}(\xB)$ that is $1$ when $\xB$ is {\it inside} an object and $-1$ otherwise. We can then define the {\it Signed Ray Distance Function} (SRDF) as $\dsr(z) = -\textrm{inside}(z\rayr) \dur(z)$. Signed functions are standard in the literature but since our setting is non-watertight, the SRDF is impossible. Now we show how to modify SRDF for non-watertight settings. 

\begin{figure}[t]
    \centering
    \noindent\includegraphics[width=0.39\textwidth]{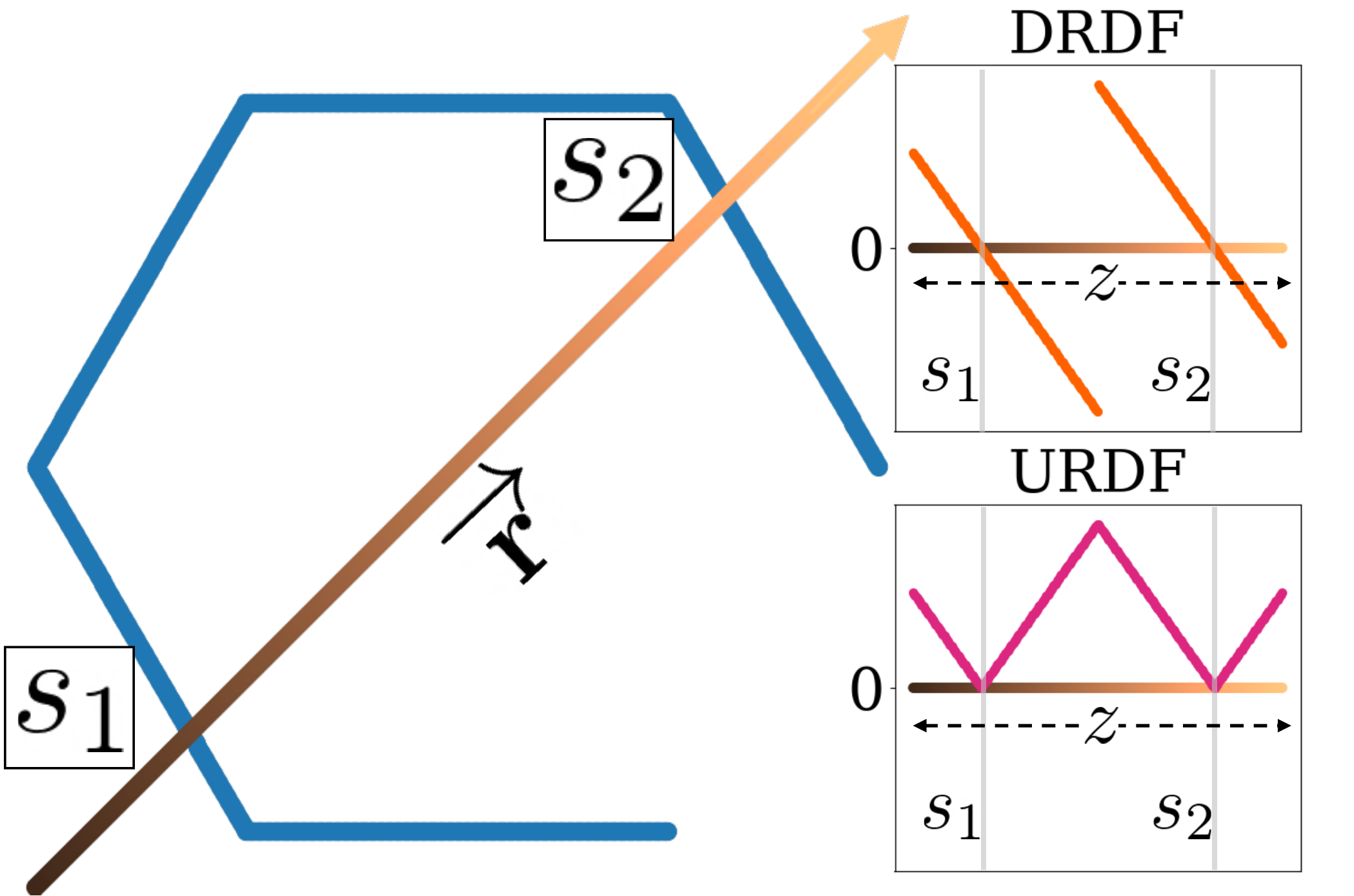} 
    \captionof{figure}{\drdf \vs \urdf in case of two intersections along the ray. Unlike \urdf, \drdf is positive and negative}
    \label{fig:toy_drdf}
\end{figure}

\parbf{Directed Ray Distance Function.} We introduce a new ray based distance function  called the {\it Directed Ray Distance Function} (DRDF). This can be seen as a modification to both \urdf and SRDF; We define $\ddr(z) = \textrm{direction}(z)\dur(z)$ where our predicate $\textit{direction}(z)$ is $\sign(s - z)$ where $s$ is the nearest intersection to $z$. In practice \drdf is positive before the nearest intersection and negative after the nearest intersection. We call it {\it Directed} because the sign depends on the positioning along the ray. Unlike SRDF, there is no notion of inside, so the \drdf can be used with unstructured scans. Near an intersection, \drdf behaves like SRDF and crosses zero. \drdf has a sharp discontinuity midway between two intersections due to a sign change. We will analyze the importance of adding directional behavior to \drdf in the subsequent sections. Fig.~\ref{fig:toy_drdf} shows the difference between \drdf \vs \urdf for multiple intersections on a ray.

\subsection{Uncertainty in Ray Distance Functions}
\label{sec:methoduncertainty}

When we predict distances in a single RGB image, the distance to an object in the scene is intrinsically uncertain. We may have a sense of the general layout of
the scene and a rough distance, but the precise location of each object to the millimeter is
not known. We investigate the consequences of this uncertainty for neural networks that predict
distance functions conditioned on single view images. We analyze a simplified setup that
lets us derive their optimal behavior. 

In particular, if the network minimizes the MSE (mean-squared-error), its optimal behavior is to produce the expected value. In many cases, the expected value is precisely what is desired like in object detection~\cite{tian2020fcos,zhou2019objects} or in ARIMA models~\cite{newbold1983arima}, weather prediction~\cite{weyn2020improving} but in others it leads to poor outcomes. For instance, in colorization~\cite{zhang2016colorful,pix2pix2017}, where one is uncertain of the precise hue, the expected value averages the options, leading to brown results; similar effects happen in rotation~\cite{Chen_2021_CVPR,mousavian20173d,kulkarni2019canonical} and 3D  estimation~\cite{Wang15,ku2019monocular,jin2021planar}.

We now gain insights into the optimal output by analyzing the expected distance functions under uncertainty about the location of a surface. For simplicity, we derive results along a ray, although the supplement shows similar results hold true for scene distances. Since there is uncertainty about the surface location, the surface location is no longer a fixed scalar $s$ but instead a random variable $S$. The distance function now depends on the value $s$ that the random variable $S$ takes on. We denote the ray distance at $z$ if the intersection is at $s$ as $d(z;s)$.
\begin{figure}[t]
    \centering
    \noindent\includegraphics[width=0.48\textwidth]{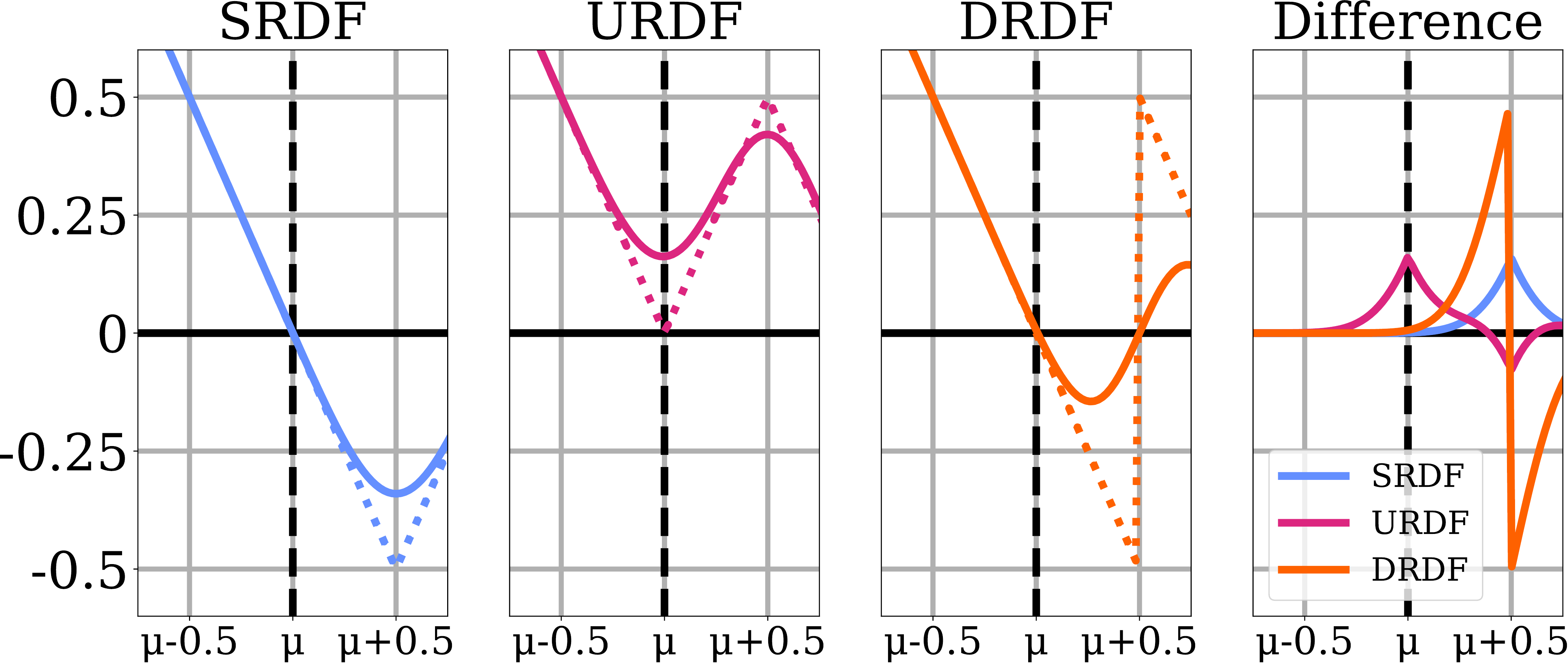} 
    \captionof{figure}{{\bf True \vs Expected distance functions under uncertainty}. Suppose the surface's location is normally distributed with mean $\mu$ at its true location and $\sigma{=}0.2$, and the next surface $1$ is unit away. We plot the expected (solid) and true (dashed) distance functions for the \textcolor{ibm1}{SRDF}, \textcolor{ibm2}{URDF}, and \textcolor{ibm3}{DRDF} and their difference (expected - true). The \textcolor{ibm1}{SRDF} and \textcolor{ibm3}{DRDF} closely match the true distance near the surface; the \textcolor{ibm2}{URDF} does not.
    }
    \label{fig:toy_dist}
\end{figure}

The network's output at a location $z$ is optimal in this setting if it equals the expected distance under $S$ or $\E[d(z;s)] = \int_{\mathbb{R}} d(z;s)  p(s) ds$ , where $p(s)$ is the density of $S$. Thus, by analyzing $\E[d(z;s)]$ we can understand the optimal behavior.  We note that this expectation is also optimal for other losses: under many conditions (see supp) $\E[d(z;s)]$ also is optimal for the L1 loss, and if $d(z;s)$ is $\{0,1\}$ such as in an occupancy function, then $\E[d(z;s)]$ is optimal for a cross-entropy loss.  For ease of derivation, we derive results for when $S$ is Gaussian distributed with its mean $\mu$ at the true intersection, standard deviation $\sigma$ and CDF $\F(s)$. Since distance functions also depend on the next intersection, we assume it is at $S$+$n$ for a constant $n \in R^{+}$.

We summarize salient results here, and a detailed analysis appears in the supplement.
Figure~\ref{fig:toy_dist} shows $\E[d(z;s)]$ for three ray distance functions (for $n=1, \sigma = 0.2$). No expected distance function perfectly matches its true function, but each varies in where the distortion occurs.  At the intersection, the expected SRDF and DRDF closely match the true function while the expected URDF is grossly distorted. Full derivations appear in the supplemental. The expected URDF has a minimum value of ${\approx}\sigma \sqrt{2/\pi}$ rather than $0$. Similarly, its previously sharp derivative is now ${\approx}2\F(z)-1$, which is close to $\pm 1$ {\it only} when $z$ is far from the intersection. In contrast, the expected DRDF's distortion occurs at $\mu+\frac{n}{2}$, and its derivative ($np(z-\frac{n}{2})-1$) is close to $-1$, except when $z$ is close to $\mu + \frac{n}{2}$.

These distortions in expected distance function disrupt the decoding of distance functions to surfaces. For instance, a true URDF can turned into to a surface by thresholding, but the expected URDF has an uncertainty-dependent minimum value ${\approx}\sigma \sqrt{2 / \pi}$, not $0$. Since a nearby intersection often has less uncertainty than a far intersection, a threshold that works for near intersections may miss far intersections. Conversely, a threshold for far intersections may dilate nearby intersections. One may try alternate schemes, e.g., using zero-crossings of the derivative. However, the expected URDF's shape is blunted; empirical results suggest that finding its zero-crossing is ineffective in practice.

DRDF is more stable under uncertainty and requires just finding a zero-crossing. The zero-crossing at the intersection is preserved except when $\sigma$ is large (e.g., $\sigma{=}\frac{n}{3}$) in such cases other distance functions also break down. This is because the distortion for DRDF occur halfway to the other intersection. The only nuance is to filter out second zero-crossing after the intersection based on the crossing direction. Further analysis appears in the supplemental.

\subsection{Implementation Details}
\label{sec:methodimplementation}
We present important implementation details here and rest in the supplement.
\parnobf{Training} Given samples
$\{\xB_i,I_i,d(\xB_i)\}_{i=1}^n$  we train our network to minimize the L1 loss,
$\frac{1}{n}\sum_{i=1}^n | d(\xB_i) - f_\theta(\xB_i,I_i)|$,
where the predictions are log-space truncated at $1$m following other methods that predict TSDFs~\cite{dai2020sg,sun2021neuralrecon}. We optimize using AdamW~\cite{loshchilov2017decoupled,kingma2014adam} with learning rate of $10^{-4}$ and weight decay of $10^{-2}$. 
We sample points $\xB_{i}$ for each scene in two ways: for each intersection at $l$ and the corresponding ray $\rayr$ through the pixel, we sample 512 points from $\mathcal{N}(l,0.1)$ along the $\rayr$; we additionally sample 512 points uniformly on $\rayr$ from $0$ to a maximum prediction distance.We train with 20 intersections/ scene for 250K iterations with 10 scenes/ batch  and freeze batch norm  after $\frac{1}{4}$th of the iterations.

\parnobf{Inference} At inference time, we extract backbone features at a regular $H\times W$ grid ($H{=}128, W{=}128$) in one forward pass of the backbone. For every ray corresponding to a grid point, we predict the distance function for $D=128$ points linearly spaced from $0$ to maximum prediction distance. This entails making $H{\times}W{\times}D$ predictions with the MLP yielding a frustum-shaped volume of locations with predictions. Methods vary in their {\it decoding} strategies to extract a surface. 
DRDF requires finding positive-to-negative zero-crossings which is trivial and hyperparameter free; we extensively optimize {\it decoding strategy} for baseline methods as explained in \S\ref{sec:experiments_baselines} and the supplement.

\section{Experiments}

\label{sec:experiments}

We evaluate DRDF on real images of scenes and compare it to alternate choices of distance functions as well as conventional approaches such as Layered Depth Images\cite{shade1998layered}. We extensively optimize decoding schemes for our baseline methods; detailed descriptions in appendix.
 
We evaluate each method's ability to predict the visible and occluded parts of the scene using standard metrics and a new metric that evaluates along rays.

\parnobf{Metrics}
We use three metrics. A single metric cannot properly quantify reconstruction performance as each metric captures a different aspect of the task~\cite{tatarchenko2019single}. The first is scene Chamfer errors. The others are accuracy/completeness~\cite{seitz2006comparison} and their harmonic mean, F1-score~\cite{tatarchenko2019single}, for scenes and rays (on occluded points).

\parnoit{Chamfer L1} We compute Symmetric Chamfer L1 error for each scene with 30K points sampled from the ground truth and the prediction. We plot the fraction of scenes with Symmetric Chamfer L1 errors that are less than t for $t \in [0,1]$m. It is more informative than just the mean across the dataset and compares performance over multiple thresholds.

\parnoit{Scene (Acc/Cmp/F1)} Like~\cite{seitz2006comparison,tatarchenko2019single}, we report accuracy/Acc (\% of predicted points within $t$ of the ground-truth), completeness/Cmp (\% of ground-truth points within $t$ of the prediction), and their harmonic mean, F1-score. This gives a overall summary of scene-level reconstruction.

\parnoit{{Rays (Acc/Cmp/F1), Occluded Points}} We also evaluate reconstruction performance along each ray independently, measuring Acc/Cmp/F1 on each ray and reporting the mean. The paper shows results for occluded points, defined as all surfaces past the first intersection; the supplement contains full results. Evaluating each ray independently is a more stringent test for occluded surfaces than a scene metric: with scene-level evaluation on a image, a prediction can miss a surface (e.g., the 2nd intersection) on every other pixel. These missing predictions will be covered for by hidden surfaces on adjacent rays. Ray-based evaluation, however, requires each pixel to have all surfaces present to receive full credit.

\begin{figure}[t]
    \centering
    \noindent\includegraphics[width=0.49\textwidth]{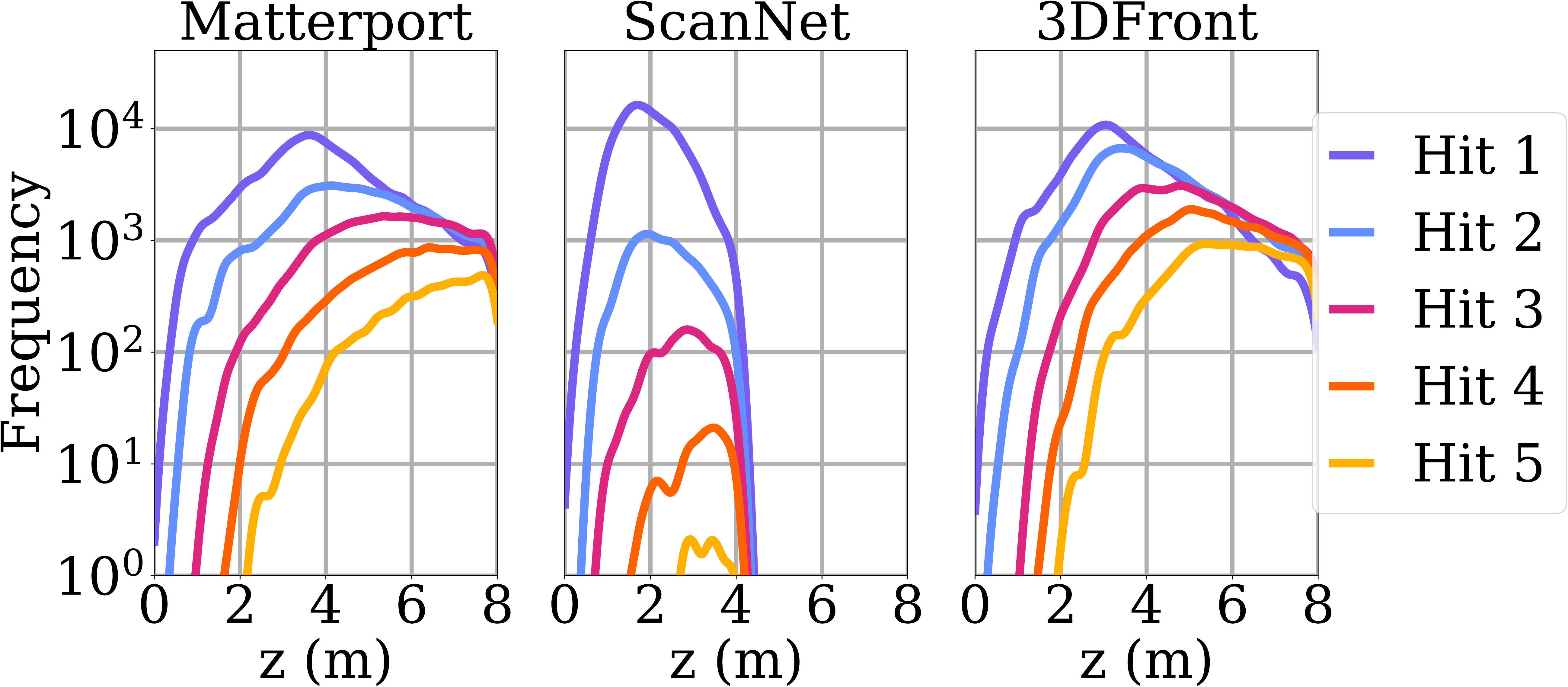} 
    \captionof{figure}{{\bf Ray hit count distribution}. We compare the distribution over surface hit (intersection) locations for first 4 hits over 1M rays. ScanNet has $\le$ 1\% rays as compared to Matterport and 3DFront which have $\ge$ 25\% rays with more than 2 hits}
    \label{fig:dataset_hits}
    \vspace{-4mm}
\end{figure}

\begin{figure*}[t]
    \centering  
    \noindent\includegraphics[width=\textwidth]{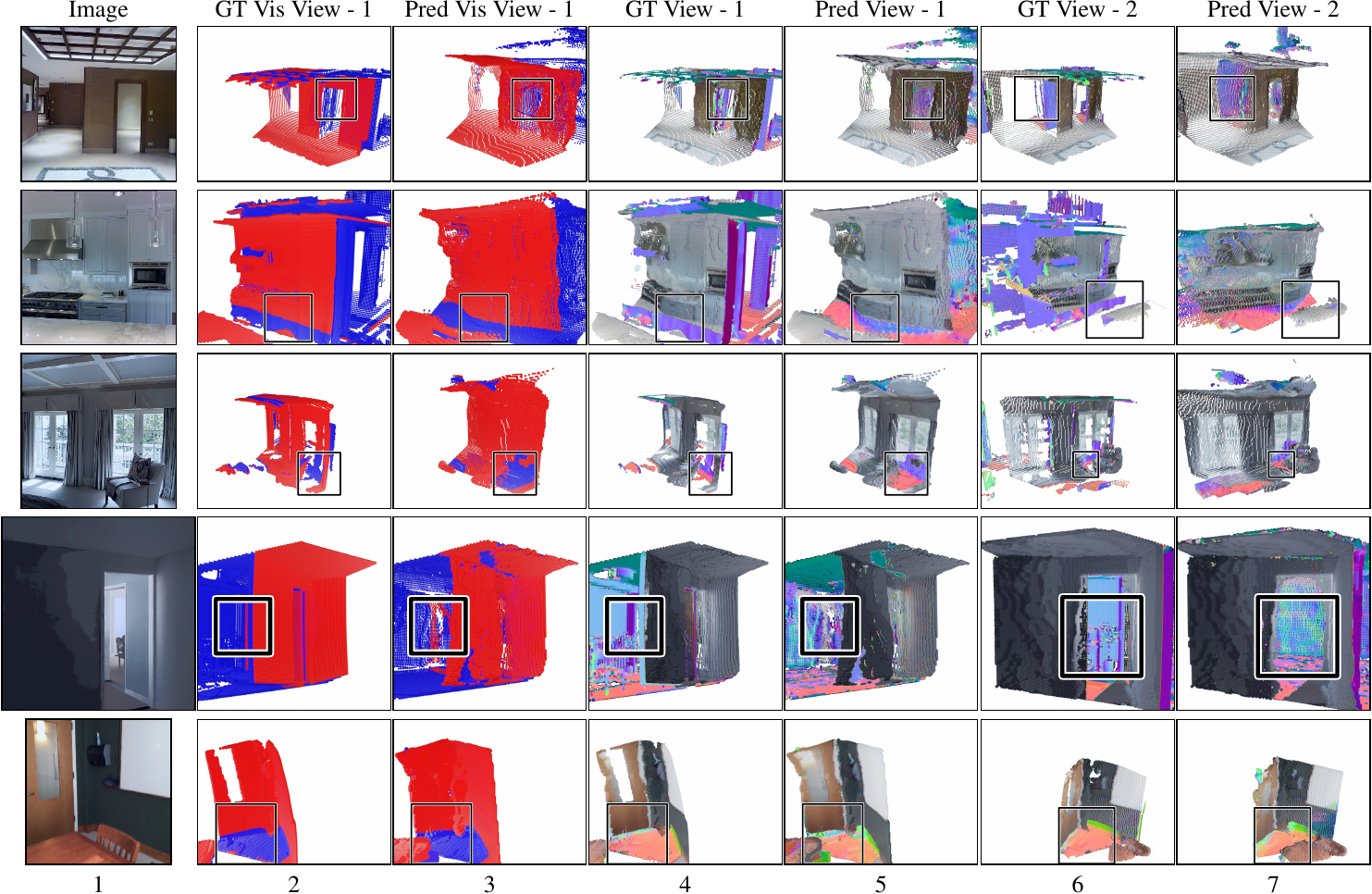} 
    \caption{{\bf Novel views from DRDF} 
Outputs from DRDF and ground-truth from new viewpoints.
Columns 2,3 show 
\textcolor{red}{visible points in red} and \textcolor{blue}{occluded points in blue}. Other columns, show the visible regions with the image and occluded regions with computed surface normals (\protect\includegraphics[height=8pt,width=8pt]{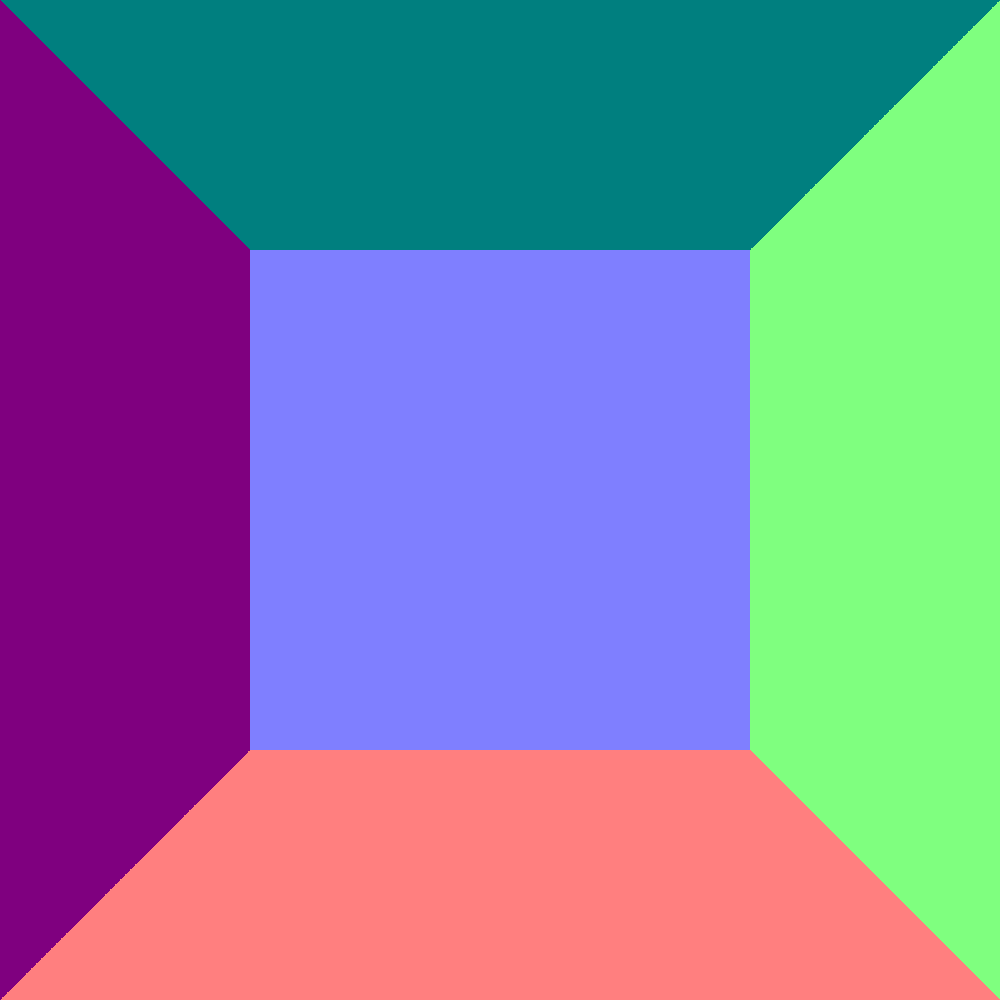}, scheme from camera inside a cube). \drdf recovers occluded regions, such as a room behind the door (row 1 \& 4), a floor behind the kitchen counter (row 2), and a wall and floor behind the chair/couch (row 3 \& 5). Rows 1-3: Matterport3D; 4: 3DFront; 5: ScanNet. 
    }
    \label{fig:qual_novel}
\end{figure*}

\parnobf{Datasets} We see three key properties for datasets: the images should be real to avoid networks using rendering artifacts; the mesh should be a real capture since imitating capture holes is a research problem; and there should be lots of occluded regions. Our main dataset is \matterport, which satisfies all properties.

We also evaluate on \tdf and \scannet. While 3DFront has no capture holes, cutting it with a view frustum creates holes. \scannet is a popular in 3D reconstruction, but has far less occluded geometry compared to the other datasets. A full description of the datasets appears in the supplement.

\parnoit{\matterport} We use the {\it raw images} captured with the Matterport camera. We split the 90 scenes into train/val/test (60/15/15) and remove images that are too close to the mesh (${\ge}60\%$ of image within 1m) or are ${>}20^\circ$ away from level. We then sample 13K/1K/1K images for train/val/test set.

\parnoit{\tdf} This is a synthetic dataset of houses created by artists with a hole-free 3D geometry. We collect 4K scenes from \tdf after removing scenes with missing annotations. We select 20 camera poses and filter for bad camera poses similar to \matterport. Our train set has 3K scenes with approximately 47K images. Val/Test sets have $500$ scenes with 1K images each.

\parnoit{\scannet} We use splits from \cite{dai2017scannet} (1045/156/312 train/val/test scenes) and randomly select 5 images per scene for train, and 10 images per scene for val/test. We then sample to a set of 33K/1K/1K images per train/val/test.

\parnoit{Dataset Scene Statistics} 
To give a sense of scene statistics, we plot the frequency of the locations of the first 5 ray hits (intersections) for each dataset (computed on 1M rays each) in Fig. \ref{fig:dataset_hits}. We show $99\%$ of ScanNet rays have 1 or 2 hits, while ${\ge}24\%$ of \matterport and \tdf rays have more than 2 hits.

\subsection{Baselines}
\label{sec:experiments_baselines}
We compare against baselines to test our contributions. 
For fair comparison, all approaches use the same ResNet-34~\cite{he2016deep} backbone and the same MLP. We extract features from multiple layers via bilinear interpolation~\cite{yu2020pixelnerf}. Thus, different distance functions are trained identically by replacing the target distance. Each method's description consists of two parts: a prediction space parameterization and a decoding strategy to convert inferred distances to surfaces.

\parnobf{Picking decoding strategies}
Most baselines predict a distance function rather than a set of intersections and need a decoding strategy to convert distances to a set of surface locations. Some baselines have trivial strategies (e.g., zero-crossings); others are more sensitive and have parameters.
We tried multiple strategies for each baseline based on past work and theoretical analysis of their behavior. We report the best one by Scene F1 on \matterport. When there are parameters, we tune them to ensure similar completeness to our method. Accuracy and completeness have a trade-off; fixing one ensures that methods are compared at similar operating points, making F1 score meaningful.

\begin{figure*}[t]
    \centering
    \noindent\includegraphics[width=\textwidth]{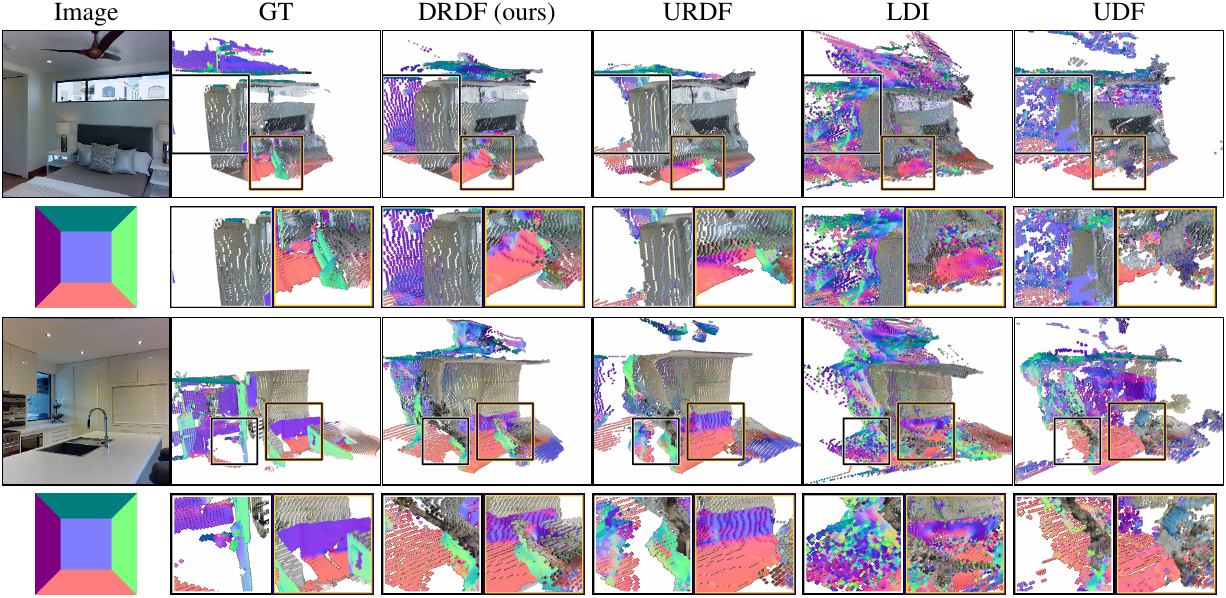} 
   \caption{ {\bf Qualitative Comparison with Baselines} 
We render the generated 3D outputs in a new view (rotated $\leftarrow$) with 2 crops for better visual comparison. Visible regions show the image; occluded regions show surface normals (legend shows a camera in a cube). \tsdf produces higher quality results compared to \mpd and \scened (row 1, 2, more consistency, smoother surface, no blobs). \udf misses parts of the floor (row 1/crop 2) and the green colored side of the kitchen counter (row2/crop 2). See supp. for more results.}
    \label{fig:qual_comp}
\end{figure*}

\parnobf{Layered Depth Images (\mpd)} To test the value of framing the problem as implicit function prediction, we train a method to predict a $k$-channel depthmap where the $i$\textsuperscript{th} output predicts the $i$\textsuperscript{th} intersection along the pixels. We use a L1 loss per pixel and intersection. 
We set $k=4$, the same number of intersections the proposed approach uses.\\
{\it Decoding.} LDI directly predicts surface locations. 

\parnobf{Layered Depth Images  with Confidence (LDI + C)} We augment the the LDI baseline with $k$ additional channels that represent the likelihood the $i$\textsuperscript{th} intersection exists. These added channels are trained with binary cross-entropy.\\
{\it Decoding}. We accept  pixel layers with likelihood $\ge 0.5$.

\parnobf{Unsigned Distance Function (\scened) \cite{chibane2020ndf}} Chibane \etal ~\cite{chibane2020ndf} fit \scened to a single 3D scene. We predict it from images.\\
{\it Decoding.} We use \texttt{scipy.argrelextrema} ~\cite{2020SciPy-NMeth} to find local extrema. We find local minima of the distance function within a 1m window along the ray. We found this works better than absolute thresholding (by 14.7 on F1). Sphere tracing and gradient-based optimization proposed by ~\cite{chibane2020ndf} performs substantially worse (25.7 on F1), likely since it assumes the predicted UDF behaves similar to a  GT UDF.

\parnobf{Unsigned Ray Distance Function (\udf)} Inspired from Chibane \etal\cite{chibane2020ndf} we compare \scened against its ray based version(\udf). Now, for direct comparison between ray distance functions we compare \urdf against \drdf.\\
{\it Decoding}. We do NMS on thresholded data with connected components on the ray with predicted distance below a tuned constant $\tau$, and keep the first prediction. This outperforms: thresholding (by 5.3 on F1); finding 0-crossings of the numerical gradient (by 11 on F1); and sphere tracing and optimization ~\cite{chibane2020ndf} (by 6.6 on F1).
\parnobf{Ray Sign-Agnostic Learning Loss (\sal) \cite{atzmon2020sal}} Traditional SDF learning is impossible due to the non-watertightness of the data and so we use the sign agnostic approach proposed by \cite{atzmon2020sal}. We initialize our architecture with the SAL initialization and train with the SAL loss. The SAL approach assumes that while the data may not be watertight due to noisy capture, the underlying model is watertight. In this case, rays start and end {\it outside} objects (and thus the number of hits along each ray is even). This is not necessarily the case on \matterport and \tdf.\\
{\it Decoding.} Following \cite{atzmon2020sal}, we find surfaces as zero-crossings of the predicted distance function along the ray.

\parnobf{Ray Occupancy (\occ)}  Traditional interior/exterior occupancy is not feasible on non-watertight data, but one can predict whether a point is within $r$ of a surface as a classification problem. This baseline tests the value of predicting ray distances, and not just occupancy. We tried several values of $r$ ([$0.1$, $0.25$, $0.5$, $1$]m) and report the best version.\\ 
{\it Decoding.} Each surface, in theory, produces two locations with probability $0.5$: an onset and offset crossing. Finding all $0.5$-crossings leads to doubled predictions. Instead, we consider all adjacent and nearby pairs of offsets and onsets, and average them; unpaired crossings are kept. This outperforms keeping just a single $0.5$-crossing (by 4.7 on F1). 

\begin{table}[t]
\setlength{\tabcolsep}{3pt} 

\centering 
\caption{{\bf Scene Acc/Comp/F1Score}. Thresholds: 0.5m (MP3D~\cite{chang2017matterport3d}, \tdf), 0.2m (\scannet). {\bf Bold is best}, \underline{underline is 2\textsuperscript{nd} best} per column. \tsdf is best in F1 and accuracy, and always comparable to the best in completeness.}
\vspace{2mm}
\label{tab:scene}
\begin{adjustbox}{max width=\linewidth}
\begin{tabular}{lm@{~}m@{~}mt@{~}t@{~}ts@{~}s@{~}s} \toprule
& \multicolumn{3}{m}{MP3D \cite{chang2017matterport3d}} & \multicolumn{3}{t}{\tdf } & \multicolumn{3}{s}{\scannet}  \\ 
Method   & Acc & Cmp & F1 & Acc  & Cmp & F1 & Acc & Cmp & F1 \\ 
\cmidrule(r){1-1}\cmidrule(r){2-4} \cmidrule(r){5-7} \cmidrule(r){8-10}
  \mpd  & 66.2 & \underline{72.4} & 67.4 & 68.6 & 46.5 & 52.7 & 19.3 & 28.6 & 21.5 \\ 
 \mpd+C  & 64.8 & 55.1 & 57.7 & 70.8 & 45.1 & 52.4 & 19.9 & 32.0 & 23.3 \\ 
 \sal \cite{atzmon2020sal}  & 66.1 & 25.5 & 34.3 & 80.7 & 28.5 & 39.5 & 51.2 & \bf 70.0 & 57.7 \\ 
 \scened  & 58.7 & \bf 76.0 & 64.7 & 70.1 & \underline{51.9} & 57.4 & 44.4 & {62.6} & 50.8 \\ 
 \occ  & 73.4 & 69.4 & \underline{69.6} & \underline{86.4} & 48.1 & \underline{59.6} & 51.5 & 58.5 & 53.7 \\ 
 \urdf  & \underline{74.5} & 67.1 & 68.7 & 85.0 & 47.7 & 58.7 & \underline{61.0} & 57.8 & \underline{58.2} \\ 
 \tsdf (ours)  & \bf 75.4 & 72.0 & \bf 71.9 & \bf 87.3 & \bf 52.6 & \bf 63.4 & \bf 62.0 & \underline {62.7} & \bf 60.9 \\ 
\bottomrule
\end{tabular}
\end{adjustbox}

\end{table}

\begin{table}[t]
\setlength{\tabcolsep}{3pt} 
\centering 
\caption{{\bf Ray Acc/Comp/F1Score on Occluded Points}. Thresholds: 0.5m (MP3D~\cite{chang2017matterport3d}, 3DFront~\cite{fu20203dfront}), 0.2m (ScanNet~\cite{dai2017scannet}). 
\tsdf is best on F1 and Acc, and is occasionally 2\textsuperscript{nd} best on Cmp. Gains on occluded points are even larger than the full scene.}
\vspace{2mm}
\label{tab:ray}
\begin{adjustbox}{max width=\linewidth}
\begin{tabular}{lm@{~}m@{~}mt@{~}t@{~}ts@{~}s@{~}s} \toprule
& \multicolumn{3}{m}{MP3D \cite{chang2017matterport3d}} & \multicolumn{3}{t}{\tdf } & \multicolumn{3}{s}{\scannet}  \\ 
Method   & Acc & Cmp & F1 & Acc  & Cmp & F1 & Acc & Cmp & F1 \\ 
\cmidrule(r){1-1}\cmidrule(r){2-4} \cmidrule(r){5-7} \cmidrule(r){8-10}
  \mpd  & 13.9 & \bf 42.8 & 19.3 & 17.8 & \underline{35.8} & 22.2 & 0.5 & 9.0 & 2.4 \\ 
 \mpd+C  & 18.7 & 21.7 & 19.3 & 17.7 & 22.6 & 19.9 & 1.1 & 2.4 & 3.5 \\ 
 \sal \cite{atzmon2020sal}  & 5.5 & 0.5 & 3.5 & 24.1 & 4.3 & 11.4 & 2.4 & \bf 38.7 & 5.6 \\ 
 \scened  & 15.5 & 23.0 & 16.6 & 29.3 & 21.3 & 23.4 & 1.8 & 7.8 & 5.5 \\ 
 \occ  & \underline{26.2} & 20.5 & \underline{21.6} & 53.2 & 22.0 & \underline{31.0} & 6.6 & 12.3 & 11.4 \\ 
 \urdf  & 24.9 & 20.6 & 20.7 & \underline{47.7} & 23.3 & 30.2 & \underline{8.4} & 11.6 & \underline{13.8} \\ 
 \tsdf (ours)  & \bf 28.4 & \underline{30.0} & \bf 27.3 & \bf 54.6 & \bf 56.0 & \bf 52.6 & \bf 9.0 & \underline{20.4} & \bf 16.0 \\ 
\bottomrule
\end{tabular}
\end{adjustbox}
\vspace{-5mm}
\end{table}

\subsection{Results}

\label{sec:results}
\parnobf{Qualitative Results} Qualitative results of our method appear throughout the paper (by itself in Fig.~\ref{fig:qual_novel} and compared to baselines in Fig.~\ref{fig:qual_comp}). Our approach is is often able to generate parts of the occluded scene, such as a room behind a door, cabinets and floor behind kitchen counters, and occluded regions behind furniture. Sometimes the method completes holes in the ground-truth that are due to scanning error. On the other hand we see our method sometimes fails to characterize the missing parts as detailed occluded 3D \eg plants. Compared to baselines, our approach does qualitatively better. LDI and UDF often have floating blobs or extruded regions, due to either predicting too many layers (LDI) or having a distance function that is challenging to predict (UDF). URDF produces qualitatively better results, but often misses points in occluded regions. 

\begin{figure}[t]
    \centering
    \includegraphics[width=0.48\textwidth]{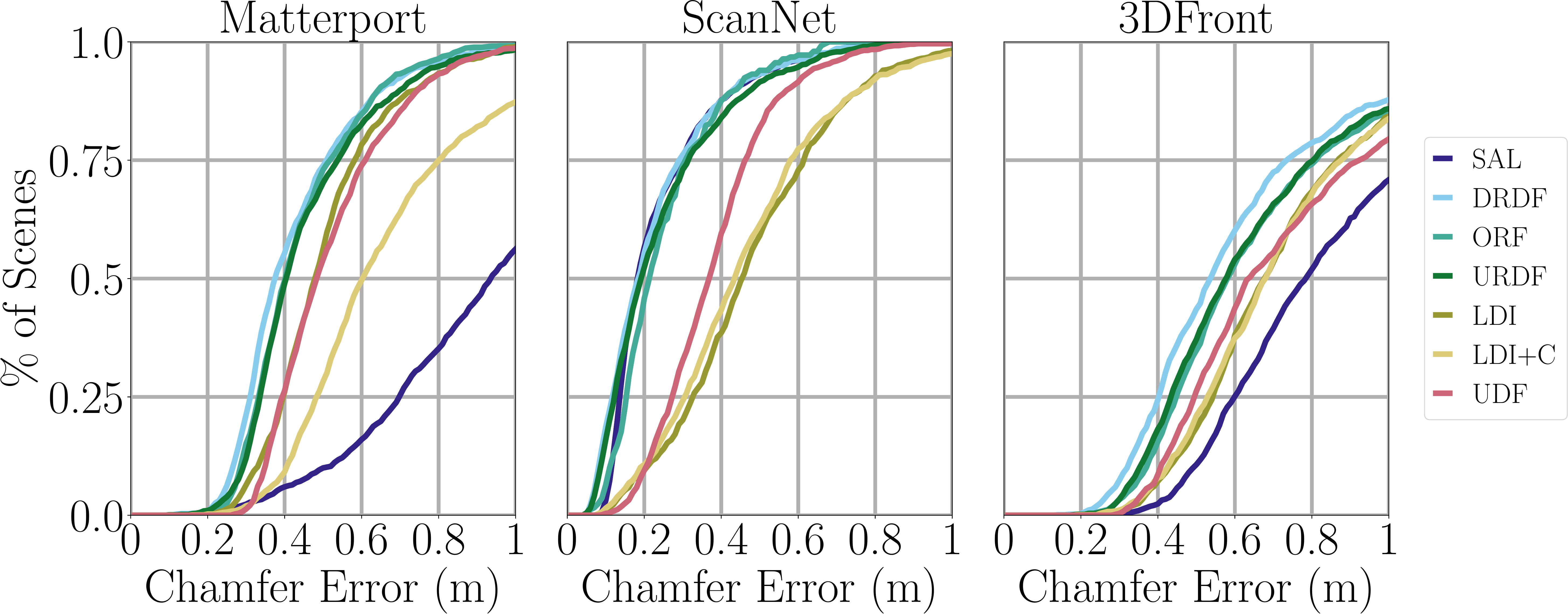} 
    \caption{{\bf Chamfer L1}: \% of scenes on Y-axis as a function of symmetric Chamfer L1 error $\le t$ on X-axis.  \textcolor{ibm6}{\tsdf} is better on Matterport and 3DFront, and comparable to the best other method on ScanNet.} 
    \label{fig:chamfer}
\end{figure}

\parnobf{Quantitative Results} These results are borne out in the quantitative results. 
\figref{fig:chamfer} shows the Chamfer plot, Table.~\ref{tab:scene} reports the scene based metrics and Table.~\ref{tab:ray} occluded surfaces metrics along rays. DRDF consistently does at least as well, or substantially better than the baselines on Chamfer. In general, DRDF does substantially better than all baselines. In a few situations, a baseline beats DRDF in completeness at the cost of substantially worse accuracy. However, a single baseline is not competitive with DRDF {\it across} datasets: SAL works well on ScanNet~\cite{dai2017scannet} and LDI works well on \matterport~\cite{chang2017matterport3d}.

\mpd performs worse than \drdf because it cannot vary its number of intersections; simply adding a second stack of outputs (\mpd + C.) is insufficient. This is because \drdf can learn {\it where} things tend to be, while LDI-based methods have to learn the order in which things occur (e.g., is the floor 2nd or the 3rd intersection at a pixel?). \sal performs competitively on ScanNet, likely because of the relatively limited variability in numbers of intersections per ray; when tested on Matterport3D and 3DFront, its performance drops substantially. 
 
We compare against a Monocular Depth Estimation (MDE) baseline with a pre-trained MiDaS~\cite{Ranftl2020} model. It has been trained on more datasets and has an optimal scale and translation fit per-image (which our models do not get). As it predicts one intersection its F1 is lower, $57.2$ \vs $71.9$ for \drdf on \matterport. Nonetheless, we see advances in MDE complementary to advances in DRDF.

The most straightforward way to learn on non-watertight data is to predict unsigned scene distances~\cite{chibane2020ndf} which has been shown to work with memorizing 3D scenes.
However, predicting it from a single image is a different problem entirely, and scene distances require integration of information over large areas. This leads to poor performance. Predicting distances on rays alleviates this challenge, but recovering intersections remains hard even with multiple decoding strategies. Thus, \tsdf outperforms \urdf. \occ similarly requires decoding strategies and is sensitive to training parameters. In contrast, by accounting for the uncertainty in surface location, \drdf requires a simple decoding strategy and outperforms other methods.

\parnobf{Conclusions} This paper introduced a new distance function, DRDF, for 3D reconstruction from an unseen image.  We use real 3D, non-watertight data at training. We showed that DRDF does not suffer from pitfalls of other distance functions and outperforms other conventional methods. DRDF achieves substantially better qualitative results and has a simple {\it decoding strategy} to recover intersections -- thanks to its stable behavior near intersections. \drdf's progress in learning 3D from real data is extendable to learning from multi-view data. Our approach, however, has societal limitations as our data does not reflect most peoples' reality: Matterport3D for instance, has many lavish houses and this may widen the technological gap. However, we are optimistic that our system will enable learning from scans collected by ordinary people rather than experts.

\parnobf{Acknowledgments}
We would like the thank Alexandar Raistrick and Chris Rockwell for their help with the 3DFront dataset. We like to thank Shubham Tulsiani, Richard Higgins, Sarah Jabour, Shengyi Qian, Linyi Jin, Karan Desai, Mohammed El Banani, Chris Rockwell, Alexandar Raistrick, Dandan Shan, Andrew Owens for comments on the draft versions of this paper. NK was supported by TRI. Toyota Research Institute (''TRI'') provided funds to assist the authors with their research but this article solely reflects the opinions and conclusions of its authors and not TRI or any other Toyota entity. 
{\small
\bibliographystyle{ieee_fullname}
\bibliography{egbib}

\begin{thebibliography}{10}\itemsep=-1pt

\bibitem{atzmon2020sal}
Matan Atzmon and Yaron Lipman.
\newblock Sal: Sign agnostic learning of shapes from raw data.
\newblock In {\em Proceedings of the IEEE/CVF Conference on Computer Vision and
  Pattern Recognition}, pages 2565--2574, 2020.

\bibitem{atzmon2020sal++}
Matan Atzmon and Yaron Lipman.
\newblock Sal++: Sign agnostic learning with derivatives.
\newblock {\em arXiv preprint arXiv:2006.05400}, 2020.

\bibitem{bae2021estimating}
Gwangbin Bae, Ignas Budvytis, and Roberto Cipolla.
\newblock Estimating and exploiting the aleatoric uncertainty in surface normal
  estimation.
\newblock In {\em Proceedings of the IEEE/CVF International Conference on
  Computer Vision}, pages 13137--13146, 2021.

\bibitem{barrow1978recovering}
Harry Barrow, J Tenenbaum, A Hanson, and E Riseman.
\newblock Recovering intrinsic scene characteristics.
\newblock {\em Comput. Vis. Syst}, 2(3-26):2, 1978.

\bibitem{chang2017matterport3d}
Angel Chang, Angela Dai, Thomas Funkhouser, Maciej Halber, Matthias Niessner,
  Manolis Savva, Shuran Song, Andy Zeng, and Yinda Zhang.
\newblock Matterport3d: Learning from rgb-d data in indoor environments.
\newblock {\em arXiv preprint arXiv:1709.06158}, 2017.

\bibitem{chang2015shapenet}
Angel~X Chang, Thomas Funkhouser, Leonidas Guibas, Pat Hanrahan, Qixing Huang,
  Zimo Li, Silvio Savarese, Manolis Savva, Shuran Song, Hao Su, et~al.
\newblock Shapenet: An information-rich 3d model repository.
\newblock {\em arXiv preprint arXiv:1512.03012}, 2015.

\bibitem{Chen_2021_CVPR}
Kefan Chen, Noah Snavely, and Ameesh Makadia.
\newblock Wide-baseline relative camera pose estimation with directional
  learning.
\newblock In {\em Proceedings of the IEEE/CVF Conference on Computer Vision and
  Pattern Recognition (CVPR)}, pages 3258--3268, June 2021.

\bibitem{chen2019learning}
Zhiqin Chen and Hao Zhang.
\newblock Learning implicit fields for generative shape modeling.
\newblock In {\em Proceedings of the IEEE/CVF Conference on Computer Vision and
  Pattern Recognition}, pages 5939--5948, 2019.

\bibitem{chibane2020ndf}
Julian Chibane, Aymen Mir, and Gerard Pons-Moll.
\newblock Neural unsigned distance fields for implicit function learning.
\newblock In {\em Advances in Neural Information Processing Systems
  ({NeurIPS})}, December 2020.

\bibitem{choy20163d}
Christopher~B Choy, Danfei Xu, JunYoung Gwak, Kevin Chen, and Silvio Savarese.
\newblock 3d-r2n2: A unified approach for single and multi-view 3d object
  reconstruction.
\newblock In {\em European conference on computer vision}, pages 628--644.
  Springer, 2016.

\bibitem{dai2017scannet}
Angela Dai, Angel~X Chang, Manolis Savva, Maciej Halber, Thomas Funkhouser, and
  Matthias Nie{\ss}ner.
\newblock Scannet: Richly-annotated 3d reconstructions of indoor scenes.
\newblock In {\em Proceedings of the IEEE Conference on Computer Vision and
  Pattern Recognition}, pages 5828--5839, 2017.

\bibitem{dai2020sg}
Angela Dai, Christian Diller, and Matthias Nie{\ss}ner.
\newblock Sg-nn: Sparse generative neural networks for self-supervised scene
  completion of rgb-d scans.
\newblock In {\em Proceedings of the IEEE/CVF Conference on Computer Vision and
  Pattern Recognition}, pages 849--858, 2020.

\bibitem{trimesh}
Michael Dawson-Haggerty.
\newblock {Trimesh [Computer Software]}, 2020.

\bibitem{Pero_2013_CVPR}
Luca Del~Pero, Joshua Bowdish, Bonnie Kermgard, Emily Hartley, and Kobus
  Barnard.
\newblock Understanding bayesian rooms using composite 3d object models.
\newblock In {\em Proceedings of the IEEE Conference on Computer Vision and
  Pattern Recognition (CVPR)}, June 2013.

\bibitem{dhamo2019object}
Helisa Dhamo, Nassir Navab, and Federico Tombari.
\newblock Object-driven multi-layer scene decomposition from a single image.
\newblock In {\em Proceedings of the IEEE/CVF International Conference on
  Computer Vision}, pages 5369--5378, 2019.

\bibitem{fan2017point}
Haoqiang Fan, Hao Su, and Leonidas~J Guibas.
\newblock A point set generation network for 3d object reconstruction from a
  single image.
\newblock In {\em Proceedings of the IEEE conference on computer vision and
  pattern recognition}, pages 605--613, 2017.

\bibitem{fidler20123d}
Sanja Fidler, Sven Dickinson, and Raquel Urtasun.
\newblock 3d object detection and viewpoint estimation with a deformable 3d
  cuboid model.
\newblock In {\em Advances in neural information processing systems}, pages
  611--619, 2012.

\bibitem{fouhey2013data}
David~F Fouhey, Abhinav Gupta, and Martial Hebert.
\newblock Data-driven 3d primitives for single image understanding.
\newblock In {\em Proceedings of the IEEE International Conference on Computer
  Vision}, pages 3392--3399, 2013.

\bibitem{fu20203dfront}
Huan Fu, Bowen Cai, Lin Gao, Lingxiao Zhang, Cao Li, Qixun Zeng, Chengyue Sun,
  Yiyun Fei, Yu Zheng, Ying Li, Yi Liu, Peng Liu, Lin Ma, Le Weng, Xiaohang Hu,
  Xin Ma, Qian Qian, Rongfei Jia, Binqiang Zhao, and Hao Zhang.
\newblock 3d-front: 3d furnished rooms with layouts and semantics.
\newblock {\em arXiv preprint arXiv:2011.09127}, 2020.

\bibitem{girdhar2016learning}
Rohit Girdhar, David~F Fouhey, Mikel Rodriguez, and Abhinav Gupta.
\newblock Learning a predictable and generative vector representation for
  objects.
\newblock In {\em European Conference on Computer Vision}, pages 484--499.
  Springer, 2016.

\bibitem{gkioxari2019mesh}
Georgia Gkioxari, Jitendra Malik, and Justin Johnson.
\newblock Mesh r-cnn.
\newblock In {\em Proceedings of the IEEE International Conference on Computer
  Vision}, pages 9785--9795, 2019.

\bibitem{groueix2018papier}
Thibault Groueix, Matthew Fisher, Vladimir~G Kim, Bryan~C Russell, and Mathieu
  Aubry.
\newblock A papier-m{\^a}ch{\'e} approach to learning 3d surface generation.
\newblock In {\em Proceedings of the IEEE conference on computer vision and
  pattern recognition}, pages 216--224, 2018.

\bibitem{hane2017hierarchical}
Christian H{\"a}ne, Shubham Tulsiani, and Jitendra Malik.
\newblock Hierarchical surface prediction for 3d object reconstruction.
\newblock In {\em 2017 International Conference on 3D Vision (3DV)}, pages
  412--420. IEEE, 2017.

\bibitem{he2016deep}
Kaiming He, Xiangyu Zhang, Shaoqing Ren, and Jian Sun.
\newblock Deep residual learning for image recognition.
\newblock In {\em Proceedings of the IEEE conference on computer vision and
  pattern recognition}, pages 770--778, 2016.

\bibitem{hedau2009recovering}
Varsha Hedau, Derek Hoiem, and David Forsyth.
\newblock Recovering the spatial layout of cluttered rooms.
\newblock In {\em 2009 IEEE 12th international conference on computer vision},
  pages 1849--1856. IEEE, 2009.

\bibitem{hoiem2005geometric}
Derek Hoiem, Alexei~A Efros, and Martial Hebert.
\newblock Geometric context from a single image.
\newblock In {\em Tenth IEEE International Conference on Computer Vision
  (ICCV'05) Volume 1}, volume~1, pages 654--661. IEEE, 2005.

\bibitem{pix2pix2017}
Phillip Isola, Jun-Yan Zhu, Tinghui Zhou, and Alexei~A Efros.
\newblock Image-to-image translation with conditional adversarial networks.
\newblock {\em CVPR}, 2017.

\bibitem{issaranon2019counterfactual}
Theerasit Issaranon, Chuhang Zou, and David Forsyth.
\newblock Counterfactual depth from a single rgb image.
\newblock In {\em Proceedings of the IEEE/CVF International Conference on
  Computer Vision Workshops}, pages 0--0, 2019.

\bibitem{izadinia2017im2cad}
Hamid Izadinia, Qi Shan, and Steven~M Seitz.
\newblock Im2cad.
\newblock In {\em Proceedings of the IEEE Conference on Computer Vision and
  Pattern Recognition}, pages 5134--5143, 2017.

\bibitem{jiang2020peek}
Ziyu Jiang, Buyu Liu, Samuel Schulter, Zhangyang Wang, and Manmohan Chandraker.
\newblock Peek-a-boo: Occlusion reasoning in indoor scenes with plane
  representations.
\newblock In {\em Proceedings of the IEEE/CVF Conference on Computer Vision and
  Pattern Recognition}, pages 113--121, 2020.

\bibitem{jin2021planar}
Linyi Jin, Shengyi Qian, Andrew Owens, and David~F Fouhey.
\newblock Planar surface reconstruction from sparse views.
\newblock {\em International Conference on Computer Vision (ICCV)}, 2021.

\bibitem{kendall2017uncertainties}
Alex Kendall and Yarin Gal.
\newblock What uncertainties do we need in bayesian deep learning for computer
  vision?
\newblock In {\em Proceedings of the 31st International Conference on Neural
  Information Processing Systems}, pages 5580--5590, 2017.

\bibitem{kingma2014adam}
Diederik~P Kingma and Jimmy Ba.
\newblock Adam: A method for stochastic optimization.
\newblock {\em arXiv preprint arXiv:1412.6980}, 2014.

\bibitem{ku2019monocular}
Jason Ku, Alex~D Pon, and Steven~L Waslander.
\newblock Monocular 3d object detection leveraging accurate proposals and shape
  reconstruction.
\newblock In {\em Proceedings of the IEEE/CVF conference on computer vision and
  pattern recognition}, pages 11867--11876, 2019.

\bibitem{kulkarni2019canonical}
Nilesh Kulkarni, Abhinav Gupta, and Shubham Tulsiani.
\newblock Canonical surface mapping via geometric cycle consistency.
\newblock In {\em Proceedings of the IEEE/CVF International Conference on
  Computer Vision}, pages 2202--2211, 2019.

\bibitem{kulkarni20193d}
Nilesh Kulkarni, Ishan Misra, Shubham Tulsiani, and Abhinav Gupta.
\newblock 3d-relnet: Joint object and relational network for 3d prediction.
\newblock In {\em Proceedings of the IEEE International Conference on Computer
  Vision}, pages 2212--2221, 2019.

\bibitem{LiSilhouette19}
Lin Li, Salman Khan, and Nick Barnes.
\newblock Silhouette-assisted 3d object instance reconstruction from a
  cluttered scene.
\newblock In {\em ICCV Workshops}, 2019.

\bibitem{lin2017learning}
Chen-Hsuan Lin, Chen Kong, and Simon Lucey.
\newblock Learning efficient point cloud generation for dense 3d object
  reconstruction.
\newblock {\em arXiv preprint arXiv:1706.07036}, 2017.

\bibitem{loshchilov2017decoupled}
Ilya Loshchilov and Frank Hutter.
\newblock Decoupled weight decay regularization.
\newblock {\em arXiv preprint arXiv:1711.05101}, 2017.

\bibitem{martin2020nerf}
Ricardo Martin-Brualla, Noha Radwan, Mehdi~SM Sajjadi, Jonathan~T Barron,
  Alexey Dosovitskiy, and Daniel Duckworth.
\newblock Nerf in the wild: Neural radiance fields for unconstrained photo
  collections.
\newblock {\em arXiv preprint arXiv:2008.02268}, 2020.

\bibitem{mescheder2019occupancy}
Lars Mescheder, Michael Oechsle, Michael Niemeyer, Sebastian Nowozin, and
  Andreas Geiger.
\newblock Occupancy networks: Learning 3d reconstruction in function space.
\newblock In {\em Proceedings of the IEEE Conference on Computer Vision and
  Pattern Recognition}, pages 4460--4470, 2019.

\bibitem{mildenhall2020nerf}
Ben Mildenhall, Pratul~P Srinivasan, Matthew Tancik, Jonathan~T Barron, Ravi
  Ramamoorthi, and Ren Ng.
\newblock Nerf: Representing scenes as neural radiance fields for view
  synthesis.
\newblock {\em arXiv preprint arXiv:2003.08934}, 2020.

\bibitem{mousavian20173d}
Arsalan Mousavian, Dragomir Anguelov, John Flynn, and Jana Kosecka.
\newblock 3d bounding box estimation using deep learning and geometry.
\newblock In {\em Proceedings of the IEEE conference on Computer Vision and
  Pattern Recognition}, pages 7074--7082, 2017.

\bibitem{newbold1983arima}
Paul Newbold.
\newblock Arima model building and the time series analysis approach to
  forecasting.
\newblock {\em Journal of forecasting}, 2(1):23--35, 1983.

\bibitem{nie2020total3dunderstanding}
Yinyu Nie, Xiaoguang Han, Shihui Guo, Yujian Zheng, Jian Chang, and Jian~Jun
  Zhang.
\newblock Total3dunderstanding: Joint layout, object pose and mesh
  reconstruction for indoor scenes from a single image.
\newblock In {\em Proceedings of the IEEE/CVF Conference on Computer Vision and
  Pattern Recognition}, pages 55--64, 2020.

\bibitem{park2019deepsdf}
Jeong~Joon Park, Peter Florence, Julian Straub, Richard Newcombe, and Steven
  Lovegrove.
\newblock Deepsdf: Learning continuous signed distance functions for shape
  representation.
\newblock In {\em Proceedings of the IEEE Conference on Computer Vision and
  Pattern Recognition}, pages 165--174, 2019.

\bibitem{poggi2020uncertainty}
Matteo Poggi, Filippo Aleotti, Fabio Tosi, and Stefano Mattoccia.
\newblock On the uncertainty of self-supervised monocular depth estimation.
\newblock In {\em Proceedings of the IEEE/CVF Conference on Computer Vision and
  Pattern Recognition}, pages 3227--3237, 2020.

\bibitem{Ranftl2020}
Ren\'{e} Ranftl, Katrin Lasinger, David Hafner, Konrad Schindler, and Vladlen
  Koltun.
\newblock Towards robust monocular depth estimation: Mixing datasets for
  zero-shot cross-dataset transfer.
\newblock {\em IEEE Transactions on Pattern Analysis and Machine Intelligence
  (TPAMI)}, 2020.

\bibitem{saito2019pifu}
Shunsuke Saito, Zeng Huang, Ryota Natsume, Shigeo Morishima, Angjoo Kanazawa,
  and Hao Li.
\newblock Pifu: Pixel-aligned implicit function for high-resolution clothed
  human digitization.
\newblock In {\em Proceedings of the IEEE International Conference on Computer
  Vision}, pages 2304--2314, 2019.

\bibitem{saito2020pifuhd}
Shunsuke Saito, Tomas Simon, Jason Saragih, and Hanbyul Joo.
\newblock Pifuhd: Multi-level pixel-aligned implicit function for
  high-resolution 3d human digitization.
\newblock In {\em Proceedings of the IEEE/CVF Conference on Computer Vision and
  Pattern Recognition}, pages 84--93, 2020.

\bibitem{saxena2008make3d}
Ashutosh Saxena, Min Sun, and Andrew~Y Ng.
\newblock Make3d: Learning 3d scene structure from a single still image.
\newblock {\em IEEE transactions on pattern analysis and machine intelligence},
  31(5):824--840, 2008.

\bibitem{seitz2006comparison}
Steven~M Seitz, Brian Curless, James Diebel, Daniel Scharstein, and Richard
  Szeliski.
\newblock A comparison and evaluation of multi-view stereo reconstruction
  algorithms.
\newblock In {\em 2006 IEEE computer society conference on computer vision and
  pattern recognition (CVPR'06)}, volume~1, pages 519--528. IEEE, 2006.

\bibitem{shade1998layered}
Jonathan Shade, Steven Gortler, Li-wei He, and Richard Szeliski.
\newblock Layered depth images.
\newblock In {\em Proceedings of the 25th annual conference on Computer
  graphics and interactive techniques}, pages 231--242, 1998.

\bibitem{sitzmann2020implicit}
Vincent Sitzmann, Julien Martel, Alexander Bergman, David Lindell, and Gordon
  Wetzstein.
\newblock Implicit neural representations with periodic activation functions.
\newblock {\em Advances in Neural Information Processing Systems}, 33, 2020.

\bibitem{song2017semantic}
Shuran Song, Fisher Yu, Andy Zeng, Angel~X Chang, Manolis Savva, and Thomas
  Funkhouser.
\newblock Semantic scene completion from a single depth image.
\newblock In {\em Proceedings of the IEEE Conference on Computer Vision and
  Pattern Recognition}, pages 1746--1754, 2017.

\bibitem{sun2021neuralrecon}
Jiaming Sun, Yiming Xie, Linghao Chen, Xiaowei Zhou, and Hujun Bao.
\newblock Neuralrecon: Real-time coherent 3d reconstruction from monocular
  video.
\newblock In {\em Proceedings of the IEEE/CVF Conference on Computer Vision and
  Pattern Recognition}, pages 15598--15607, 2021.

\bibitem{sun2018pix3d}
Xingyuan Sun, Jiajun Wu, Xiuming Zhang, Zhoutong Zhang, Chengkai Zhang, Tianfan
  Xue, Joshua~B Tenenbaum, and William~T Freeman.
\newblock Pix3d: Dataset and methods for single-image 3d shape modeling.
\newblock In {\em Proceedings of the IEEE Conference on Computer Vision and
  Pattern Recognition}, pages 2974--2983, 2018.

\bibitem{tatarchenko2019single}
Maxim Tatarchenko, Stephan~R Richter, Ren{\'e} Ranftl, Zhuwen Li, Vladlen
  Koltun, and Thomas Brox.
\newblock What do single-view 3d reconstruction networks learn?
\newblock In {\em Proceedings of the IEEE Conference on Computer Vision and
  Pattern Recognition}, pages 3405--3414, 2019.

\bibitem{tian2020fcos}
Zhi Tian, Chunhua Shen, Hao Chen, and Tong He.
\newblock {FCOS: A Simple and Strong Anchor-Free Object Detector}.
\newblock {\em TPAMI}, 2020.

\bibitem{tulsiani2018factoring}
Shubham Tulsiani, Saurabh Gupta, David~F Fouhey, Alexei~A Efros, and Jitendra
  Malik.
\newblock Factoring shape, pose, and layout from the 2d image of a 3d scene.
\newblock In {\em Proceedings of the IEEE Conference on Computer Vision and
  Pattern Recognition}, pages 302--310, 2018.

\bibitem{tulsiani2017learning}
Shubham Tulsiani, Hao Su, Leonidas~J Guibas, Alexei~A Efros, and Jitendra
  Malik.
\newblock Learning shape abstractions by assembling volumetric primitives.
\newblock In {\em Proceedings of the IEEE Conference on Computer Vision and
  Pattern Recognition}, pages 2635--2643, 2017.

\bibitem{lsiTulsiani18}
Shubham Tulsiani, Richard Tucker, and Noah Snavely.
\newblock Layer-structured 3d scene inference via view synthesis.
\newblock In {\em ECCV}, 2018.

\bibitem{2020SciPy-NMeth}
Pauli Virtanen, Ralf Gommers, Travis~E. Oliphant, Matt Haberland, Tyler Reddy,
  David Cournapeau, Evgeni Burovski, Pearu Peterson, Warren Weckesser, Jonathan
  Bright, St{\'e}fan~J. {van der Walt}, Matthew Brett, Joshua Wilson, K.~Jarrod
  Millman, Nikolay Mayorov, Andrew R.~J. Nelson, Eric Jones, Robert Kern, Eric
  Larson, C~J Carey, {\.I}lhan Polat, Yu Feng, Eric~W. Moore, Jake
  {VanderPlas}, Denis Laxalde, Josef Perktold, Robert Cimrman, Ian Henriksen,
  E.~A. Quintero, Charles~R. Harris, Anne~M. Archibald, Ant{\^o}nio~H. Ribeiro,
  Fabian Pedregosa, Paul {van Mulbregt}, and {SciPy 1.0 Contributors}.
\newblock {{SciPy} 1.0: Fundamental Algorithms for Scientific Computing in
  Python}.
\newblock {\em Nature Methods}, 17:261--272, 2020.

\bibitem{wang2018pixel2mesh}
Nanyang Wang, Yinda Zhang, Zhuwen Li, Yanwei Fu, Wei Liu, and Yu-Gang Jiang.
\newblock Pixel2mesh: Generating 3d mesh models from single rgb images.
\newblock In {\em ECCV}, 2018.

\bibitem{Wang15}
X. Wang, David~F. Fouhey, and A. Gupta.
\newblock Designing deep networks for surface normal estimation.
\newblock In {\em CVPR}, 2015.

\bibitem{weyn2020improving}
Jonathan~A Weyn, Dale~R Durran, and Rich Caruana.
\newblock Improving data-driven global weather prediction using deep
  convolutional neural networks on a cubed sphere.
\newblock {\em Journal of Advances in Modeling Earth Systems},
  12(9):e2020MS002109, 2020.

\bibitem{xu2019disn}
Qiangeng Xu, Weiyue Wang, Duygu Ceylan, Radomir Mech, and Ulrich Neumann.
\newblock Disn: Deep implicit surface network for high-quality single-view 3d
  reconstruction.
\newblock In {\em Advances in Neural Information Processing Systems}, pages
  492--502, 2019.

\bibitem{yu2020pixelnerf}
Alex Yu, Vickie Ye, Matthew Tancik, and Angjoo Kanazawa.
\newblock {pixelNeRF}: Neural radiance fields from one or few images.
\newblock In {\em CVPR}, 2021.

\bibitem{zhang2020nerfplusplus}
Kai Zhang, Gernot Riegler, Noah Snavely, and Vladlen Koltun.
\newblock Nerf++: Analyzing and improving neural radiance fields.
\newblock {\em arXiv preprint arXiv:2010.07492}, 2020.

\bibitem{zhang2016colorful}
Richard Zhang, Phillip Isola, and Alexei~A Efros.
\newblock Colorful image colorization.
\newblock In {\em ECCV}, 2016.

\bibitem{zhou2019objects}
Xingyi Zhou, Dequan Wang, and Philipp Kr{\"a}henb{\"u}hl.
\newblock Objects as points.
\newblock {\em arXiv preprint arXiv:1904.07850}, 2019.

\end{thebibliography}
}
\newpage
\setcounter{section}{0}
\renewcommand\thesection{\Alph{section}}
\section*{Appendix}




\noindent We present additional details on experiments,  qualitative results as figures and videos in the following sections. We present complete derivations and further detailed analysis in this supplemental.

Details of experiments for scene and ray based evaluation appear under \S \ref{sec:add_exps} along with new ray based evaluations. In \S \ref{sec:decode_exps} we provide full experiment evaluation of different decoding strategies of baselines. In \S \ref{sec:dataset} we discuss complete details for all of the datasets.  Then in \S \ref{sec:analysis} discuss the detail behavior of different distance functions under uncertainty and graph them. We follow this up with additional qualitative results on all the three datasets on randomly sampled images in the test set in \S \ref{sec:qual}. In \S \ref{sec:deriv} we derive the results mathematically to showcase the analysis technique for other  distance functions.
\section{Experiments}

\subsection{Additional Evaluation for Baselines}
\noindent In the main paper we presented results from Tab. \ref{tab:scene} (scene based evaluation) and Tab. \ref{tab:ray_inv} (ray based evaluation on occluded points). Additionally we also present ray based evaluation on all intersections/points along the ray in Tab. \ref{tab:ray_all}. We present results on all the three datasets \hlc[dred]{Matterport}, \hlc[dblue]{3DFront}, and \hlc[dpurple]{ScanNet} like in the main paper. Now for the sake of completeness we revisit the metrics again.

\parnobf{Scene (Acc/Cmp/F1)} Like~\cite{seitz2006comparison,tatarchenko2019single}, we report accuracy/Acc ( \% of predicted points within $t$ of the ground-truth), completeness/Cmp ( \% of ground-truth points within $t$ of the prediction), and their harmonic mean, F1-score. This gives a summary of overall scene-level accuracy. Results are reported in Tab. \ref{tab:scene} as presented in the main paper.

\begin{table}[h]
    \setlength{\tabcolsep}{3pt} 
    
    \centering 
    \caption{\redtext{(From Main Paper) }{\bf Scene Acc/Comp/F1Score}. Thresholds: 0.5m (MP3D~\cite{chang2017matterport3d}, \tdf), 0.2m (\scannet). {\bf Bold is best}, \underline{underline is 2\textsuperscript{nd} best} per column. \tsdf is best in F1 and accuracy, and always comparable to the best in completeness.}

    \label{tab:scene}
    \begin{adjustbox}{max width=\linewidth}
    \begin{tabular}{lm@{~}m@{~}mt@{~}t@{~}ts@{~}s@{~}s} \toprule
    & \multicolumn{3}{m}{MP3D \cite{chang2017matterport3d}} & \multicolumn{3}{t}{\tdf } & \multicolumn{3}{s}{\scannet}  \\ 
    Method   & Acc & Cmp & F1 & Acc  & Cmp & F1 & Acc & Cmp & F1 \\ 
    \cmidrule(r){1-1}\cmidrule(r){2-4} \cmidrule(r){5-7} \cmidrule(r){8-10}
      \mpd  & 66.2 & \underline{72.4} & 67.4 & 68.6 & 46.5 & 52.7 & 19.3 & 28.6 & 21.5 \\ 
     \mpd+C  & 64.8 & 55.1 & 57.7 & 70.8 & 45.1 & 52.4 & 19.9 & 32.0 & 23.3 \\ 
     \sal \cite{atzmon2020sal}  & 66.1 & 25.5 & 34.3 & 80.7 & 28.5 & 39.5 & 51.2 & \bf 70.0 & 57.7 \\ 
     \scened  & 58.7 & \bf 76.0 & 64.7 & 70.1 & \underline{51.9} & 57.4 & 44.4 & {62.6} & 50.8 \\ 
     \occ  & 73.4 & 69.4 & \underline{69.6} & \underline{86.4} & 48.1 & \underline{59.6} & 51.5 & 58.5 & 53.7 \\ 
     \urdf  & \underline{74.5} & 67.1 & 68.7 & 85.0 & 47.7 & 58.7 & \underline{61.0} & 57.8 & \underline{58.2} \\ 
     \tsdf (ours)  & \bf 75.4 & 72.0 & \bf 71.9 & \bf 87.3 & \bf 52.6 & \bf 63.4 & \bf 62.0 & \underline {62.7} & \bf 60.9 \\ 
    \bottomrule
    \end{tabular}
    \end{adjustbox}
    
    \end{table}

\parnobf{{Rays (Acc/Cmp/F1), Occluded Points}} We additionally evaluate each ray independently, measuring Acc/Cmp/F1 on each ray and reporting the mean. Occluded points are defined as all surfaces past the first for both ground-truth and predicted. Evaluating each ray independently applies a more stringent test for occluded surfaces compared to scenes: with scene-level evaluation on a high resolution image, a prediction can miss a hidden surface (e.g., the 2nd surface) on every other pixel since the missing predictions will be covered for by hidden surfaces in adjacent rays. Ray-based evaluation requires each pixel to have all hidden surfaces present to receive full credit.
Results are reported in Tab. \ref{tab:ray_inv}. 

\parnobf{{Rays (Acc/Cmp/F1), All Points}} We evaluate each ray independently, measuring Acc/Cmp/F1 on each ray and reporting the mean. Unlike the occluded version of this metric we do not drop the first surface and evaluate using all the ground truth and predicted intersections. This metric has similar properties as the {\it Occluded} metric but applies the stringent test to all intersections. Results are reported in Tab. \ref{tab:ray_all}.  We note that, except for \sal on \scannet which gives higher Cmp. as compared to \tsdf at the cost of accuracy where \tsdf is the next best;  \tsdf always outperforms all the baselines on Acc/Cmp/F1

\begin{table}[t]
\setlength{\tabcolsep}{3pt} 
\centering 
\caption{\redtext{(From Main Paper)} {\bf Ray Acc/Comp/F1Score on Occluded Points}. Thresholds: 0.5m (MP3D~\cite{chang2017matterport3d}, 3DFront~\cite{fu20203dfront}), 0.2m (ScanNet~\cite{dai2017scannet}). 
\tsdf is best on F1 and Acc, and is occasionally 2\textsuperscript{nd} best on Cmp. Gains on occluded points are even larger than the full scene.}
\label{tab:ray_inv}
\begin{adjustbox}{max width=\linewidth}
\begin{tabular}{lm@{~}m@{~}mt@{~}t@{~}ts@{~}s@{~}s} \toprule
& \multicolumn{3}{m}{MP3D \cite{chang2017matterport3d}} & \multicolumn{3}{t}{\tdf } & \multicolumn{3}{s}{\scannet}  \\ 
Method   & Acc & Cmp & F1 & Acc  & Cmp & F1 & Acc & Cmp & F1 \\ 
\cmidrule(r){1-1}\cmidrule(r){2-4} \cmidrule(r){5-7} \cmidrule(r){8-10}
    \mpd  & 13.9 & \bf 42.8 & 19.3 & 17.8 & \underline{35.8} & 22.2 & 0.5 & 9.0 & 2.4 \\ 
    \mpd+C  & 18.7 & 21.7 & 19.3 & 17.7 & 22.6 & 19.9 & 1.1 & 2.4 & 3.5 \\ 
    \sal \cite{atzmon2020sal}  & 5.5 & 0.5 & 3.5 & 24.1 & 4.3 & 11.4 & 2.4 & \bf 38.7 & 5.6 \\ 
    \scened  & 15.5 & 23.0 & 16.6 & 29.3 & 21.3 & 23.4 & 1.8 & 7.8 & 5.5 \\ 
    \occ  & \underline{26.2} & 20.5 & \underline{21.6} & 53.2 & 22.0 & \underline{31.0} & 6.6 & 12.3 & 11.4 \\ 
    \urdf  & 24.9 & 20.6 & 20.7 & \underline{47.7} & 23.3 & 30.2 & \underline{8.4} & 11.6 & \underline{13.8} \\ 
    \tsdf (ours)  & \bf 28.4 & \underline{30.0} & \bf 27.3 & \bf 54.6 & \bf 56.0 & \bf 52.6 & \bf 9.0 & \underline{20.4} & \bf 16.0 \\ 
\bottomrule
\end{tabular}
\end{adjustbox}

\end{table}

\begin{table}[t]
    \setlength{\tabcolsep}{3pt} 
    \centering 
    \caption{\redtext{(Supplemental Table)} {\bf Ray Acc/Comp/F1Score on All Points}. Thresholds: 0.5m (MP3D~\cite{chang2017matterport3d}, 3DFront~\cite{fu20203dfront}), 0.2m (ScanNet~\cite{dai2017scannet}). 
    \tsdf is best on F1 and Acc, and is occasionally 2\textsuperscript{nd} best on Cmp. Gains on occluded points are even larger than the full scene.}
    \label{tab:ray_all}
    \begin{adjustbox}{max width=\linewidth}
    \begin{tabular}{lm@{~}m@{~}mt@{~}t@{~}ts@{~}s@{~}s} \toprule
    & \multicolumn{3}{m}{MP3D \cite{chang2017matterport3d}} & \multicolumn{3}{t}{\tdf } & \multicolumn{3}{s}{\scannet}  \\ 
    Method   & Acc & Cmp & F1 & Acc  & Cmp & F1 & Acc & Cmp & F1 \\ 
    \cmidrule(r){1-1}\cmidrule(r){2-4} \cmidrule(r){5-7} \cmidrule(r){8-10}
    \mpd  & 28.8 & \underline{50.7} & 35.7 & 34.1 & 49.8 & 39.0 & 7.3 & 18.2 & 11.7 \\ 
    \mpd+C  & 26.8 & 30.1 & 27.8 & 30.6 & 33.9 & 31.6 & 4.7 & 7.4 & 8.0 \\ 
    \sal \cite{atzmon2020sal}  & 27.2 & 19.1 & 22.4 & 43.9 & 25.8 & 31.6 & 31.0 & \bf 60.4 & \underline{40.8} \\ 
    \scened  & 32.7 & 45.3 & 36.8 & 51.5 & 57.1 & 52.0 & 27.8 & 38.9 & 32.5 \\ 
    \occ  & \underline{46.4} & 49.8 & \underline{46.9} & \underline{71.8} & \underline{66.2} & \underline{67.1} & 34.1 & 39.7 & 36.6 \\ 
    \urdf  & 45.2 & 46.6 & 44.8 & 66.0 & 56.9 & 59.3 & \underline{37.3} & 39.4 & 38.7 \\ 
    \tsdf  & \bf 48.3 & \bf 55.0 & \bf 50.3 & \bf 74.9 & \bf 76.3 & \bf 74.1 & \bf 40.3 & \underline{45.7} & \bf 43.0 \\     
    \bottomrule
    \end{tabular}
    \end{adjustbox}
    
    \end{table}

\label{sec:add_exps}

\subsection{Effect of different Decodings}
Decoding strategies are important and different for all distance functions. The performance of all methods that predict distance function depend's on their decoding strategy. Therefore methods with almost no hyper-parameters  in decoding strategies are desirable. \scened, \urdf, \occ all require rigorous optimization of a decoding strategy. DRDF's decoding strategy is simple and hyperparameter free which involves only finding positive to negative zero crossings. For other methods we optimize this strategy extensively.
\label{sec:decode_exps}
We discuss the impact of alternate decoding strategies for baselines in detail here. For the sake of brevity we reported numbers only on Scene-F1 score for all decoding strategies in the text of the main paper . Here we report the numbers on all the metrics for baseline and their alternate decoding strategies in Tab \ref{tab:decoding}. We first describe \scened, followed by \urdf and then followed by \occ.

\parnobf{\scened} We tried two other decoding strategies with \scened namely, absolute thresholding \emph {(\scened + Th.)} and Sphere tracing followed by gradient based optimization \emph {(\scened + Sph.)} as proposed by Chibane \etal~\cite{chibane2020ndf}. We observe as reported in the main paper that these two strategies do slightly worse on Scene F1 Score than our best reported strategy of using \texttt{scipy.argrelextrema} to find minimas of the distance function along the ray.

On other metrics of Ray based Acc/Cmp/F1 we observe that our strategy does especially well on discovering occluded regions. We speculate that using absolute thresholding is especially bad because of the behavior of global unsigned distance under uncertainty.  Moreover, due to the model's inability to mimic the GT \udf we find that using sphere tracing as proposed by Chibane \etal~\cite{chibane2020ndf} is not as effective.\\

\begin{table}[h]
    \setlength{\tabcolsep}{3pt} 
    
    \centering 
    \caption{\redtext{(Supplemental Table) }{\bf Effect of different decodings}. Thresholds: 0.5m (MP3D~\cite{chang2017matterport3d}). {\bf Bold is best} per column and section (created by horizontal lines). We compare alternate decoding strategies for baseline methods and report their performance on the three metrics ; Scene Acc/Cmp/F1, Ray Acc/Cmp/F1 All, Ra Acc/Cmp/F1 Occluded }
    
    \label{tab:decoding}
    \begin{adjustbox}{max width=\linewidth}
    \begin{tabular}{lm@{~}m@{~}mt@{~}t@{~}ts@{~}s@{~}s} \toprule
    & \multicolumn{3}{m}{Scene} & \multicolumn{3}{t}{Ray All} & \multicolumn{3}{s}{Ray Occluded}  \\ 
    Method   & Acc & Cmp & F1 & Acc  & Cmp & F1 & Acc & Cmp & F1 \\ 
    \cmidrule(r){1-1}\cmidrule(r){2-4} \cmidrule(r){5-7} \cmidrule(r){8-10}
    \scened+ Th.  & 79.3 & 49.7 & 58.8 & 46.2 & 27.6 & 32.9 & 18.0 & 2.6 & 5.2 \\ 
    \scened+ Sph.  & 50.8 & 32.8 & 37.5 & 19.3 & 22.6 & 20.1 & 11.4 & 2.1 & 4.7 \\ 
    \scened  & 58.7 & 76.0 & \bf 64.7 & 32.7 & 45.3 & 36.8 & 15.5 & 23.0 & 16.6 \\ 
    \midrule
    
    \urdf+ Th.  & 82.5 & 55.1 & 63.3 & 56.7 & 47.7 & 50.1 & 16.6 & 30.6 & 18.6 \\ 
    \urdf+ Grd.  & 48.5 &  75.8 & 57.4 & 24.6 & 50.9 & 32.2 & 11.4 & 37.2 & 16.1 \\ 
    \urdf+ Sph.  & 59.1 & 69.4 & 62.4 & 45.9 & 46.5 & 44.8 & 23.8 & 15.3 & 16.7 \\ 
    \urdf  & 74.5 & 67.1 & \bf 68.7 & 45.2 & 46.6 & 44.8 & 24.9 & 20.6 &  20.7 \\ 
    \midrule
    \occ+ Sngle.  & 70.9 & 62.9 & 64.7 & 36.6 & 37.5 & 36.2 & 22.0 & 18.9 & 18.9 \\ 
    \occ  & 73.4 & 69.4 & \bf 69.6 & 46.4 & 49.8 & 46.9 & 26.2 & 20.5 & 21.6 \\ 
    \midrule
    \tsdf (ours)  & 75.4 & 72.0 & 71.9 & 48.3 & 55.0 & 50.3 & 28.4 & 30.0 & 27.3 \\ 
    \bottomrule
    \end{tabular}
    \end{adjustbox}
    
    \end{table}




\parnobf{\urdf} We use three alternate decoding strategies to best recover the surface locations for model trained with unsigned ray distance. First, we use absolute thresholding \emph {(\urdf + Th.) }on the predicted distance by considering all points with value distance prediction $\le \tau$. We choose $\tau$ by cross-validation. Second, we use the numerical gradient to find the zero crossings of the gradient functions hence detecting the minimas \emph{(\urdf + Grd.)}. Thirdly, we use sphere tracing followed by gradient based optimization from Chibane \etal~\cite{chibane2020ndf} \emph{(\urdf + Sph.)}. As reported in the main text all these strategies perform worse on Scene F1 score with regards to our strategy that does non-maximum suppression on the thresholded data by finding connected components of the ray that have predicted distance below a tuned-constant $\tau$.

On other metrics of Ray based Acc/Cmp/F1 we observe that \urdf with our decoding strategy outperforms all alternate choices on considerably on the occluded points. \urdf + Th. and \urdf + Grad. tend to high F1 scores on occluded points but this is due high completion scores that these methods have as compared to their accuracy. \urdf + Sph. does reasonably well on the occluded points but is outperformed likely as it assumes that the predicted \urdf behaves like a GT \urdf. 

\parnobf{\occ} Our choice of decoding strategy for \occ is based on the fact that \occ predicts an onset and a offset zero crossing. We keep the average of onset and the offset crossings when we find pairs and keep the single crossing otherwise. An alternate decoding strategy is to only keep only one of the zero crossings. We report the scene F1 score for keeping on a single zero crossing under \emph{\occ+Sngl.} and see that it under performs our strategy by (4.9 points). Additionally our \occ decoding strategy also outperforms on Ray based metrics.

\section{Dataset Details}
We discuss the complete details for all datasets from the main paper. We only use the images and corresponding 3D meshes from these datasets. We use Trimesh~\cite{trimesh} to process the meshes and compute ray intersections using the embree backend support.
\label{sec:dataset}
\parnobf{\matterport} Matterport3D is a real dataset of scans collected in houses using the Matterport3D camera. The datasets provides access to images and their corresponding ground truth 3D in the form on non-watertight meshes. Since houses in Matterport3D are big, we clip scene to 8 meters in depth from the camera and only considering the mesh that is within $8m$ camera frustum. We use this mesh to compute the ground truth intersections and distance function values. Such a large range is necessary as Matterport3D is a collection of rooms and this allows other models to predict additional rooms behind walls.

\parnobf{\tdf} Similar to Matterport3D, 3DFront is also a collection of houses with only key difference being it is synthetic. In 3DFront also we clip scene to 8 meters in depth from the camera so only considering the mesh that is within $8m$ to compute the locations of ground truth intersections.

\parnobf{\scannet} We use splits from \cite{dai2017scannet} (1045/156/312 train/val/test) and randomly select 5 images per scene for train set, and 10 images per scene for val/test set. We then sample to a set of 33K/1K/1K images per train/val/test. For ScanNet we clip the scene t 4 meters in depth from the camera so only considering mesh that is within $4m$ to compute the locations of ground truth intersections. We use a smaller range than \matterport and \tdf because this dataset mostly has individual rooms for which this range suffices.


\section{Distance Functions Under Uncertainty}
\label{sec:analysis}

When predicting distance functions for a complex 3D scene for single image predicting exact distances for the geometry is intrinsically uncertain. It is hard to predict the exact location of a particular object accurately, but it is often easier to know the general layout of the scene. 

This uncertainty in predicting exact distances coupled with using a network that is trained to minimized the MSE loss encourages networks to produce the expected values of distance functions. The predicted distance function lack critical properties of actual distance function, and we discuss the how these differ across different distance functions. This section presents all the results that are important to understand the differences in behavior of the ray-based distance function  and we directly state them. All the derivations for these results are in \S \ref{sec:deriv} (towards the end of this supplement). 

We analyze a single intersection along ray and behavior of distance functions with it. We will start with the setup, then show key results for different ray distance functions. 

\parnobf{Setup} 
We assume we are predicting a distance-like function along a ray. Given surface geometry the ray intersects, the distance function's value is a function of the distance $z$ along the ray.

We analyze the case of uncertainty about a single intersection.  We assume the intersection's location is a random variable $S$ that is normally distributed with mean $\mu$ (the intersection's location) and standard deviation $\sigma$. 
Throughout, WLOG, we assume the intersection is at $\mu = 0$ for convenience. This considerably simplifies some expressions, and
can be done freely since we are free to pick the coordinate system. The rest follows the main paper.
Let $p(s)$ denote the density and $\F(s)$ denote the CDF for samples from $S$. We assume that the distance to the second intersection is $n \in \mathbb{R}^+$, which we will assume is not a random variable for simplicity (i.e., the second intersection is at $S+n$, which is normally distributed with mean $n$ and standard deviation $\sigma$).

Given a value $s$ for the intersection location, we can instantiate the distance function. We denote the value of the distance function at $z$ if the intersection is at $s$ as $d(z;s): \mathbb{R} \to \mathbb{R}$ that maps a location $z$ on the ray to a (possibly signed) distance. The distance if $s$ is at the real location is $d(z;0)$.

\parnobf{Training} 
During training, a function approximator is trained to minimize a loss
function that measures the distance between its predictions and the
ground-truth. The optimal behavior of this function approximator is 
to output the value that minimizes the loss function. One critical value is the {\it expected} value of the distance function 
\begin{equation}
\label{eqn:expected}
\E[d(z;s)] = \int_{\mathbb{R}} d(z;s) p(s) ds.
\end{equation}
Eqn.~\ref{eqn:expected} is important for various loss functions:
\begin{itemize}
    \item Mean Squared Error (MSE): $\E[d(z;s)]$ is the optimal value when minimizing the MSE. 
    \item Cross-Entropy: If $d(z;s)$ is an indicator function (i.e., producing either $1$/positive or $0$/negative), then $\E(d(z;s)]$ minimizes the cross-entropy loss as well. This follows from the fact that $\E[d(z;s)]$ is the chance $z$ is positive, and a cross-entropy loss is minimized by matching frequencies.
    \item L1 Loss: The median (rather than the mean) is the optimizer for the L1 loss. However, the median and mean are the same for symmetric distributions. If one calculates $d(z;S)$ (where $S$ is the random variable rather than a particular value), one obtains a new random variable. If this distribution is symmetric, then the mean and median are the same, and therefore $\E[d(z;s)]$ minimizes the L1 loss too. In practice (see \S\ref{sec:analysis_median}, Fig.~\ref{fig:median}), we empirically find that any deviations between the mean and median are small, and thus the mean and median are virtually identical almost all of the time. 
\end{itemize}


We can think of this setting from two angles:
\begin{enumerate}
\item $\E[d(z;s)]$ as a 1D function of $z$. In our setting, our neural networks are incentivized to minimize their distance from this value; the $S$ is implicit. This is the primary way that we look at the problem since it gives us a function of $z$. We can then do things like compute $\frac{\partial}{\partial z}$.
\item $d(z;S)$ as a distribution over the distance for some fixed $z$. We use this angle to explain why the mean and median are similar in most cases.
\end{enumerate}

\begin{figure*}
\centering
\begin{tabular}{cc}
\multicolumn{2}{c}{\bf Signed Ray Distance Function} \\
\includegraphics[height=1.6in]{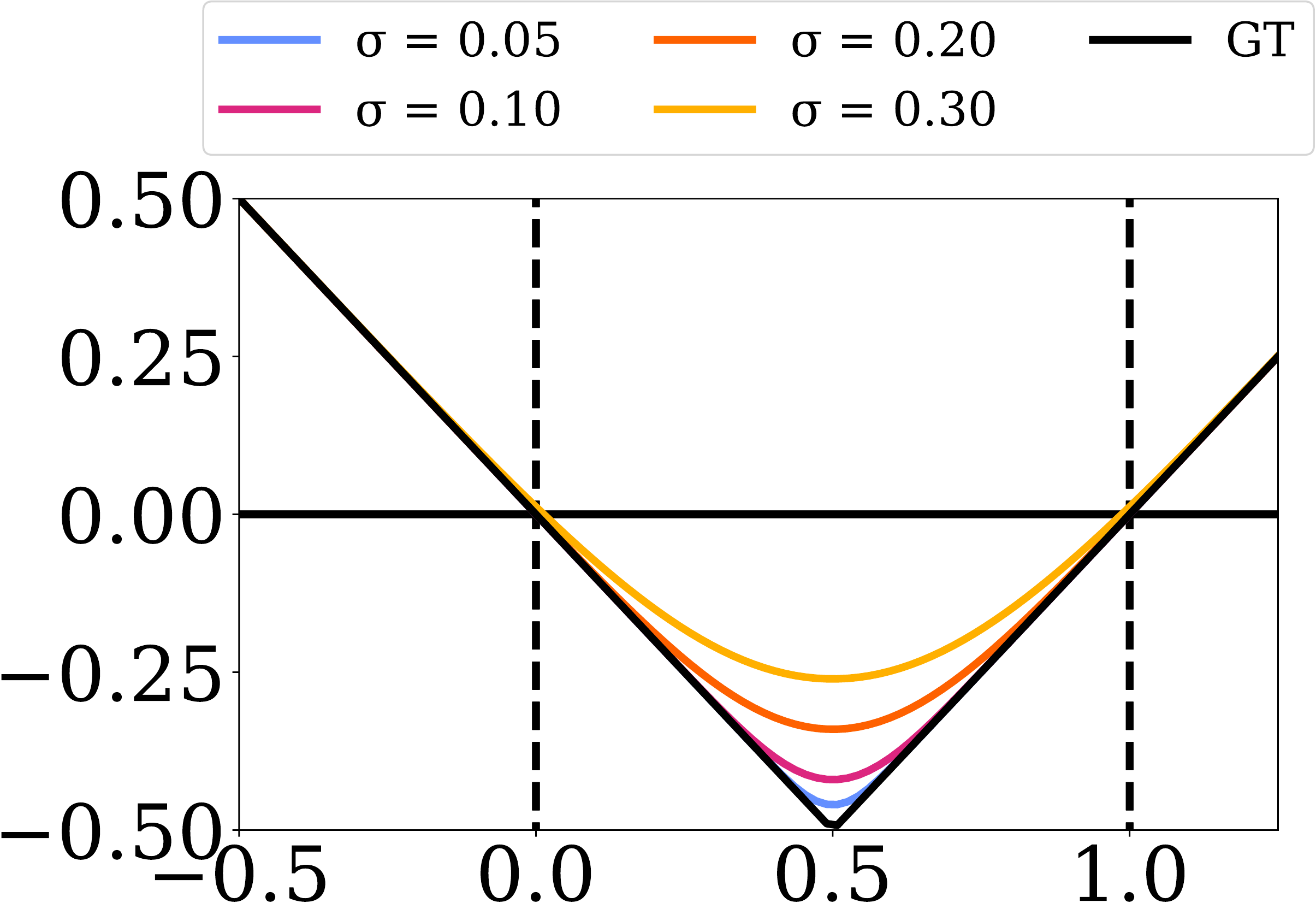} &
\includegraphics[height=1.6in]{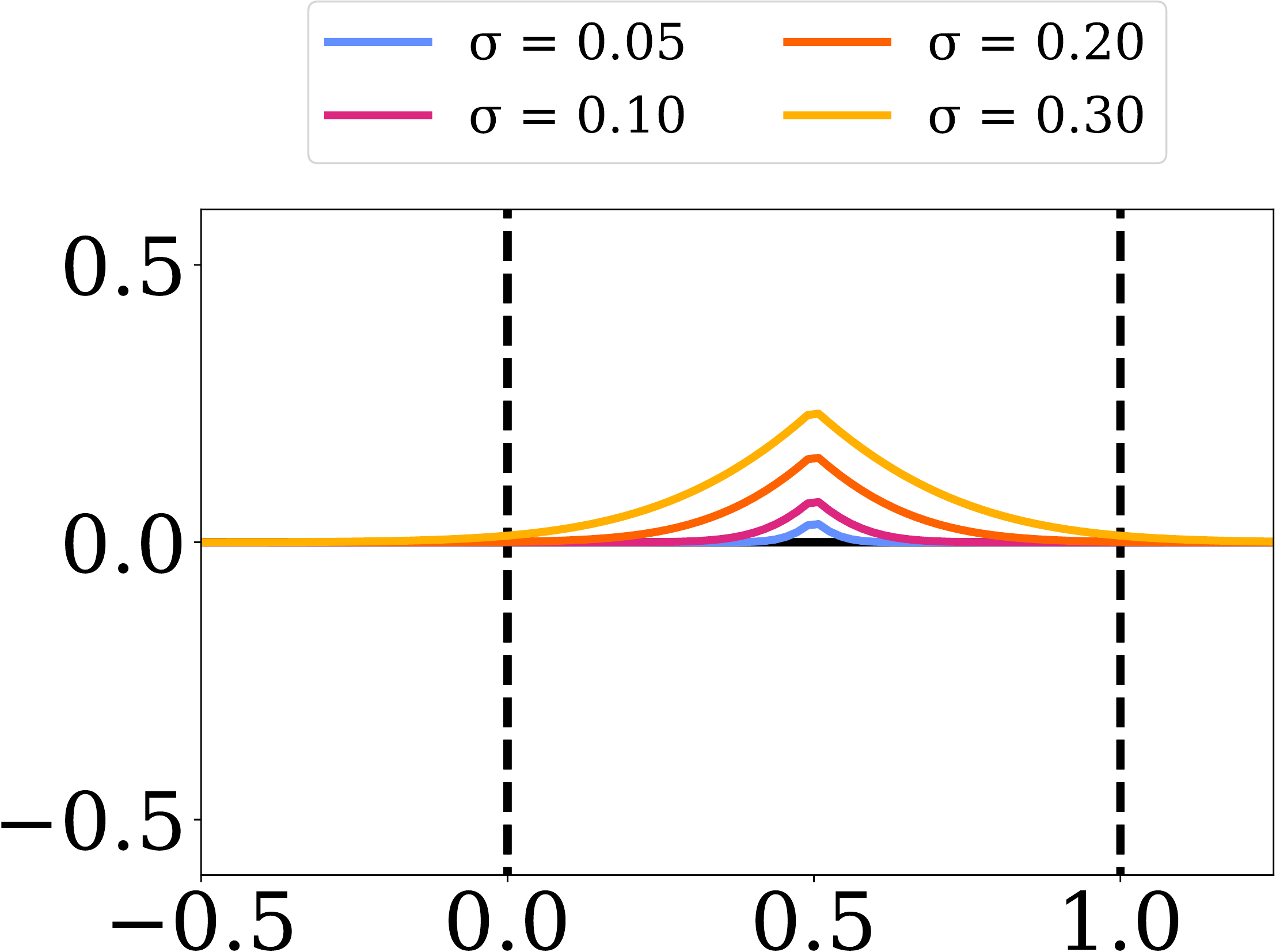} \\

\multicolumn{2}{c}{\bf Unsigned Ray Distance Function} \\
\includegraphics[height=1.6in]{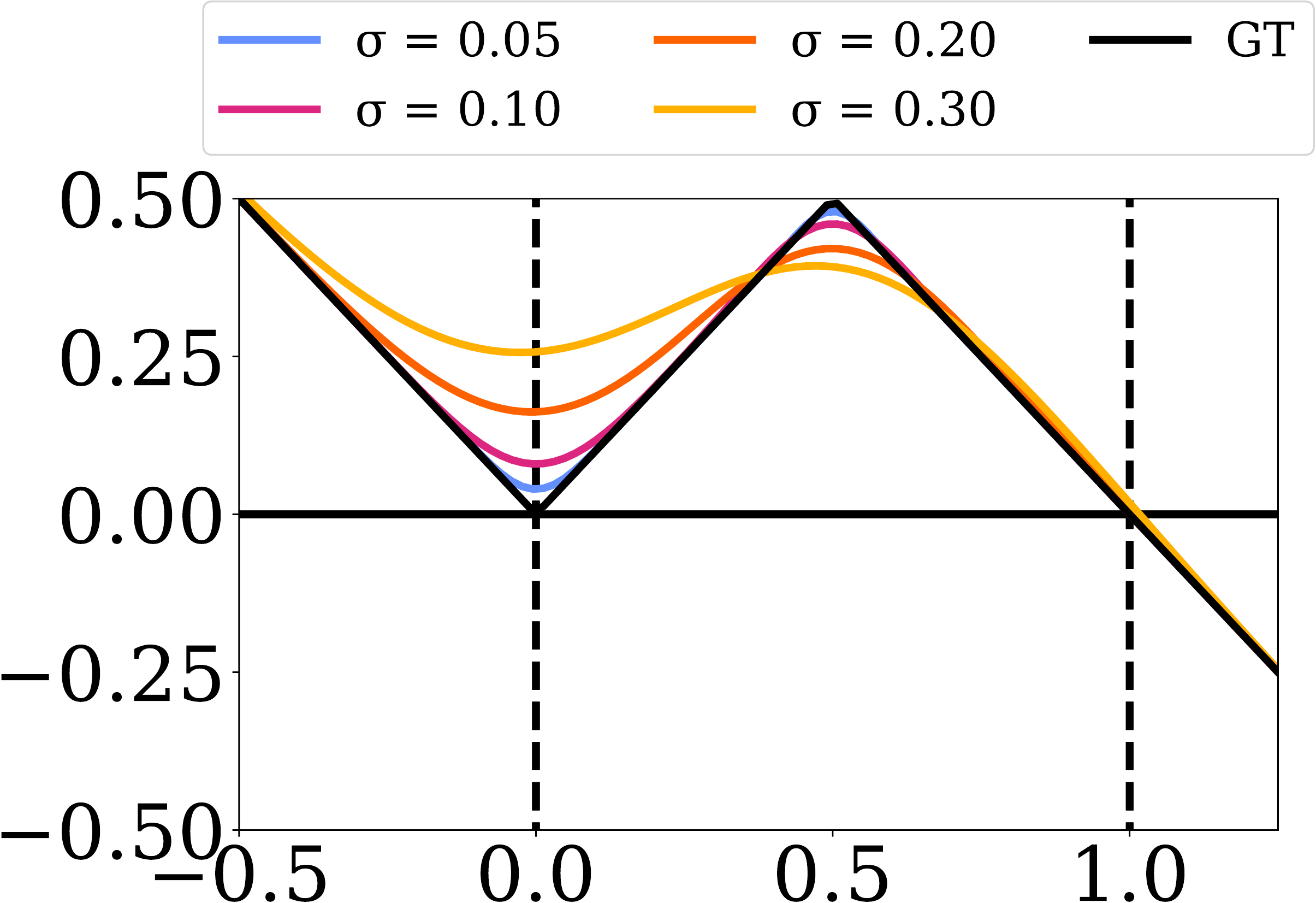} &
\includegraphics[height=1.6in]{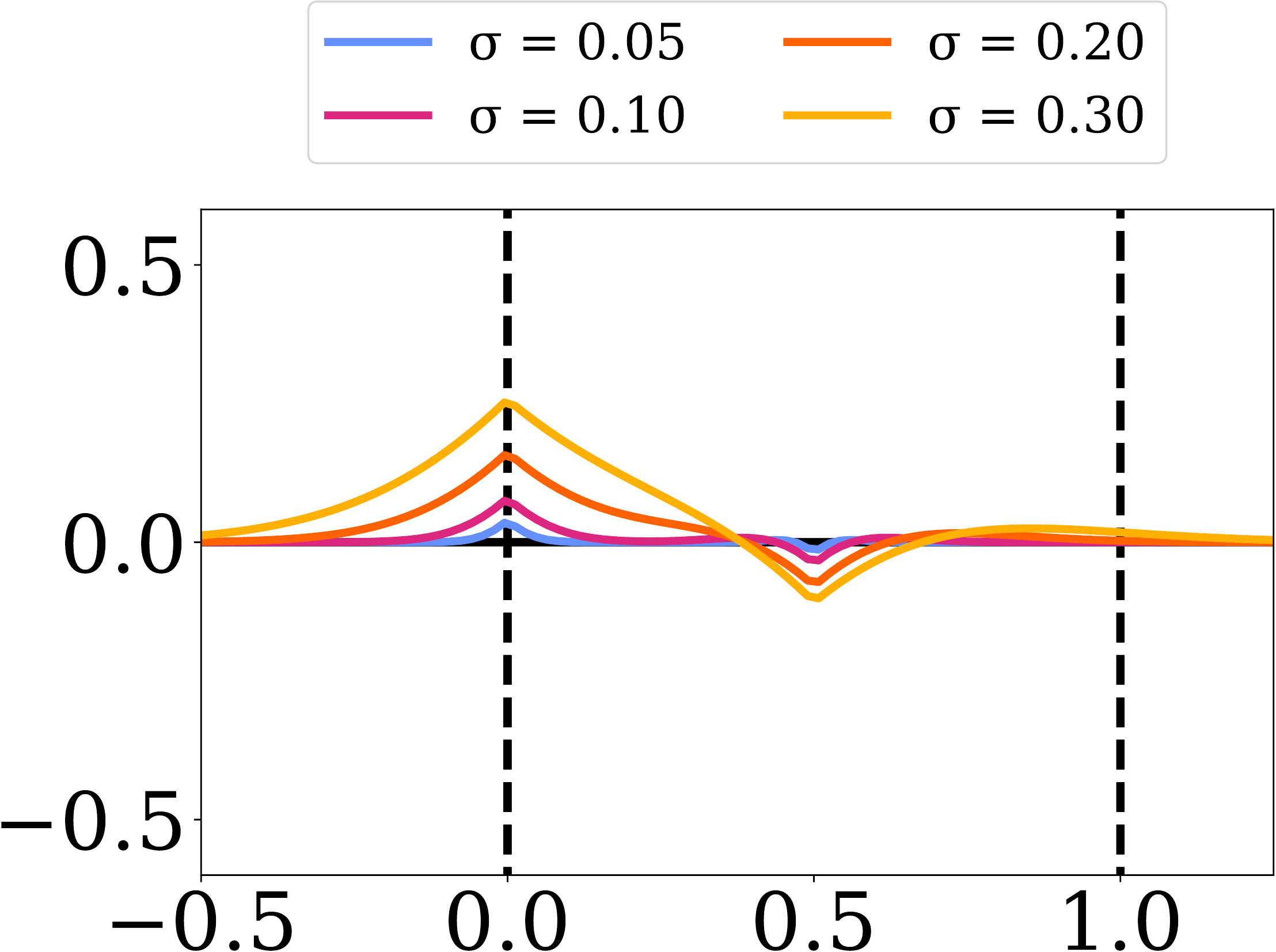} \\
\end{tabular}
\caption{Expected distance functions and their deviation from the real distance function. We plot the expected distance function $\E[d(z;s)]$ ({\bf left}) and
the residual between the the expectation and the real distance function $\E[d(z;s)] - d(z;s)]$ ({\bf right}). In each case, we plot the expectation for four $\sigma$. In all cases the next intersection
is $n=1$ away, and so if the units were $m$, one could think of the noise as $5$, $10$, $20$, and $30$cm.
For the signed and unsigned distance functions, we plot the full versions that also account for the next intersection.}
\label{fig:distance}
\end{figure*}

\begin{figure*}
\centering
\begin{tabular}{cc}
\multicolumn{2}{c}{\bf Occupancy Ray Function} \\
\includegraphics[height=1.6in]{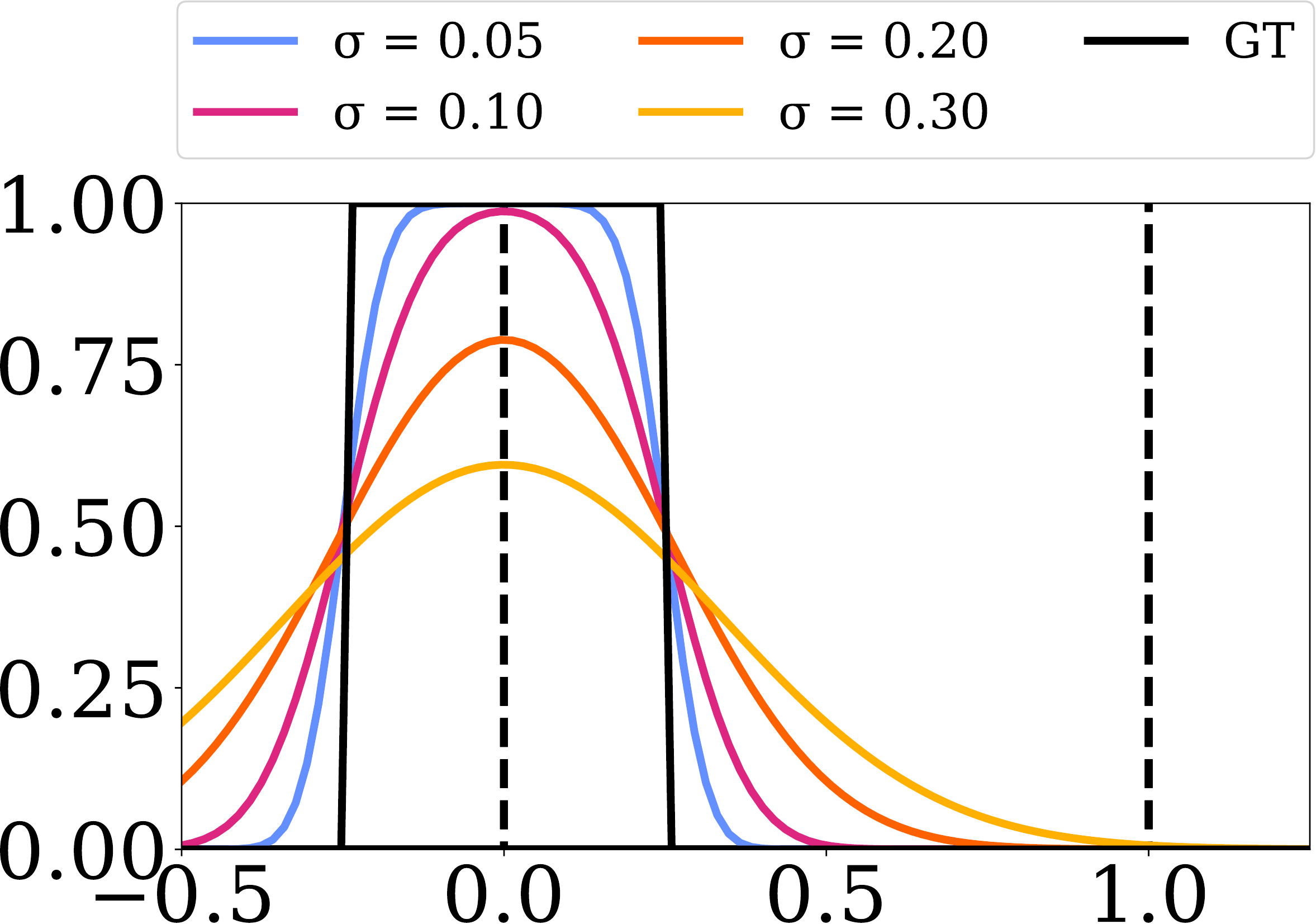} &
\includegraphics[height=1.6in]{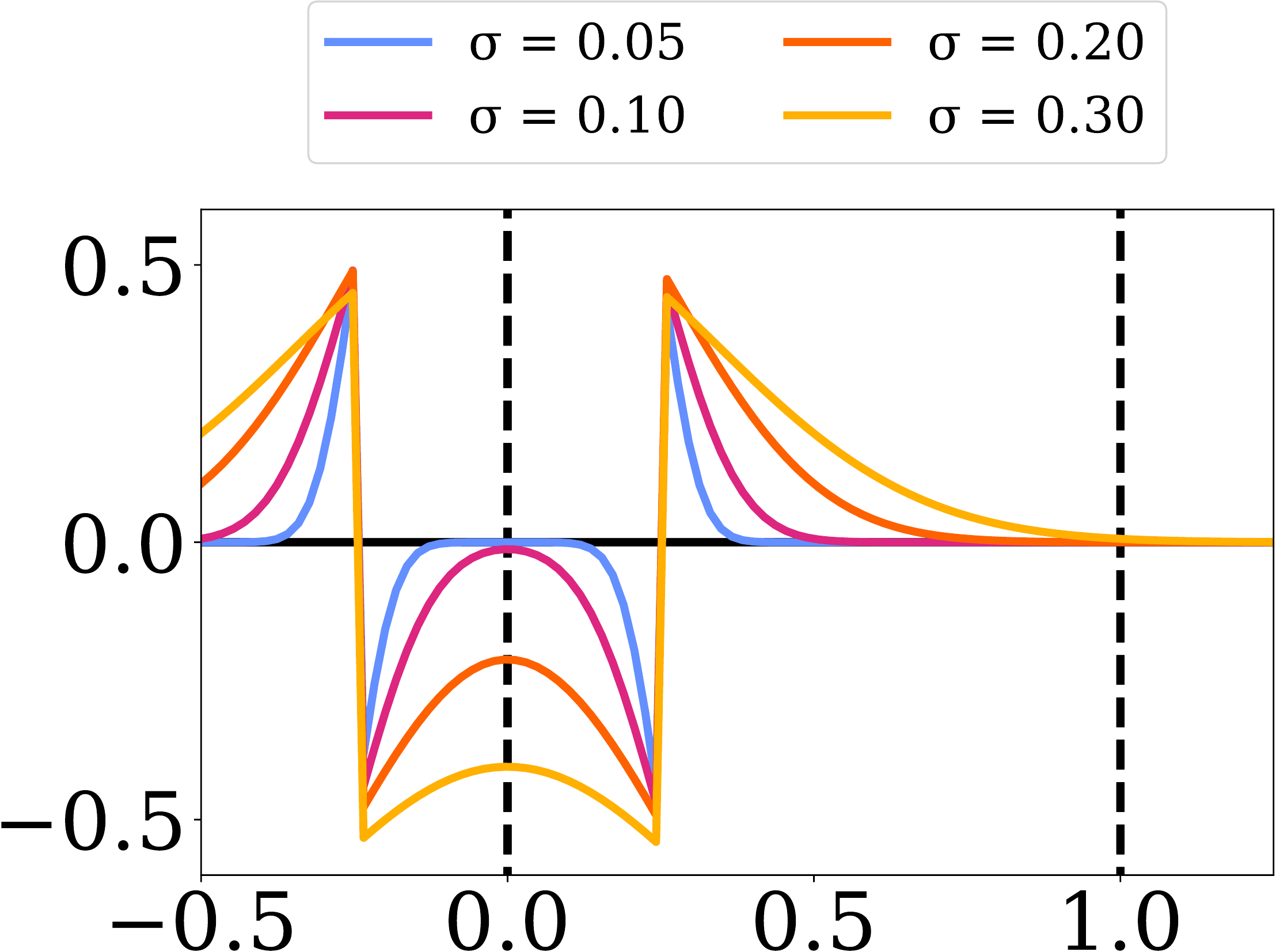} \\

\multicolumn{2}{c}{\bf Directed Ray Distance Function} \\
\includegraphics[height=1.6in]{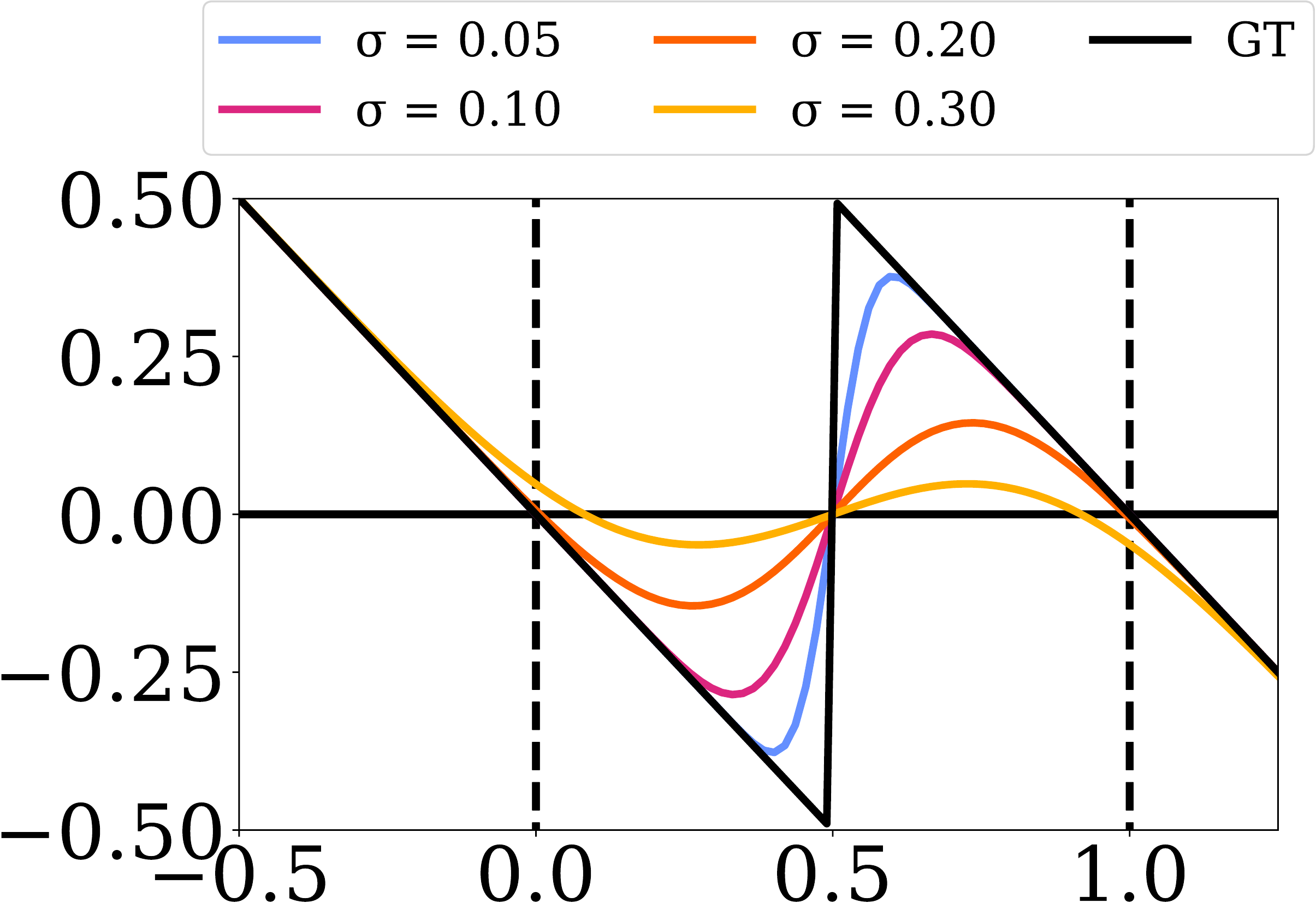} &
\includegraphics[height=1.6in]{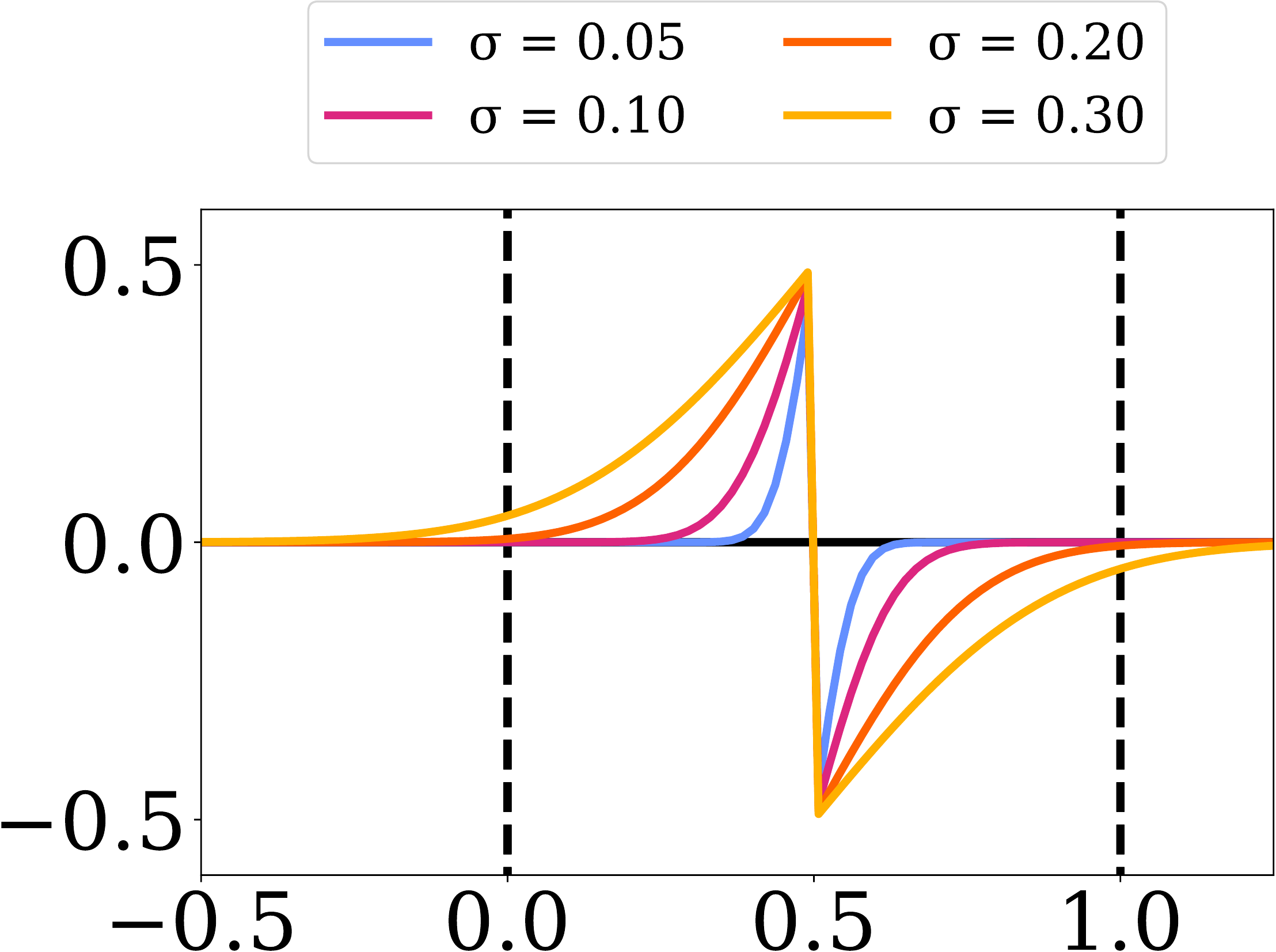} \\
\end{tabular}
\caption{Expected distance functions and their deviation from the real distance function. We plot the expected distance function $\E[d(z;s)]$ ({\bf left}) and
the residual between the the expectation and the real distance function $\E[d(z;s)] - d(z;s)]$ ({\bf right}). In each case, we plot the expectation for four $\sigma$. In all cases the next intersection
is $n=1$ away, and so if the units were $m$, one could think of the noise as $5$, $10$, $20$, and $30$cm.}
\label{fig:distance2}
\end{figure*}

We can then analyze the {\it expected} distance function ($\E[d(z;s)]$) for various distance functions, as well as the derivative $\frac{\partial}{\partial z}\E[d(z;s)]$, and the difference between the expected distance function and the underlying distance function ($\E[d(z;s)] - d(z;s)$). We plot distance functions and their difference from the ground truth in Figs.~\ref{fig:distance}, \ref{fig:distance2}.

The expected distance functions usually have two regimes: a regime in which they closely mimic the underlying distance function and a regime in which there are substantial distortions that are usually dependent on the level of uncertainty. These distance functions vary in where the distortions occur -- some have them near the intersection and others have them far away. When analyzing the expected functions, these regimes are caused by the PDF $p$ going to $0$ or the CDF $\F$ going to either $0$ or $1$.

\subsection{Signed Ray Distance Function (SRDF)}

Ignoring the second intersection, which has limited impact near the first intersection, the signed ray distance function (SRDF) is 
\begin{equation}
\dsr(z;s) = s-z
\end{equation}
assuming WLOG that $z<s$ is outside 
and positive. The expected distance function and its derivative are  
\begin{equation}
\E[\dsr(z;s)] = -z,~~~~\frac{\partial}{\partial z} \E[\dsr(z;s)] = -1.
\end{equation}
Considering the second intersection at $n$ creates additional terms in the expected SRDF that are negligible near 0, specifically
\begin{equation}
(2z-n) \F\left(z-\frac{n}{2}\right) + 2 \int_{z-\frac{n}{2}}^\infty s p(s) ds.
\end{equation}

\parnobf{Finding intersections} Finding the intersection is straightforward, since it is a zero-crossing and the expected function behaves like the actual function near the intersection.

\subsection{Unsigned Ray Distance Function (URDF)}

Likewise ignoring the second intersection, the unsigned ray distance function (URDF) is
\begin{equation}
\dur(z;s) = |s-z|.
\end{equation}
The expected distance function consists of three
terms that trade off in magnitude over the values of $z$: 
\begin{equation}
\E[\dur(z;s)] = z \F(z) + - z(1-\F(z)) + 2 \int_z^{\infty} s p(s) ds,
\end{equation}
which induces three regimes: 
$z$ when $z \gg 0$, $-z$ when $z \ll 0$, and a transitional regime near $0$. The trade off
between the regimes is controlled by $\F$ and $\int_z^{\infty} s p(s) ds$ (which is $\approx 0$ when
$z \gg 0$ or $z \ll 0$). The function's minimum is $0$, but the minimum value of expectation is $\sigma \sqrt{2 / \pi}$.
The derivatives is 
\begin{equation}
\frac{\partial}{\partial z} \E[\dur(z;s)] = 2 \F(z)-1, 
\end{equation}
which again has three regimes: $-1$ for $z \ll 0$, $+1$ for $z \gg 0$, and a transitional regime near $0$. Thus, the expected URDF has distance-function-like properties away from the intersection.

\parnobf{The second intersection} Like the SRDF, accounting for the second intersection leads to a more complex expression. The expected second intersection also includes
the terms
\begin{equation}
(n-2z) \F\left(z-\frac{n}{2}\right) +\int_{-\infty}^{z-\frac{n}{2}} s p(s) ds,
\end{equation}
which are negligible near $0$ and produce distortion at the half-way point $n/2$.

\parnobf{Finding intersections} Finding the intersection is challenging again
due to how $\sigma$ substantially alters the function at the minimum.
Thresholding is challenging because the minimum value is uncertainty-dependent;
searching for where the gradient approaches zero is difficult because the expected
value is substantially more blunted.

\subsection{(Proximity) Occupancy Ray Function (ORF)}
A traditional occupancy function (i.e., inside positive, outside negative) is impossible to train on non-watertight meshes. One can instead train an occupancy network to represent the presence of surface.
The occupancy ray function (ORF) is:
\begin{equation} 
\dor(z;s) = \oneB_{\{x:|x-s|<r\}}(z).
\end{equation}
Its expectation is the fraction of the density within a radius $r$ of $z$, or
\begin{equation}
\E[\dor(z;s)] = \F(z+r) - \F(z-r),
\end{equation}
which has a peak at $z=0$. 

\parnobf{Finding intersections} Finding the intersection is challenging due to the strong interaction between the radius $r$ and the the uncertainty $\sigma$. For instance, suppose one looks for when the occupancy probability crosses a threshold $\tau$ (e.g., $\tau = 0.5$). This crossing may never happen, since if $r=\frac{1}{2}\sigma$, then $\max_z \E[\dor(z;s)] \approx 0.38$. Moreover, setting a global threshold is difficult: the distance from which the $\tau$-crossing is from the true peak depends entirely on $\sigma$. On the other hand, looking for a peak is also challenging: if $r > 3\sigma$, then many values are near-one and roughly equal, since $(z+r,z-r)$ will cover the bulk of the density for many $z$; if $r < \sigma$, then the peak's magnitude is less than one.

\begin{figure}[t]
\centering
\includegraphics[width=0.6\linewidth]{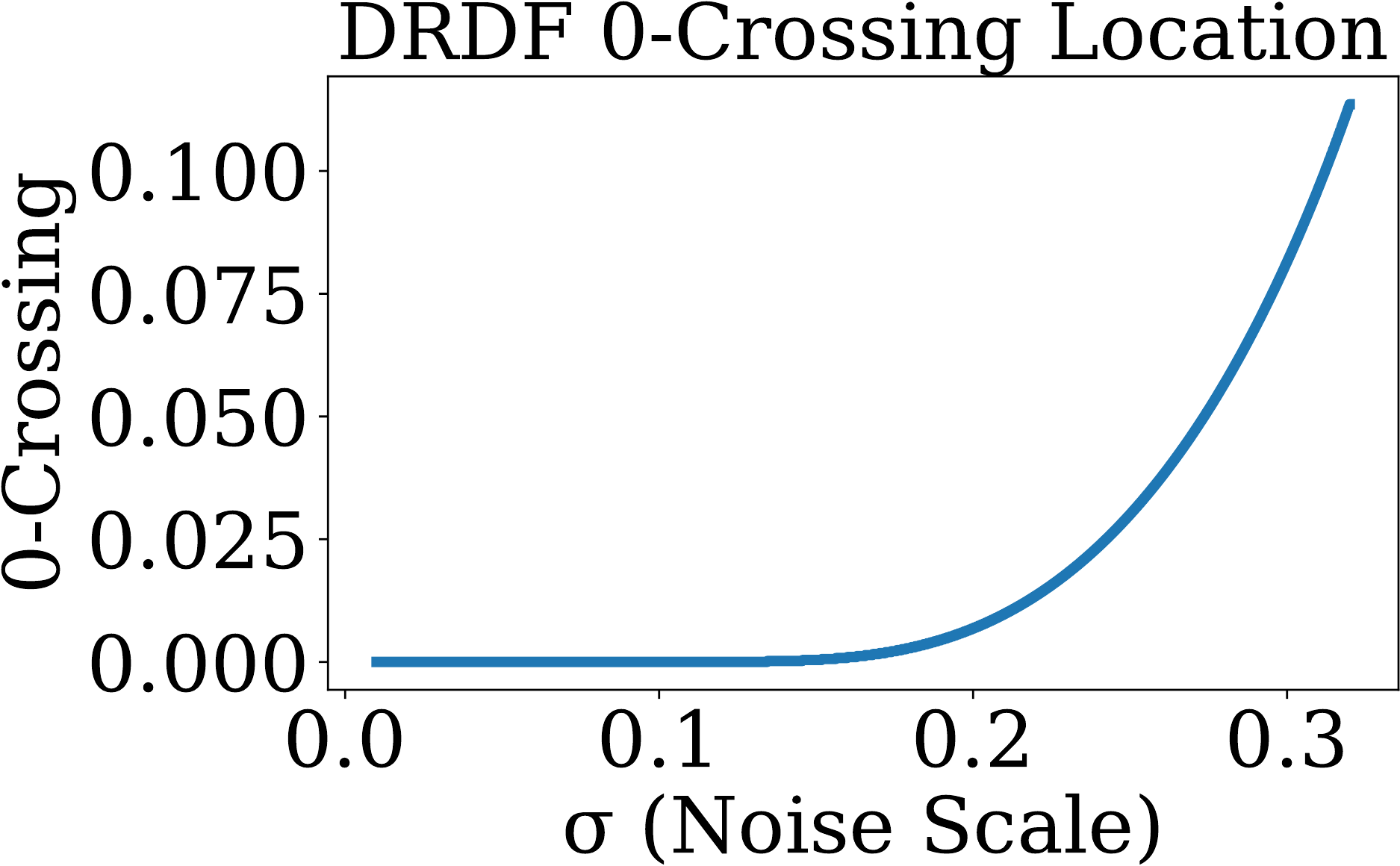}
    \caption{Location of the zero crossing of the DRDF as a function of $\sigma$ (for $n=1$; this scales with $n$). The zero-crossing is virtually at zero until $\sigma$ becomes a substantial fraction of the distance to the next intersection. The smallest $\sigma$ for which the zero-crossing location exceeds 0.01 (i.e., 1cm if $n=1$m) is 0.21; for 0.05, it is 0.27.}
    \label{fig:drdfzerocross}
\end{figure}

\subsection{Directed Ray Distance Function}
We show the equation of directed ray distance function in case of two intersections. This is special case of our general equation presented in the main paper that calculates the DRDF for any number of intersections. 
The directed ray distance function in case of two intersections at $s$ and $n$ is.
\begin{equation}
\label{eqn:dtsdf}
\ddr(z;s) = 
\begin{cases}
s-z &: z\le n/2+s \\
n+s-z &: z > n/2+s. \\
\end{cases}
\end{equation}
Despite the complexity of the function, the expectation is relatively a straightforward
\begin{equation}
\E[\ddr(z;s)] = n \F\left(z-\frac{n}{2}\right) - z, 
\end{equation}
which can be seen to have three regimes: $-z$ when $z \ll n/2$, $n-z$ when $z \gg n/2$,
and a transition near $n/2$. These regimes are traded off by whether $\F(z-\frac{n}{2})$ is $0$, $1$, or something in between. Moreover, so long as $p(z-n/2) \approx 0$ (i.e. the uncertainty is smaller than the distance to the next intersection), the function has a zero-crossing at ${\approx}0$ -- note that if one sets $n=1$ (fixing the scale) the value at $z=0$ is $\F(-1/2)$. The derivative of the expected distance function is
\begin{equation}
\frac{\partial}{\partial z} \E[\ddr(z;s)] = n p\left(z-\frac{n}{2}\right) - 1, 
\end{equation}
which again has two regimes: $-1$ when $p(z-\frac{n}{2}) \approx 0$, which happens
when $z$ is far from $\frac{n}{2}$, which in turn happens for $z \ll n/2$ and $z \gg n/2$; and
a transitional regime near $n/2$, where the derivative is not $-1$. 

The location of the zero-crossing is controlled by $\sigma$. For most $\sigma$ of interest, the zero-crossing is nearly at zero. This value can be computed as the $\hat{z}$ such that $n \F(\hat{z}-\frac{n}{2}) - \hat{z} = 0$. We plot the location of the zero-crossing $\hat{z}$ as a function of $\sigma$ in Fig.~\ref{fig:drdfzerocross}, assuming $n=1$ (note that $n$ scales with $\sigma$). $\hat{z}$ first crosses $0.01$ (i.e., 1cm error) when $\sigma = 0.21$, or when the standard deviation of the uncertainty about surface location is $20\%$ of the distance to the next intersection. The DRDF does break down at for large $\sigma$ (e.g., $\sigma = 0.3$, where it is off by ${\approx}0.1$.

\parnobf{Finding intersections} Finding the intersection is made substantially easy because the uncertainty-dependent contortions of the function are pushed elsewhere. The discontinuity at $n/2$ does create a phantom zero-crossing, but this is easily recognized as a transition from negative to positive.

\subsection{Median vs Expectation}
\label{sec:analysis_median}
All the analysis presented in the above sections has been under the assumption for networks trained with L2 loss, and the above analysis also holds for networks trained with L1 loss as in case of random variable under symmetric distribution about the mean the analysis follows as is. The above analysis is in terms of the expected distance function since this is easiest to derive. However, the median distance function closely tracks the expected distance function for the distance functions we study. 

\begin{figure}[t]
\centering
\begin{tabular}{cc}
\includegraphics[width=0.45\linewidth]{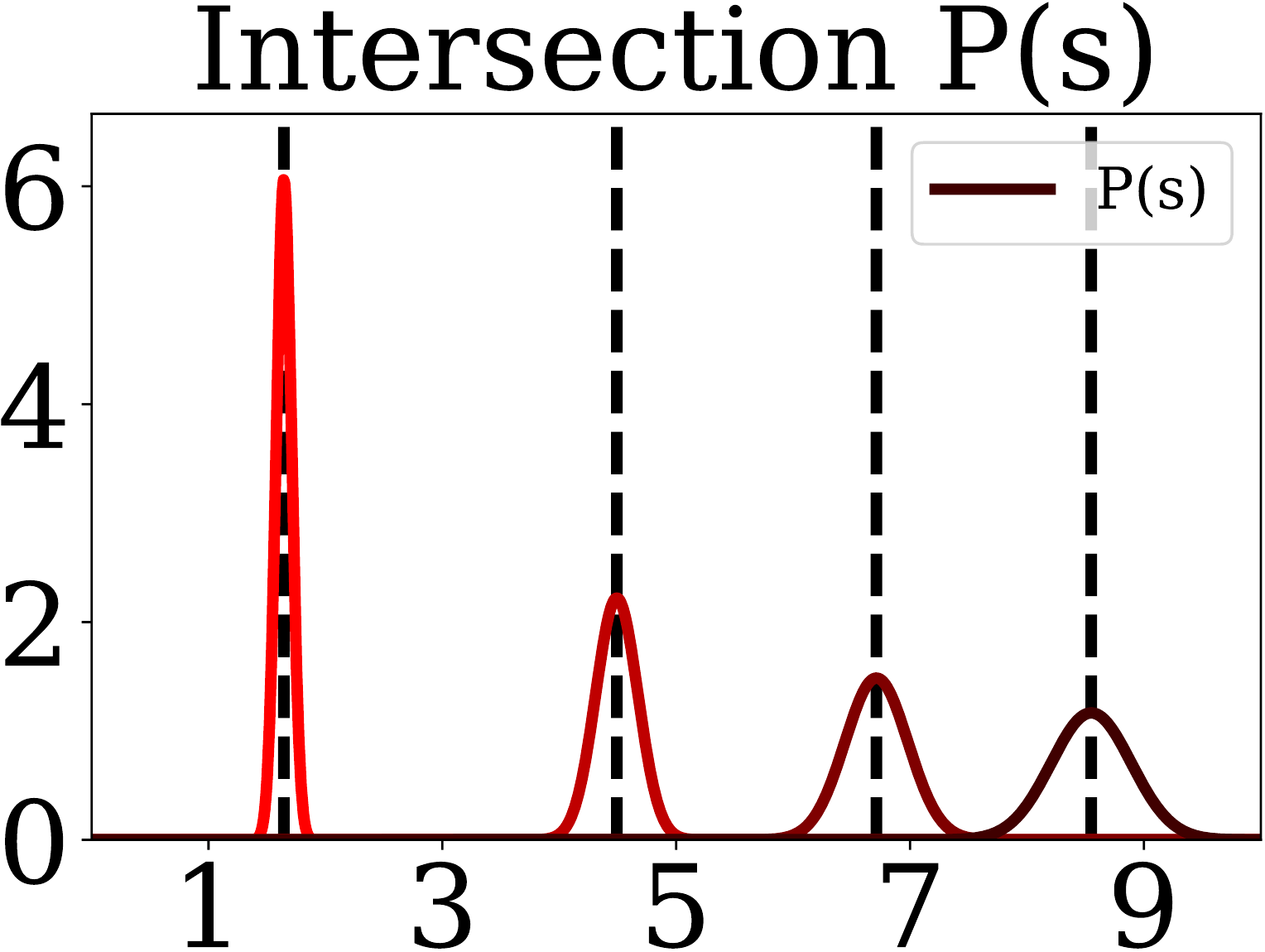} &
\includegraphics[width=0.45\linewidth]{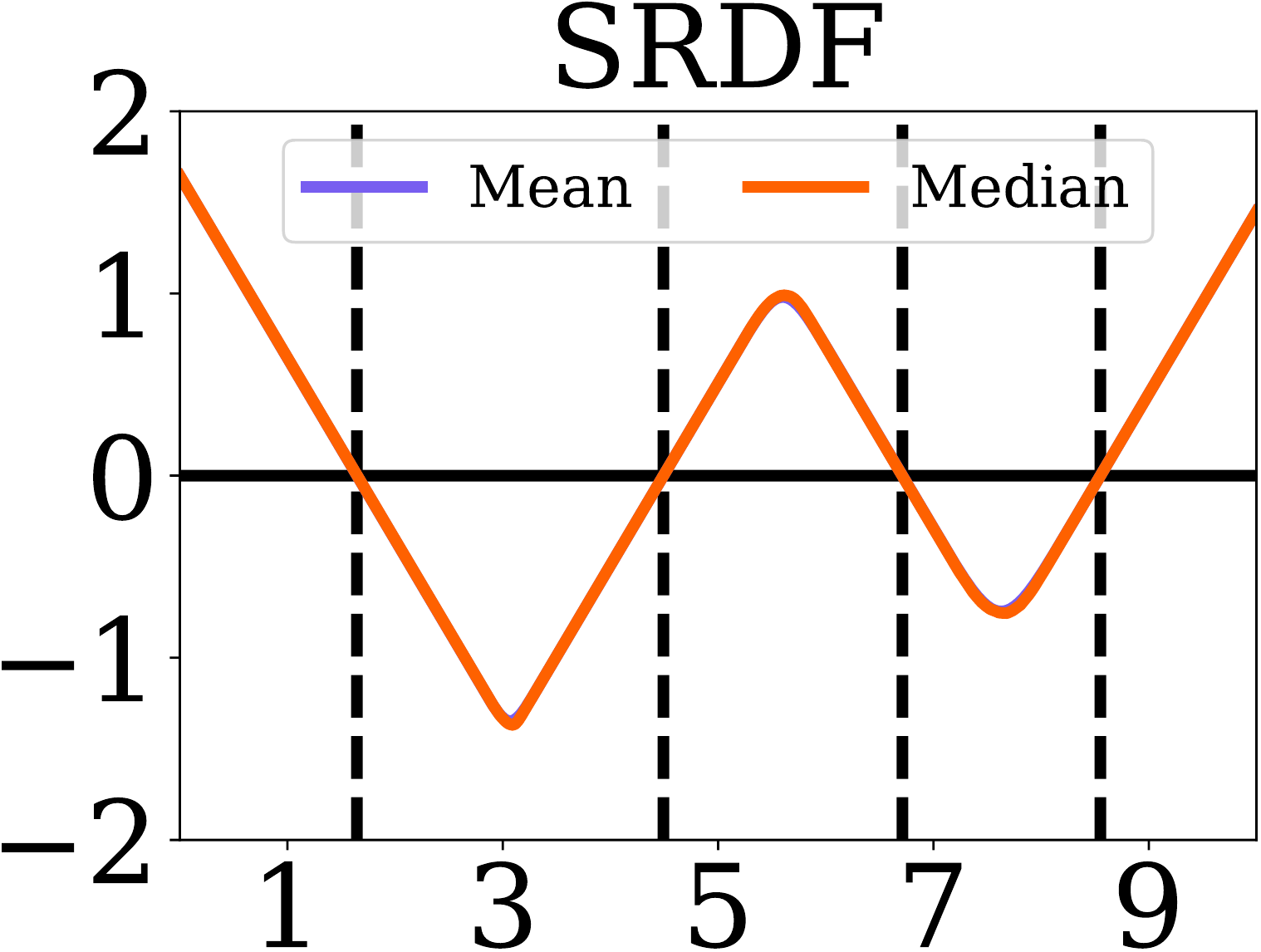} \\ 
\includegraphics[width=0.45\linewidth]{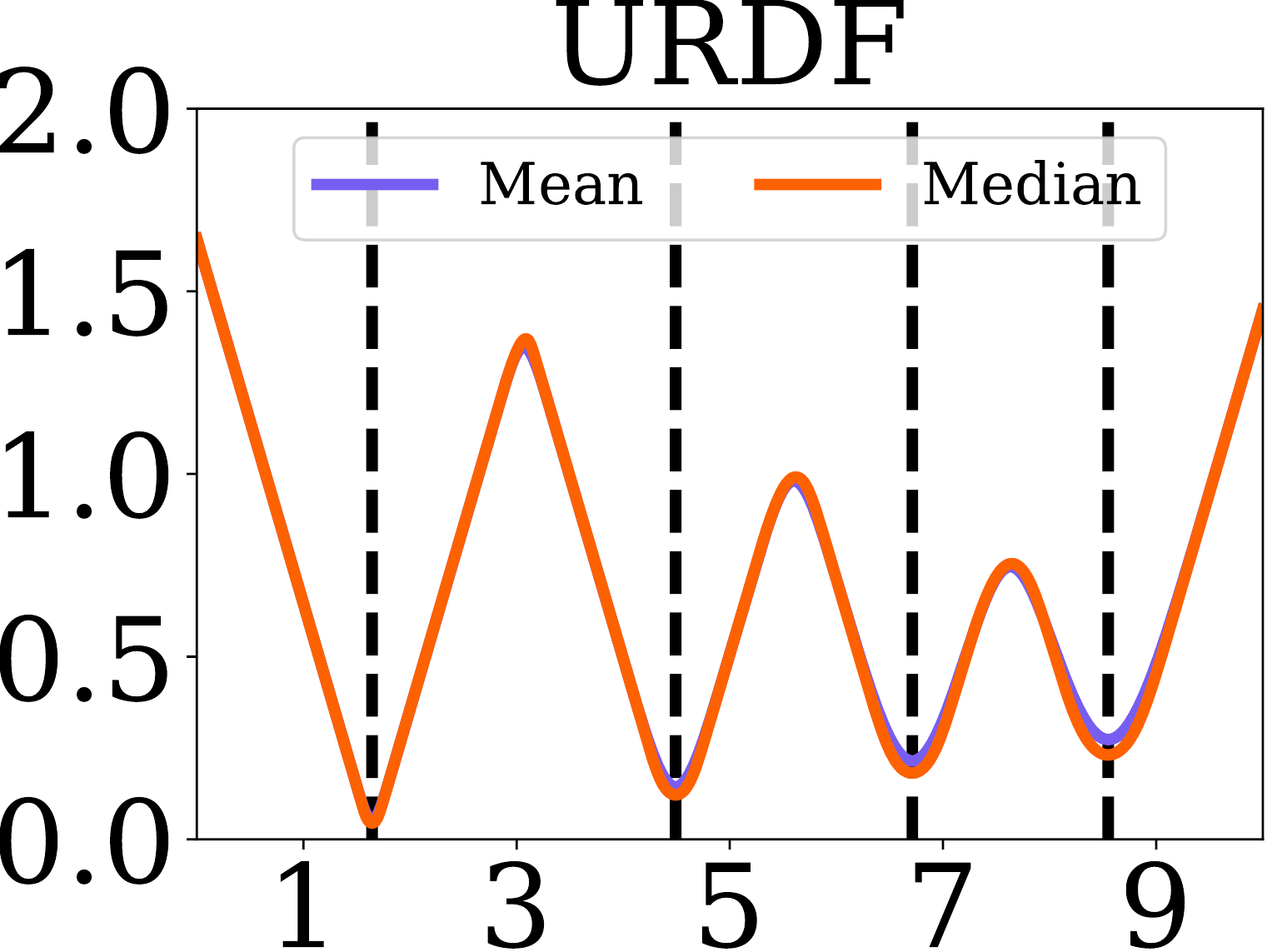} &
\includegraphics[width=0.45\linewidth]{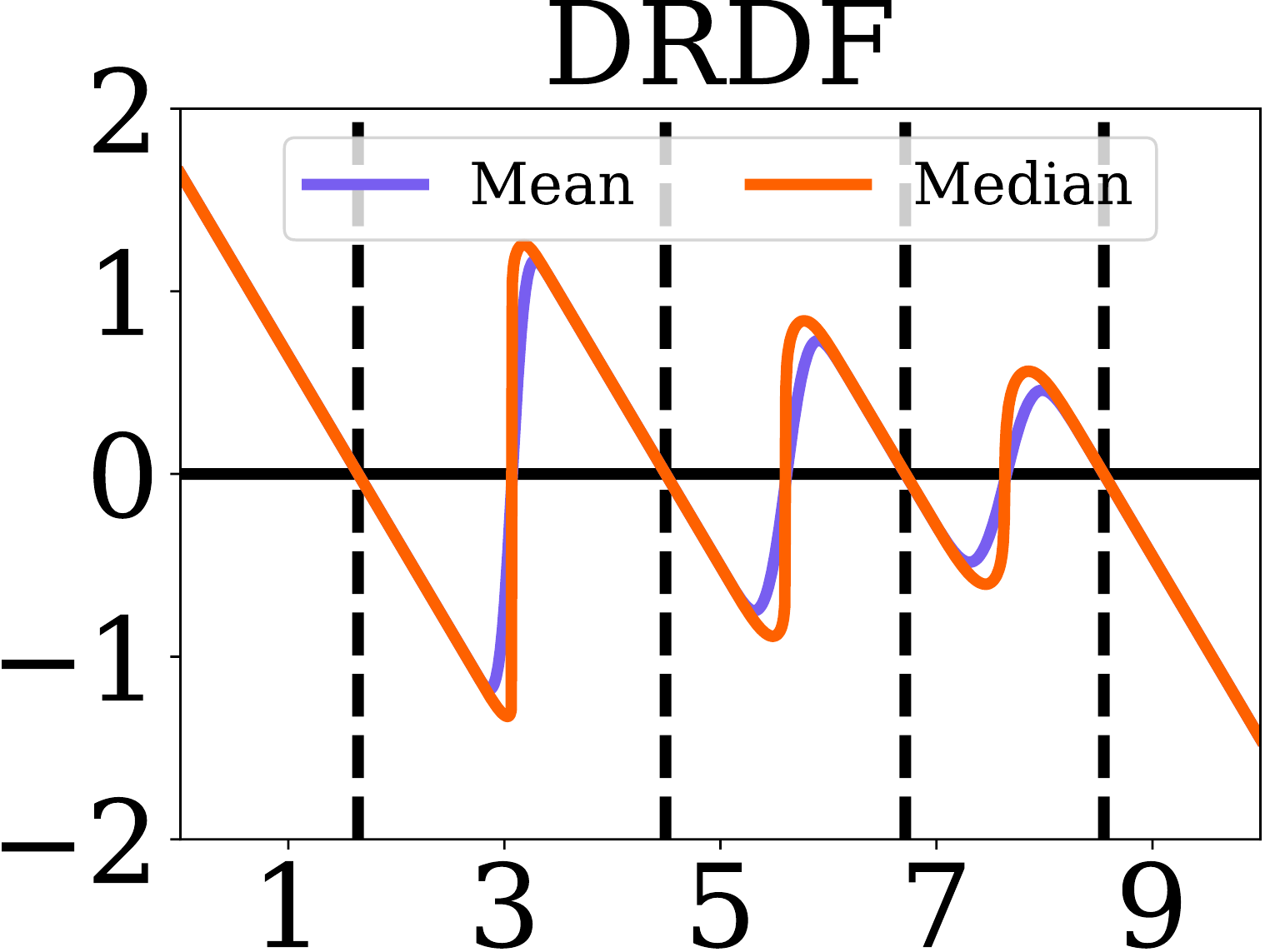} \\
\end{tabular}
\caption{Median-vs-Expectation/Mean for UDF, SDF, and DRDF. We sample intersections from a set of per-intersection distributions (top left). We then compute at each point along the ray, the mean and median distance function. The mean and median are virtually identical apart from a slight shift in the UDF minimums, and sharpening in the DRDF near the discontinuity.}
\label{fig:median}
\end{figure}

\parnobf{Empirical results} Empirically, the results for the median are
virtually identical. We sample intersections independently from the
distributions shown in Fig.~\ref{fig:median}, where the variance is
depth-dependent. We then numerically calculate the expectation/mean and the
median over $1$M samples from this distribution. The plots are virtually
identical. Two small differences are visible: the median URDF is slightly
smaller than the mean URDF near intersections, and the median DRDF more closely
resembles the ground-truth DRDF by better capturing the discontinuity. We
do not plot the ORF since cross-entropy training minimizes the mean.

\parnobf{Analysis}
These empirical results occur because if we treat the distance function at a
location $z$ as a random variable, then the mean and median are similar. More
specifically, for a fixed $z$, if we plug in the random variable $S$ into the
distance function $d(z;S)$, we can analyze a new random variable for the
distance to the surface at location $z$. For instance, the SRDF at location $z$
is $S-z$ if we ignore the second intersection; in turn, $S-z$ is normally
distributed with mean $-z$. The mean and median are identical for the normal
since it is symmetric.

\parnoit{URDF}
A more involved case is the URDF.  The URDF at location $z$ is $|S-z|$, which
is a folded normal with mean $-z$ and standard deviation $\sigma$ (with the
$\sigma$ inherited from the uncertainty about the intersection location).
We'll focus on bounding the gap between the mean and median value at each
location.

To the best of our knowledge, there is no closed form expression for the median, and 
Chebyshev gives vacuous bounds, and so we therefore compute it numerically
(using the fact the the median of the folded normal is the $m$ such that
$\F_z(m)=0.5+\F_z(-m)$ where $\F_z$ is the CDF of a normal centered at $z$).
Note that when $z$ is far from the intersection, the folded normal and normal
are virtually identical -- close to none of the normal's density is on negative
values. In general, one can bound the gap between mean and
median by numerical search over different possible values for $z$. For $\sigma=1$, the largest difference is
${\approx}0.135$. Changing $\sigma$ scales this: $\sigma=0.5$ yields
${\approx}0.067$. Note that the minimum is $\sigma \sqrt{2/\pi} = 0.797\sigma$,
so for $\sigma = 1$, the median's minima is $\ge0.663$. Thus, in general, the median has to be quite close to the mean.

\parnoit{DRDF}
The DRDF has a larger distortion near $\frac{n}{2}$ because the random variable near $\frac{n}{2}$ is bimodal. The mean splits the difference between the modes while the median sharply transitions depending on which mode is more likely. This discrepancy, however, occurs far from the intersection and is therefore not of concern. For $z$ near the intersection, the resulting random variable resembles a normal distribution.

\subsection{Unsigned Distance Function to A Plane in 3D}
\label{sec:deriv_plane}

Suppose we are given a plane consisting of a normal $\nB \in \mathbb{R}^3$ with $||\nB||_2 = 1$ and offset $o$ (where points $\xB$ on the plane satisfy $\nB^T \xB + o = 0$). Then $d_{U}(\xB;\nB,o)$ is the unsigned distance function (UDF) to the plane, or 
\begin{equation}
d_{\textrm{U}}(\xB) = | \nB^T \xB + o |.
\end{equation} 
Suppose there is uncertainty about the plane's location in 3D. Specifically, let us assume that the uncertainty is some added vector $\sB{\sim}N(\zeroB,\sigma^2 \IB)$ where $\IB$ is the identity matrix and $\zeroB$ a vector of zeros. Then the expected UDF at $\xB$  is 
\begin{equation}
E_{\sB}[d_\textrm{U}(\xB;\sB)] = \int_{\mathbb{R}^3} |\nB^T (\xB + \sB) + o|~(\sB)~d\sB.
\end{equation}
This ends up being the expected URDF, but replacing distance with point-plane distance.
Specifically, if $p = |\nB^T \xB + o|$, then
\begin{equation}
E_{\sB}[d_{U}(\xB;\sB)] = p \F(p) - p (1-\F(p)) + 2 \int_{p}^{\infty} s p(s) ds.
\end{equation}
Thus, the minimum value remains $\sigma \sqrt{2/\pi}$. 

One nuance is that the rate at which $p$ changes is different for different rays through the scene and is proportional to the cosine between the normal $\nB$ and the ray. So the function is stretched along its domain.

\clearpage

\section{Qualitative Results}
\begin{figure*}[h]
    \centering
    \noindent\includegraphics[width=\textwidth]{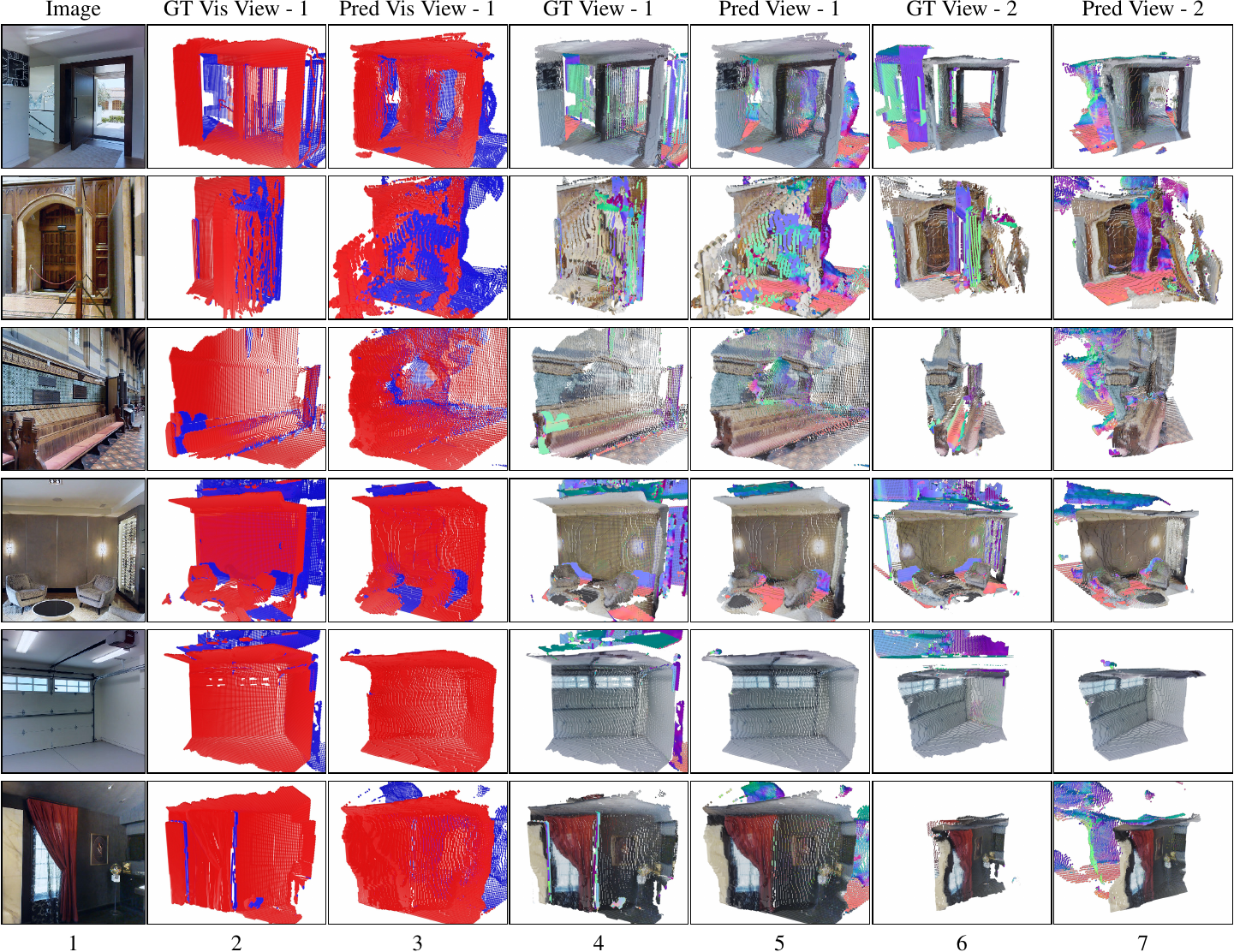} 
   \caption{{\bf Matterport Novel Views} We \emph{randomly sample} examples from the test set and show results and show generated 3D outputs in a new view for them in novel view. Video results for Row 1 and Row 4 are present in the \texttt{supp\_qual.mp4}. Cols 2 \& 3 show regions in \textcolor{red}{red as visible} in camera view and \textcolor{blue}{blue as occluded} in camera view. We colors the \textcolor{red}{visible regions} with image textures and the \textcolor{blue}{occluded regions} with surface normals(\protect\includegraphics[height=8pt,width=8pt]{figures/normal_map2.png}, scheme from camera inside a cube) in Col 4, 5, 6, 7. We observe that our model is able to recover parts of the occluded scene shown in blue and colored with surface normal map; floor behind the wall(row 1), walls and floor behind couch (row 4).}
    \label{fig:matterport_novel}
    \vspace{-3mm}
\end{figure*}

\begin{figure*}[h]
    \centering
    \noindent\includegraphics[width=\textwidth]{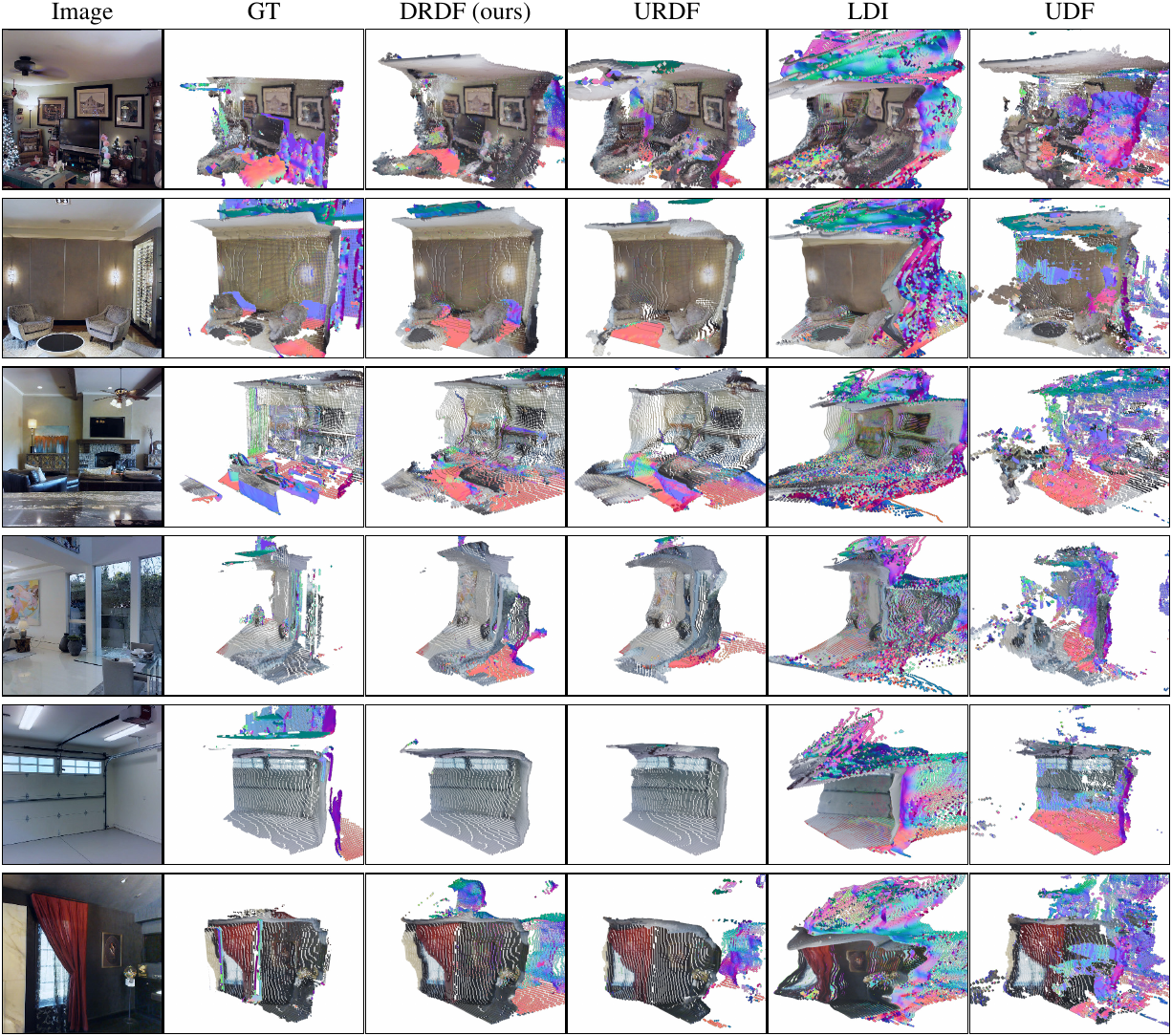} 
   \caption{{\bf Matterport Comparison with Baselines} We \emph{randomly sample} samples from the test set and show results comparing DRDF to other baselines. DRDF shows consistently better results as compared to \scened and \mpd. Both \scened and \mpd have blobs and inconsistent surfaces in output spaces (all rows). \urdf always is unable to recover hidden regions (row 3 behind the couch on the right), \urdf is missing the floor on lower right (row 4) as compared to \tsdf.}
    \label{fig:matterport_comp}
    \vspace{-3mm}
\end{figure*}

\begin{figure*}[h]
    \centering
    \noindent\includegraphics[width=\textwidth]{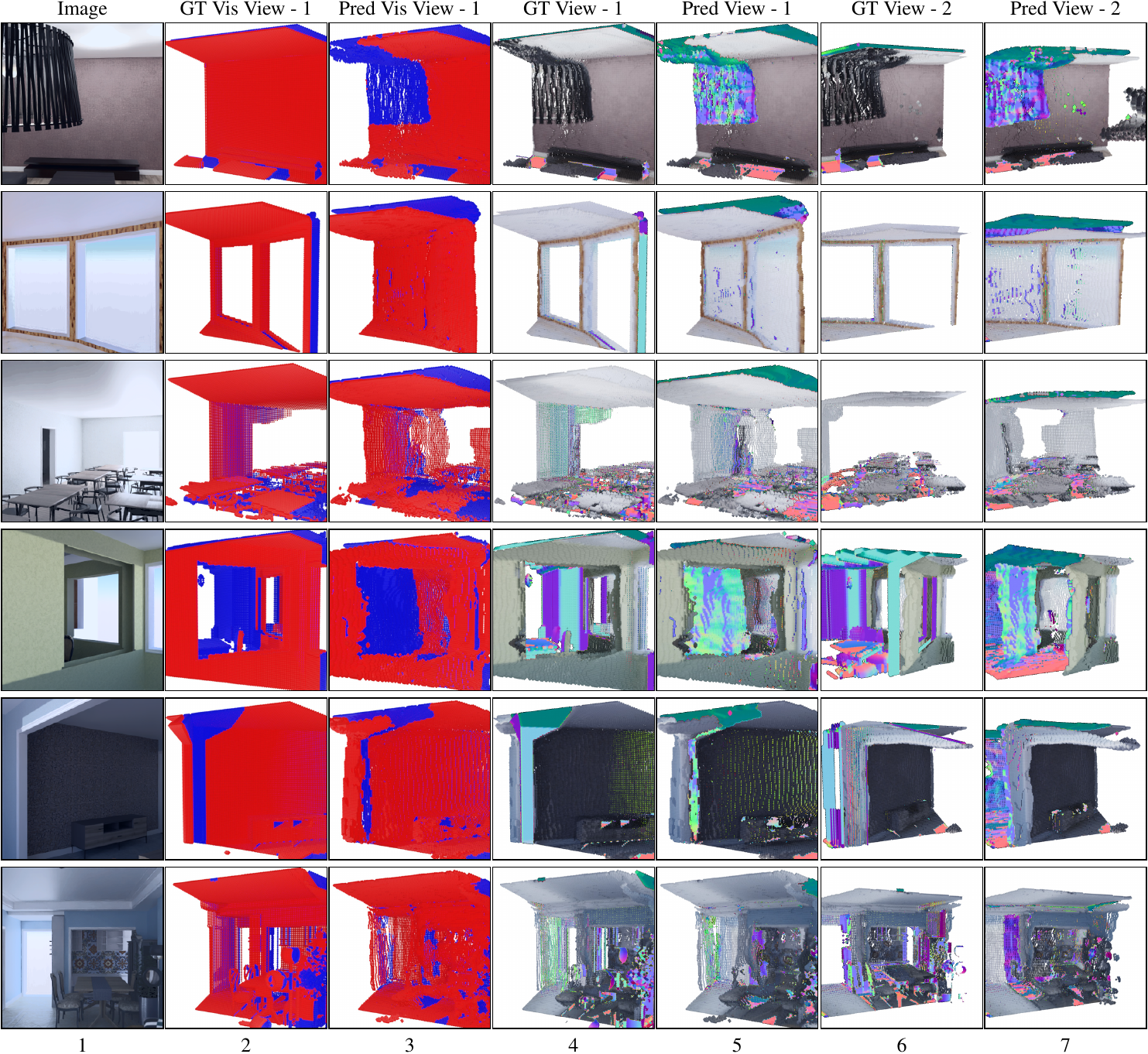} 
   \caption{{\bf 3DFront Novel Views} We \emph{randomly sample} examples from the \tdf test set and show results. Video results are available for row 2, 6 in the \texttt{supp\_qual.mp4}. We observe our model recovers portion of floor occluded by the table (row 2, bottom right of the image) ; our model is also able to identify small occluded regions in a complicated scene (row 6)}
    \label{fig:threedf_novel}
    \vspace{-3mm}
\end{figure*}

\begin{figure*}[h]
    \centering
    \noindent\includegraphics[width=\textwidth]{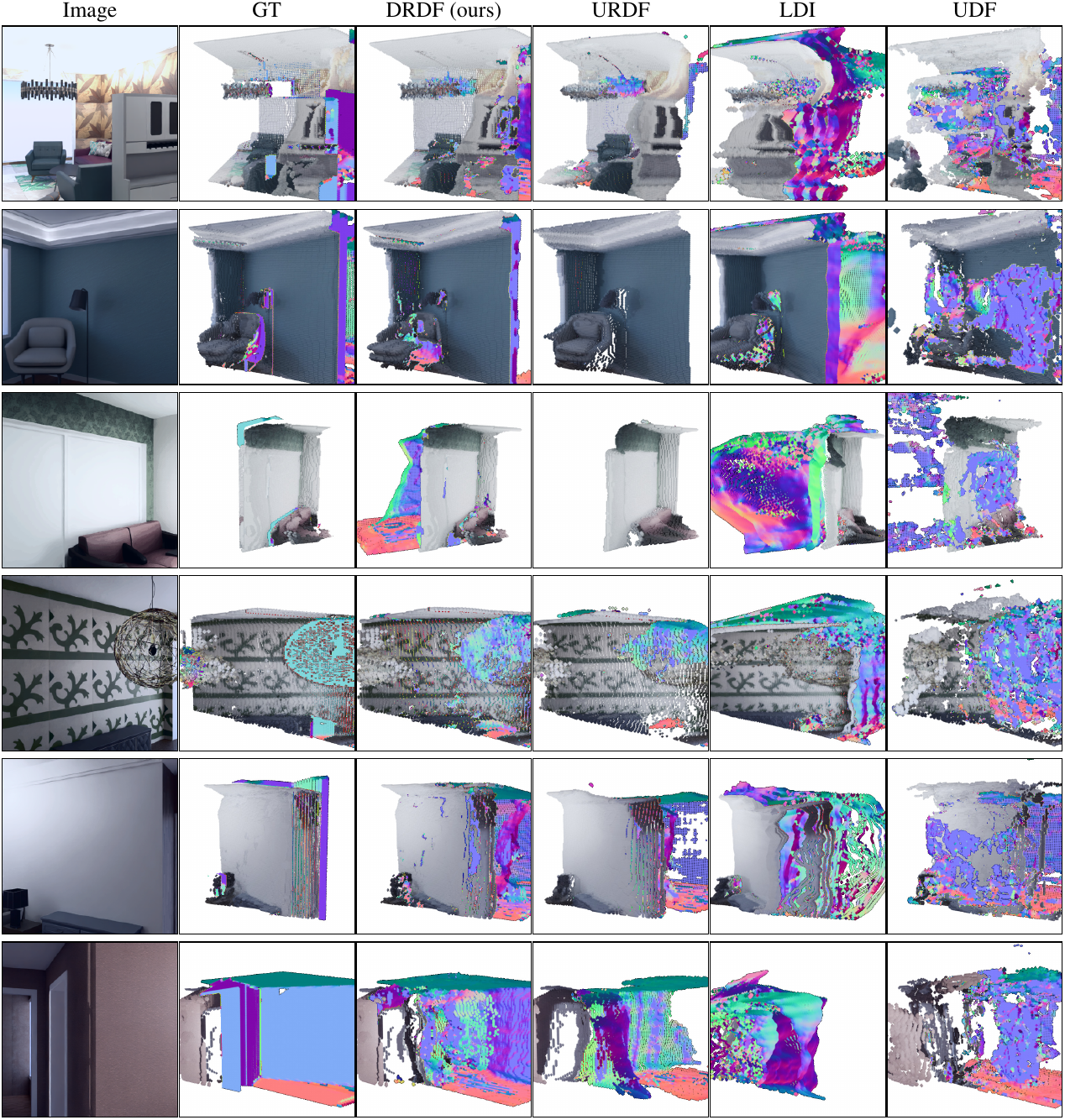} 
   \caption{{\bf 3DFront Comparison with Baselines}  We \emph{randomly sample} samples from the test set and show results comparing DRDF to other baselines. DRDF shows consistently better results as compared to \scened and \mpd. Both \scened and \mpd have blobs and inconsistent surfaces in output spaces (all rows). \urdf always is unable to recover hidden regions (row 2 behind the couch on the right) while \tsdf does. \tsdf also speculates another room in the scene (row 3, 5)}
    \label{fig:threedf_comp}
    \vspace{-3mm}
\end{figure*}

\begin{figure*}[h]
    \centering
    \noindent\includegraphics[width=\textwidth]{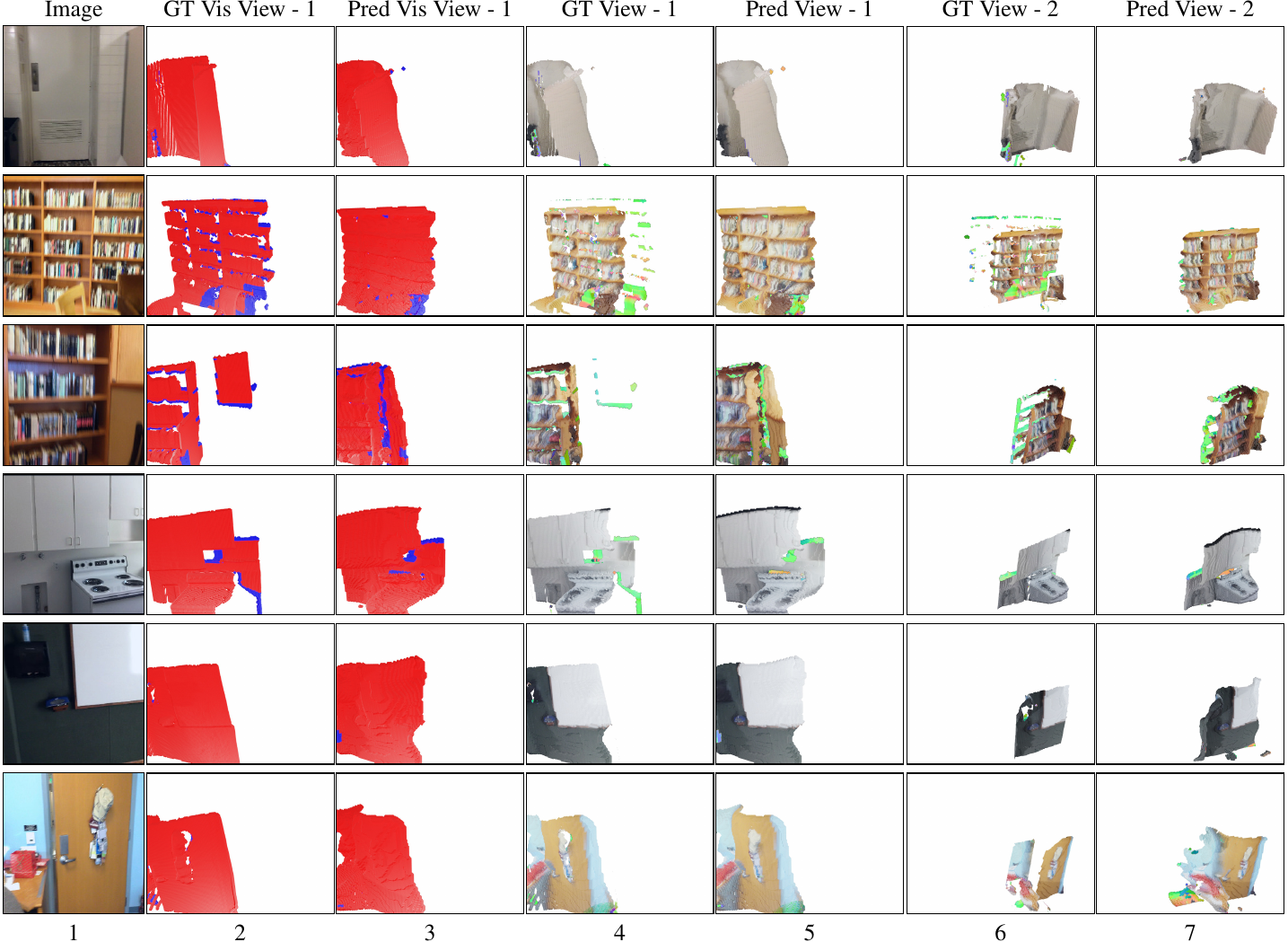} 
   \caption{{\bf ScanNet Novel Views}  We \emph{randomly sample} examples from the \scannet test set and show results. Video results are available for row 2, 4 in the \texttt{supp\_qual.mp4}. We observe our model recovers portion of wall occluded by the chair (row 2, bottom right of the image View 1) ; ScanNet does not have lot of occluded surfaces as we can see from ground-truth and hence a lot of regions in novel views are visible in the camera view.}
    \label{fig:scannet_novel}
    \vspace{-3mm}
\end{figure*}

\begin{figure*}[h]
    \centering
    \noindent\includegraphics[width=\textwidth]{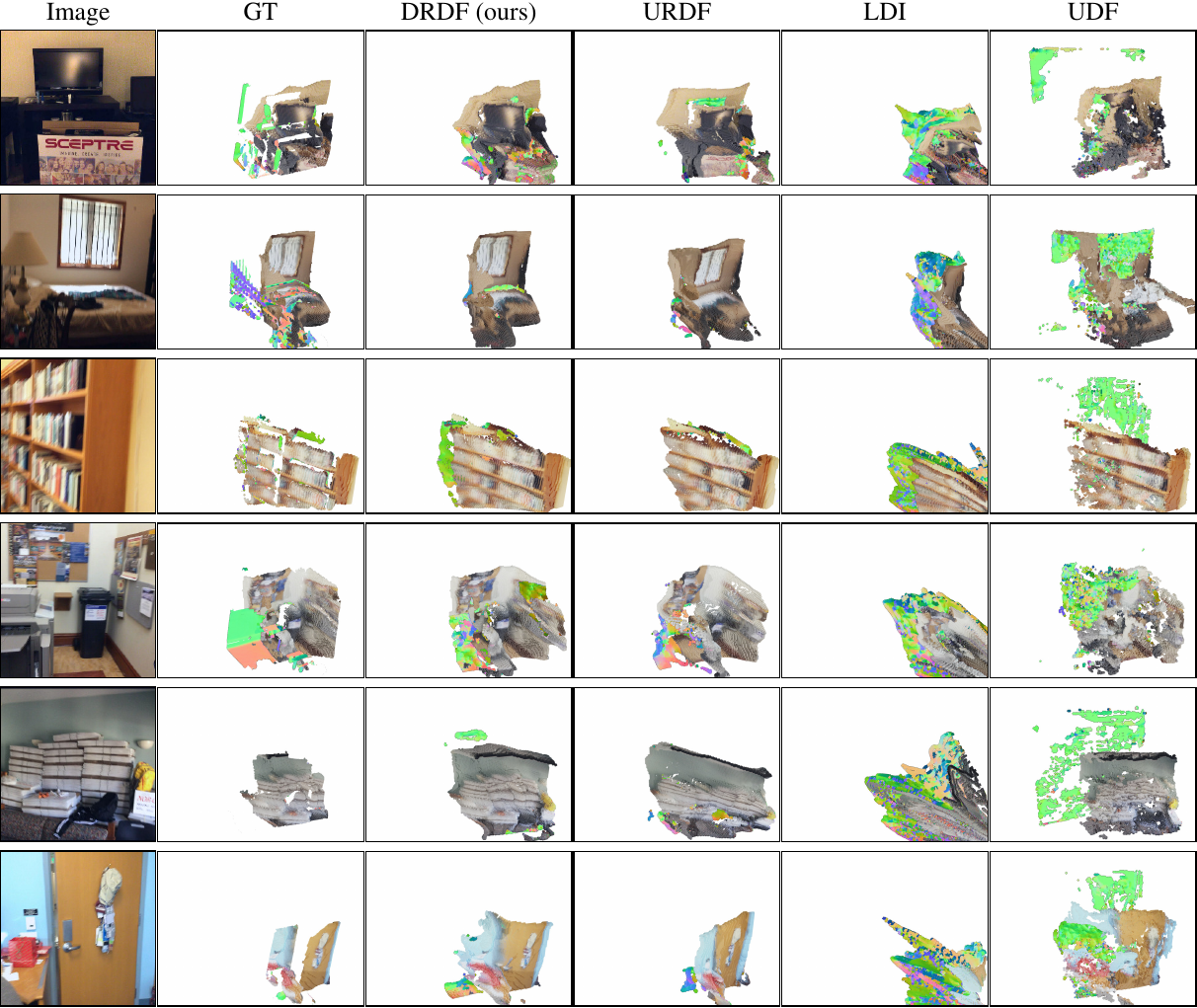} 
   \caption{{\bf ScanNet Comparison with Baselines} We \emph{randomly sample} samples from the test set and show results comparing DRDF to other baselines. DRDF shows consistently better results as compared to \scened and \mpd. Both \scened and \mpd have blobs and inconsistent surfaces in output spaces (all rows). \tsdf outputs look more closer to the groud-truth as compared to \urdf (row 4)}
    \label{fig:scannet_comp}
    \vspace{-3mm}
\end{figure*}

We show qualitative results on r\emph{randomly sampled} images from the test set on 3 datasets. In \figref{fig:matterport_novel} and \figref{fig:matterport_comp} we compare on \matterport with respect to ground truth and baselines. Similarly we show results for \tdf on \figref{fig:threedf_novel} and \figref{fig:threedf_comp}. Results for in \scannet in novel views in \figref{fig:scannet_novel} and comparison against baseline is in \figref{fig:scannet_comp}. We show some selected results from these random samples (2 per dataset) in the \texttt{supp\_qual.mp4}. We recommend watching this video.
\label{sec:qual}

\label{sec:tdf_random}

\clearpage \clearpage

\section{Derivations}
\label{sec:deriv}
\newcommand{\ddz}{\frac{\partial}{\partial z}}

For completeness, we show the derivation of some of the results presented in \S\ref{sec:analysis}. This is meant to help in verifying the solutions or deriving the solution for another function.  Assume $S \sim N(0,\sigma)$ with density $p(s)$ and CDF $\F(s)$.

Useful identities:
\begin{enumerate}
\item 
$\int_{-\infty}^{\infty} s p(s) ds = 0$~~~($\E[s] = 0$).
\item $c \int_{-\infty}^{\infty} p(s) ds = c$~~~(Total probability is $1$)
\item $\int_{-\infty}^{a} s p(s) ds = \int_{-\infty}^{\infty} s p(s) ds - 
\int_{a}^{\infty} s p(s) ds$ 
\item $\int_{-\infty}^{a} s p(s) ds = -\int_{a}^{\infty} s p(s) ds$ (since $\int_{-\infty}^{\infty} s p(s) ds = 0$).
\item $\int_{-\infty}^{a} = \F(a)$
\item $\int_{a}^{\infty} = (1-\F(a))$
\end{enumerate}

\subsection{Signed Ray Distance Function}
\label{sec:deriv_sdf}

We will start with 
a signed ray distance function. The expected
signed ray distance function is
\begin{equation}
\E[\dsr(z;s)] = \int_{\mathbb{R}} (s-z) p(s) ds, 
\end{equation}
which can be rewritten as 
\begin{equation}
\int_{\mathbb{R}} s p(s) ds - \int_{\mathbb{R}} z p(s) ds.
\end{equation}
The first term is the expected value of $S$, or $0$. The second term is
$-z \int_{\mathbb{R}} p(s) ds$. Since $\int p(s) ds = 1$, this reduces
to $-z$. This yields 
\begin{equation}
\E[\dsr(z;s)] = -z.
\end{equation} 
Alternately, one can recognize $S-z$ as normally distributed with mean $-z$, 
which has a mean of $-z$.

\parnobf{General form}
The more general form of $\dsr$ that accounts for the second intersection is 
done in two cases:
\begin{equation}
\dsrp(z;s) = 
\begin{cases}
\textcolor{blue}{s-z} &: \textcolor{blue}{z < s + n/2} \\
\textcolor{red}{z-n-s} &: \textcolor{red}{z \ge s + n/2}. \\
\end{cases}
\end{equation}
The expectation $\E[\dsrp(z;s)] = $ can be computed in two parts
\begin{equation}
\textcolor{red}{\int_{-\infty}^{z-\frac{n}{2}} (z-n-s) p(s) ds} +
\textcolor{blue}{\int_{z-\frac{n}{2}}^{\infty} (s-z) p(s) ds}.
\end{equation}
For notational cleanliness, let $t = z-\frac{n}{2}$, and pull
out constants and re-express any integrals as CDFs. Then 
the first integral expands to 
$\textcolor{red}{z \F(t) - n \F(t) - \int_{-\infty}^{t} s p(s) ds}$,
and the second integral expands to
$\textcolor{blue}{\int_{t}^{\infty} s p(s) ds - z (1-\F(t))}$. 
We can then group and apply $\int_{-\infty}^{t} s p(s) ds = - \int_{t}^{\infty} s p(s) ds$, which
yields $(2z-n)\F(t) - z + 2\int_{t}^{\infty} s p(s) ds$.
A little re-arranging, and expanding out $t$ yields:
\begin{equation}
\label{eqn:edsp}
\begin{split}
\E[\dsrp](z;s) = & -z + (2z-n) \F\left(z-\frac{n}{2}\right) +  \\
& 2 \int_{z-\frac{n}{2}}^\infty s p(s) ds. \\
\end{split}
\end{equation}
This expression has $\E[\dsr]$ in it ($-z$), plus terms (all but the first one)
that activate once $z$ approaches $\frac{n}{2}$.

\subsection{Unsigned Ray Distance Function}
\label{sec:deriv_udf}

The expected
unsigned ray distance function is
\begin{equation}
\E[\dur(z;s)] = \int_{\mathbb{R}} |s-z| p(s) ds.
\end{equation}

Before calculating it in general, we can quickly check what value the expected distance function 
takes on at the actual intersection by plugging in $z=0$, or
\begin{equation}
\E[\dur(0;s)] = \int_{\mathbb{R}} |s| p(s) ds.
\end{equation}
This integral evaluates to $\sigma \sqrt{2/\pi}$, which can be
quickly obtained by noting that it is the expected value of
a half-normal distribution. Indeed, the distribution over $\dur(z;S)$ is a folded normal
distribution with mean $-z$. 

We can then derive the more general form, by calculating the integral in two parts:
$\E[\dur(z;s)]$ is 
\begin{equation}
\textcolor{red}{\int_{-\infty}^{z} (z-s) p(s) ds} + 
\textcolor{blue}{\int_{z}^{\infty} (s-z) p(s) ds}.
\end{equation}
We can expand and shuffle to yield
\begin{multline}
\textcolor{red}{z \int_{-\infty}^z p(s) ds} \textcolor{blue}{-z \int_{z}^{\infty} p(s) ds} + \\
\textcolor{blue}{\int_{z}^{\infty} s p(s) ds} \textcolor{red}{- \int_{-\infty}^z s p(s) ds}. 
\end{multline}
The first two terms can be written in terms of the CDF $\F$, and the last term can be further simplified by noting
$\int_{-\infty}^{z} s p(s) ds = - \int_{z}^{\infty} s p(s) ds$.  This yields a final form
for $\E[\dur(z;s)]$,
\begin{equation}
\label{eqn:expdu}
 z \F(z) -z (1-\F(z)) + 2 \int_{z}^{\infty} s p(s) ds.
\end{equation}
As seen before, when $z$ is zero, the result is $\sigma \sqrt{2/\pi}$, which
is the minimum. For $z \ll 0$, both $\F(z)$ and the integral can be ignored, leading
a value of ${\approx} -z$. Symmetrically, the value is ${\approx} z$ if $z \gg 0$.
Near zero, the function is more complex.

\parnobf{Derivative} The derivative of Equation~\ref{eqn:expdu} can be calculated out in three parts
\begin{equation}
\begin{split}
\ddz z \F(z) & = z p(z) + \F(z)
\\
\ddz z (1-\F(z)) & = zp(z) - \F(z) + 1
\\
\ddz 2 \int_{z}^{\infty} s p(s) ds & = -2z p(z) 
\\
\end{split}
\end{equation}
Adding the first, subtracting the second, and adding the third yields the final 
result:
\begin{equation}
\frac{\partial}{\partial z} \E[\dur(z;s)] = 2 \F(z) - 1. 
\end{equation}
In the tails, $\F(z)$ splits to $0$ and $1$, and thus $\frac{\partial}{\partial z} \E[\dur(z;s)]$ splits to $-1$ and $1$. When $z$ is not in the tail, the derivatives are not one.

\parnobf{General Form} The more general form of $\dur$ that accounts for the second intersection is
\begin{equation}
\durp(z;s) = 
\begin{cases}
\textcolor{red}{s-z} &: \textcolor{red}{z < s} \\
\textcolor{blue}{z-s} &: \textcolor{blue}{z > s, z-\frac{n}{2} < s} \\
\textcolor{purple}{n-z} &: \textcolor{purple}{z-\frac{n}{2} > s.} \\
\end{cases}
\end{equation}
This can be computed in three parts. Again, let $t = z-\frac{n}{2}$ to reduce notational clutter. Then $\E[\durp(z;s)]$ is
\begin{equation}
\begin{split}
& \textcolor{purple}{\int_{-\infty}^{t} (n-z) p(s) ds} + \\
& \textcolor{blue}{\int_{t}^{z} (z-s) p(s) ds} + \\
& \textcolor{red}{\int_{z}^{\infty} (s-z) p(s) ds}.
\end{split}
\end{equation}
As usual, we pull out constants and rewrite integrals in terms of the CDF or 1 minus the CDF.  This yields
\begin{equation}
\begin{split}
& \textcolor{purple}{n\F(t) - z\F(t)} + \\
& \textcolor{blue}{z (\F(z)-\F(t)) - \int_{t}^{z} s p(s) ds} + \\ 
& \textcolor{red}{\int_{z}^{\infty} s p(s) ds - z (1-\F(z))}.
\end{split}
\end{equation}
If we gather terms involving $\F(t)$ and $\F(z)$, as well as the integrals, we get
\begin{equation}
\begin{split}
& (n-2z) \F(t) + 2 z \F(z) - z + \\
& -\int_{t}^{z} s p(s) ds + \int_{z}^{\infty} s p(s) ds.
\end{split}
\end{equation}
The value $-\int_{t}^{z} s p(s) ds = \int_{-\infty}^{t} s p(s) ds + \int_{z}^\infty s p(s) ds$, which lets us rewrite the integrals, yielding 
\begin{equation}
\begin{split}
& (n-2z) \F(t) + \textcolor{orange}{2 z \F(z) - z} + \\
& \textcolor{orange}{2 \int_{z}^{\infty} s p(s) ds} + \int_{-\infty}^{t} s p(s) ds,
\end{split}
\end{equation}
where terms from the original URDF are highlighted in \textcolor{orange}{orange} (note that $z \F(z) - z(1-\F(z)) = 2 \F(z) - z$). Re-arranging, and re-substituting back in $t = z-\frac{n}{2}$ yields
\begin{equation}
\label{eqn:edup}
\begin{split}
& z\F(z) + -z (1-\F(z)) + 2 \int_z^\infty s p(s) ds + \\
& (n-2z) \F\left(z-\frac{n}{2}\right) +\int_{-\infty}^{z-\frac{n}{2}} s p(s) ds. \\
\end{split}
\end{equation}
Again, this is like $\E[\dur]$ but with additional terms (those in the second line) that activate once $z$ approaches
$\frac{n}{2}$. 

\subsection{Occupancy Ray Function}

The standard occupancy function (i.e., positive is interior, negative is exterior) is not defined on non-watertight meshes. We can define an alternate occupancy function which is positive near a surface and negative away from a surface. 

Specifically the expected occupancy function is 
\begin{equation}
\label{eqn:expocc}
\E[\dor(z;s)] = \int_{\mathbb{R}} \oneB_{\{x:|x-s|<r\}}(z) p(s) ds, 
\end{equation}
where $\oneB$ is the indicator function. Equation~\ref{eqn:expocc} can be simplified as
\begin{equation}
\int_{z-r}^{z+r} p(s) ds = \F(z+r) - \F(z-r).
\end{equation}

\subsection{Directed Ray Distance Function}

We propose instead, to use 
\begin{equation}
\ddr(z;s) = 
\begin{cases}
\textcolor{red}{s-z} &: \textcolor{red}{z\le n/2+s} \\
\textcolor{blue}{n+s-z} &: \textcolor{blue}{z > n/2+s}, \\
\end{cases}
\end{equation}
which switches over signs halfway to the next intersection. The expectation
can be done the two cases. Let $t=z-\frac{n}{2}$ for clarity, then the expectation is
\begin{equation}
\textcolor{blue}{\int_{-\infty}^{t} (n+s-z) p(s) ds} + 
\textcolor{red}{\int_{t}^{\infty} (s-z) p(s) ds}.
\end{equation}
These can be broken, grouped by content of the integrals, and had constants pulled out 
to produce
\begin{equation}
\begin{split}
& \textcolor{blue}{n \int_{-\infty}^{t} p(s) ds} + \\
& \quad \quad \textcolor{blue}{\int_{-\infty}^{t} s p(s) ds} + \textcolor{red}{\int_{t}^{\infty} s p(s) ds} + \\
& \quad \quad \textcolor{blue}{- z \int_{-\infty}^{t} p(s) ds} - \textcolor{red}{z \int_{t}^{\infty} p(s) ds}.
\end{split}
\end{equation}
From here, one can rewrite the first line as $n \F(z-\frac{n}{2})$. The second
line is $0$, since it groups to be $\int_\mathbb{R} s p(s) ds = 0$.
The third line is $-z$, since the integrals group to cover all the reals, and 
$\int_{-\infty}^{\infty} p(s) ds =1 $. This leaves the final result
\begin{equation}
\E[\ddr(z;s)] = n \F\left(z-\frac{n}{2}\right) - z
\end{equation}

The derivative of this expression is
\begin{equation}
\frac{\partial}{\partial z} \E[\ddr(z;s)] = n p\left(z-\frac{n}{2}\right) - 1
\end{equation}
because $\frac{\partial}{\partial z} \F(z) = p(z)$. This expression is $-1$ unless
$n p(z - \frac{n}{2})$ is large.

\subsection{Planes}

We are given a plane consisting of a normal $\nB \in \mathbb{R}^3$ with $||\nB||_2 = 1$ and offset $o$ (where points $\xB$ on the plane satisfy $\nB^T \xB + o = 0$). Our uncertainty about the plane's location in 3D is $\sB{\sim}N(\zeroB,\sigma^2 \IB)$ where $\IB$ is the identity matrix and $\zeroB$ a vector of zeros. Then $d_{\textrm{U}}(\xB;\sB)$ is the 3D unsigned distance function
\begin{equation}
d_{\textrm{U}}(\xB) = | \nB^T \xB + o |.
\end{equation}  
We will then compute the expected distance
\begin{equation}
E_{\sB}[d_\textrm{U}(\xB;\sB)] = \int_{\mathbb{R}^3} |\nB^T (\xB + \sB) + o|~(\sB)~d\sB.
\end{equation}

First, note that we are free to pick the coordinate system, and so we pick it so that the plane passes through the origin and is perpendicular to the $z$-axis. Thus, $\nB = [0,0,1]$ and $o = 0$. This does not require the plane to be perpendicular to the ray; this is merely placing the arbitrary coordinate system to be in a mathematically convenient configuration. Geometrically, this is precisely identical to the ray case: any uncertainty that is perpendicular to the plane does not alter the distance to the plane, leaving a single source of uncertainty (in $z$).

Algebraically, one can verify this as well. The distance to the plane for any point $\xB$ is $|\nB^T \xB - o|$. We can add the uncertainty about the plane's location by subtracting it off the point, placing the point at $\xB-\sB$. Then the distance is $|\nB^T (\xB - \sB) -o|$. Since $\nB = [0,0,1]$ and $o = 0$, this simplifies to $|\xB_z - \sB_z|$, where $\xB_z$ is the z coordinate of $\xB$ and likewise for $\sB_z$. The final expected value of the 3D distance is
\begin{equation}
\iint_{\mathbb{R}^2}
\left(\int_{-\infty}^{\infty} |\xB_z - \sB_z| p(\sB_z) d \sB_z\right) p(\sB_x) p(\sB_y) d \sB_x d\sB_y.
\end{equation}
Since the inner integral is constant with respect to $\sB_x$ and $\sB_y$, we can pull it out; we can also rewrite $|\xB_z - \sB_z|$ as $|\sB_z - \xB_z|$ to match convention, yielding:
\begin{equation}
\left(\int_{-\infty}^{\infty} |\sB_z - \xB_z| p(\sB_z) d \sB_z\right)
\iint_{\mathbb{R}^2}
 p(\sB_x) p(\sB_y) d \sB_x d\sB_y.
\end{equation}
The right double integral is $1$, leaving the expected unsigned distance function
\begin{equation}
\int_{-\infty}^{\infty} |\sB_z - \xB_z| p(\sB_z) d \sB_z.
\end{equation}

A few things follow from this setup. First, the minimum value at the real intersection will still be $\sigma \sqrt{2/\pi}$. Second, the only uncertainty that matters is the variance in the direction perpendicular to the plane: if $s \sim N(\zeroB,\textrm{diag}[\sigma_x^2, \sigma_y^2, \sigma_z^2])$, then only $\sigma_z^2$ controls the distortion of the UDF.
Finally, the expected distance along a ray that is not perpendicular to the plane be stretched proportionally to the cosine between the ray and the normal. Thus, the qualitative behavior (i.e., where the sign of the derivative changes) will be similar, but the rate at which things change will not be.

\clearpage

\end{document}